\documentclass[11pt]{article} % use larger type; default would be 10pt
\usepackage{soul}
\usepackage[utf8]{inputenc} % set input encoding (not needed with XeLaTeX)

\usepackage{url}

\usepackage{geometry} % to change the page dimensions
\usepackage{graphicx} % support the \includegraphics command and options

\usepackage{booktabs} % for much better looking tables
\usepackage{array} % for better arrays (eg matrices) in maths
\usepackage{paralist} % very flexible & customisable lists (eg. enumerate/itemize, etc.)
\usepackage{verbatim} % adds environment for commenting out blocks of text & for better verbatim
\usepackage{subfig} % make it possible to include more than one captioned figure/table in a single float
\usepackage{fancyhdr} % This should be set AFTER setting up the page geometry
\usepackage{sectsty}
\allsectionsfont{\sffamily\mdseries\upshape} % (See the fntguide.pdf for font help)
\usepackage[nottoc,notlof,notlot]{tocbibind} % Put the bibliography in the ToC
\usepackage[titles,subfigure]{tocloft} % Alter the style of the Table of Contents

\usepackage{natbib}
\usepackage{pdflscape}
\usepackage{afterpage}
\usepackage{chngcntr}

\usepackage{natbib}
\usepackage{amsfonts}
\usepackage{amsmath}
 
\usepackage[parfill]{parskip} % no indentation for new paragraphs

\usepackage[hidelinks]{hyperref} 

\title{Logistic Regression makes small LLMs \\strong and explainable ``tens-of-shot" classifiers}
\author{Marcus Buckmann and Ed Hill \\ Advanced Analytics, Bank of England\footnote{The views expressed in this paper are those of the authors, and not necessarily those of the Bank of England or its committees.}}

\begin{document}

\maketitle

\begin{center}
\large
\end{center}
\vspace{1.5cm}

\begin{abstract}
For simple classification tasks, we show that users can benefit from the advantages of using small, local, generative language models instead of large commercial models without a trade-off in performance or introducing extra labelling costs. These advantages, including those around privacy, availability, cost, and explainability, are important both in commercial applications and in the broader democratisation of AI. Through experiments on 17 sentence classification tasks (2--4 classes), we show that penalised logistic regression on the embeddings from a small LLM equals (and usually betters) the performance of a large LLM in the ``tens-of-shot" regime. This requires no more labelled instances than are needed to validate the performance of the large LLM. Finally, we extract stable and sensible explanations for classification decisions.

\end{abstract}

\section{Introduction}

This paper looks at simple sentence classification tasks, a widespread use-case of natural language processing. One model family that can be used for this task are flagship generative large language models such as GPT-4 \citep{openai2024gpt4}, Claude 3 \citep{anthropic2024claude} and Gemini \citep{team2023gemini}, which have exhibited impressive zero-shot performance across a wide range of tasks\footnote{In this work we compare against GPT-4 because in late 2023, when the work was performed, GPT-4 was a strong and standard benchmark (and remains so across many tasks \citep[][table 1]{anthropic2024claude}, also Section \ref{sec:lit}). Explicitly, we are not specifically discussing the use, advantages or disadvantages of GPT-4 versus other flagship LLMs, rather using it as a standard baseline against which to test our methods.}. However, these models have the disadvantages associated with any Cloud Software-as-a-Service around privacy, connectivity, and financial cost; as well as a lack of explainability and consistency (since changing or deprecating the model is outside the control of the end user). 

Alternatively, non-generative transformer models, such as fine-tuned versions of encoder-only models, can achieve good performance in text classification  \citep[e.g.][]{li2023chatgpt}. But fine-tuning such models comes at a significant cost in time, expertise and computation. 

\begin{figure}[!b]
% \rule[1ex]{\linewidth}{0.5pt}
\begin{center}
\includegraphics[width=.44\linewidth]{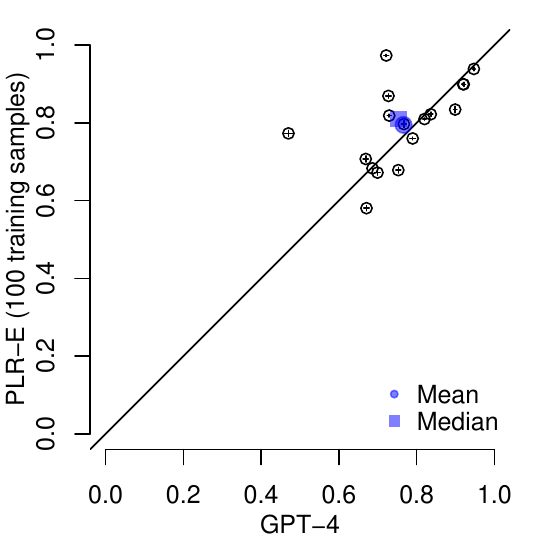} 
\includegraphics[width=.44\linewidth]{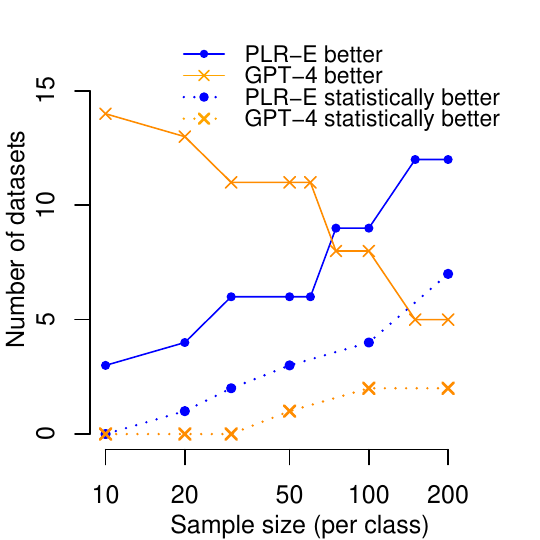} 

\caption{\small Left: Comparing the accuracies from GPT-4 and our method (PLR-E) over the 17 classification tasks. Right: We show, for increasing training sample sizes (divided by the number of classes), for how many datasets our method is outperforms than GPT-4 on accuracy, and vice versa (solid lines). We also count in how many of these cases we can confidently declare a `winner' above statistical noise (significance level: $\alpha  = 0.1$, two-sided)}
\label{first_figure}
% \includegraphics[width=.49\linewidth]{learning_res_rotten_yesno_llama2_7b-chat_q4_0_acc.pdf}
% \rule[1ex]{\linewidth}{0.5pt}
\end{center}
\end{figure}

In this study we show that we can realise the advantages of using local generative models for sentence classification , without incurring trade-offs in performance or significant other costs. 

Specifically, we show three key results, supported by a quantitative and qualitative discussion:
\begin{enumerate}
\item Penalised logistic regression (PLR) on the embeddings produced by a smaller, open-source and locally hosted generative model (we use quantised (q.4.0) Llama2 7B as a baseline \citep{touvron2023llama}) can equal or exceed the performance of GPT-4 on sentiment analysis and classification tasks. By contrast, the text output of the local models cannot compete with that of GPT-4 in most datasets.
\item In the majority of datasets, only 60--75 training samples per class are required to train a PLR model that beats GPT-4. By contrast, we require substantially more instances to obtain small enough confidence bounds to state that GPT-4 has a \textit{statistically} better performance than PLR and vice versa.

\item This PLR model provides stable word- and phrase- level explanations in this ``tens-of-shot" regime: We validate that these explanations are sensible against human annotations. \end{enumerate}

Taken together these results show a lack of performance trade-off (Result 1), or additional costs (Result 2), and that explanations of classification decisions -- a key possible advantage of local models -- can be realised in practice (Result 3). 

Figure \ref{first_figure} highlights the first two key results: The left panel compares the performance of GPT-4 and our PLR-E model (in these plots, a point lying above the diagonal shows a `win' for the model on the vertical axis) with an average of $30$ instances per class. It shows that, on average, our method outperforms GPT-4.

With fewer training instances, our supervised model is naturally weaker, and with more, stronger. This is reflected in the right panel, which shows the number of labelled instances beyond which one model can be declared the `winner' -- where the average (over randomly chosen training sets of that size) performance of one model is better than the other. We see that for  more than $75$ samples per class our PLR-E model outperforms GPT-4 in more than half of the tasks. The panel also shows the number of tasks on which each model can be declared the winner `statistically' for a given sample size. This corresponds to the real-world use case where we have a single set of labelled instances which is used to train the PLR-E model and to measure both its and GPT-4's performance.
The large sample sizes needed to reduce the statistical noise enough to confidently declare a winner provide substantial training sets leading to strong performance from the PLR-E method.

The paper is structured as follows: Section \ref{sec:lit} reviews the literature and Section \ref{sec:methods} describes our methodology. Section \ref{sec:learning} presents our results, beginning with a standard learning curve analysis, where models trained with varying amounts of data are assessed on their performance on a large test set. We show how performance is affected by prompting strategies and the choice of the language model. Section \ref{sec:uncertainty} then moves to the setting which mimics the case where only limited labelled data is available for both training and testing, like when a dataset or task is being approached for the first time. In this setting the quantity of labelled data influences both the performance of PLR models and the uncertainty of the performance estimates of both the PLR model and of the large LLM. Section \ref{sec:embeddings} discusses the structure and characteristics of the embeddings from local model and Section \ref{sec:explain} considers the stability and accuracy of feature importance explainability methods. Section \ref{conclusion} concludes. 

\subsection{Literature review} \label{sec:lit}

Recent studies have shown that large general-purpose LLMs such as GPT-4 are competitive text classifiers \citep{rathje2023gpt}, including on tasks that require specialised domain expertise \citep{savelka2023}. \citet{zhang2023sentiment} conducted a large scale empirical assessment across 13 sentiment analysis tasks on 26 data sets and conclude that ``Even in a zero-shot setting, [LLMs'] performance can match or surpass fine-tuned smaller language models, and with little sensitivity to different prompt designs." However, other studies found that human annotators or specialised models calibrated on human annotations outperform large LLMs \citep{liyanage2023gpt,li2023chatgpt,toney2024ai}.

Our approach -- learning a linear model on the hidden states of a large neural networks -- is known in the literature as linear probing \citep{belinkov2022probing,alain2016understanding}. This technique has mostly been used to understand what information hidden states of language models represent \citep{jawahar2019does,zhu2024language,gurnee2023language,chen2023beyond,campbell2023localizing}, but has previously been used to improve the accuracy of generative language model predictions: \citet{cho2023palp} used linear probing on 12 classification tasks and showed that it outperforms in-context learning for both GPT-J and GPT-2. As in our work, the best performance was obtained when augmenting the text to classify with a prompt stating the classification task . Other work \citep{jiang2023scaling, zhang2024simple} similarly uses a surrounding prompt to improve the quality of embeddings for downstream tasks.

Instead of learning a linear model on the embeddings, \citet{abbas2024enhancing} used linear probing to calibrate the token probability, an approach we use as a baseline in our paper as well (PLR-L below).

Another strand of the literature aims to improve the predictions of generative models using weak supervision on embeddings. \citet{chen2022shoring} and \citet{cho2023celda} use the embeddings of the generative language model and estimate labels exploiting local smoothness of the embedding space. \citet{guha2024} use embeddings from other models for weak supervision. %\citet{wang2023improving} and \citet{SFR-embedding-2} fine tune LLMs to improve their usefulness as sentence embedding models.

Our paper also relates to the literature on edge computing, and to the cost-effective use and `democratisation' of AI. Despite recent advances in the efficiency of fine-tuning \citep{hu2021lora,liu2024dora}, fine-tuning BERT-type models to a bespoke classification problem requires the collection of appropriate data, and access to sufficient computational resources and expertise. By contrast, learning a simple linear model on top of the embedding of an open-source generative model is computationally cheap, and penalised logistic regression is amongst the best known and widely understood prediction methodologies across disciplines. While inference with a several billion parameter model as in this paper is substantially slower than using sub-billion parameter previous-generation models, we note that the model used is already a year old, and improvements in model quantization and pruning \citep{lin2023awq,sun2023simple,dettmers2023spqr,ma2024era} and the development of more efficient model architectures \citep{gu2023mamba,peng2023rwkv} will continue to decrease the memory and computational footprint of models with equivalent performance. 

Furthermore, using proprietary LLMs such as GPT-4 or Gemini for text classification has several disadvantages including the exposure of private data (both to the provider of the service and, possibly, the communication system), payment for the service, the requirement of (stable) internet access, and the lack of model consistency against deprecation or updating. Not having access to the model's parameters removes flexibility, limiting options around fine-tuning and model adaptation, and also means that tasks which cannot be easily expressed as a prompt (for example classifying text according to an individual's personal preferences) cannot be performed.\footnote{The relative importance of these issues will depend on the user's situation. Regarding cases where the user wants to limit, but need not eliminate, external LLM usage, this work complements model cascade and selection methods \citep{chen2023frugalgpt,vsakota2023fly}, possibly superseding them in the simple classification case.} And while many approaches exist to explain various aspects of the behaviours and limitations of LLMs \citep{zhao2023explainability}, these usually cannot be applied without having access to the full model. Explainabilty is important both to inform their practical commercial use, and for diagnosing and avoiding pathological behaviours for robust performance \citep[e.g.][]{du2023shortcut} and legal compliance \citep{aiact2024}.  

Our case study around explainability in Section \ref{sec:explain} uses the Financial Phrases dataset \citep{Malo2014GoodDO} which is of interest to us due to our field of work within the United Kingdom's Central Bank, and this paper therefore links specifically to the widespread use of text classification in economics and finance. In central banking more specifically, text classification has been used on central bank communications for sentiment analysis \citep{bennani2017,picault2017, lee2021,chen2023} and other classification tasks \citep{bertsch2022,pfeifer2023}. Furthermore, text classification models can play a crucial role in banking supervision by helping to detect the sentiment and topics in reports or minutes of board meetings of supervised firms, and can also be building blocks of macroeconomic forecasting methods that learn from text \citep{kalamara2022,ellingsen2022}.

\subsection{Methodology} \label{sec:methods}

Our method has three steps: prompt construction, text embedding, and penalised logistic regression (PLR). We will describe these steps with signposts to the robustness and other checks. We will evaluate performance, using accuracy as the primary metric, on 17 classification tasks from diverse domains, including movie reviews, news headlines, Youtube comments, tweets, and Reddit posts. We do not make use of all instance of the larger data sets but rely on sub-sampling. The data sets and the sizes of the samples we draw are described in detail in Appendix \ref{datasets}. 

\subsubsection{Prompt construction} \label{sec:prompt_construction}

We add a contextualising prefix and a suffix indicating the classification task to be performed around the text to be classified. An example from the Financial Phrases dataset is 

\texttt{\small  I am extremely delighted with this project and the continuation of cooperation with Viking Line.}

This is extended to 

\texttt{\small The following sentence contains financial news: I am extremely delighted with this project and the continuation of cooperation with Viking Line. Does the sentence have (a) positive, (b) negative, (c) neutral sentiment? Answer:\ (}

Similarly, for a classification problem with two classes, such as the irony data set we extend the text 

\texttt{\small Today is going to be a great day . \#not. } 

to 

\texttt{\small Consider the following tweet: Today is going to be a great day . \#not.  Is this tweet ironic? Answer with Yes or No. Answer:}

We show that adding this surrounding text substantially improves performance relative to using just the text and that the results are robust to the precise wording and direction of the surrounding prompt (Section \ref{sec:rob_exact_text}).

\subsubsection{Embedding}

The prompt is sent to the LLM which outputs the embedding which will be used for the classification task. The embedding is the final layer activations before the prediction head, which is $4096$ dimensional for our baseline model. We show that our results are robust to the size and quantisation of the model in Section \ref{sec:rob_mod_size_quant}.

\subsubsection{Text prediction} 

We also assess the quality of the text output. The LLMs usually provide an answer where the first token is one of the candidate tokens specificed in the prompt (e.g. ``Yes", or ``No" for binary classification tasks or ``a", ``b", or ``c" for a three-class task). To catch cases where the token with the maximum logit out of all possible tokens is not in the candidate set, we instead record as the answer the token with the maximum logit out of the relevant candidate tokens.

\subsubsection{Penalised logistic regression}

Finally, we perform penalised logistic regression (PLR) on the embeddings. Specifically, we use ridge regression, that is, $l_2$ regularisation. In the binary case this minimises the log-likelihood
\begin{equation}
\ell = \sum_{k=1}^K y_k \ln(p(\boldsymbol{e_k}))+\sum_{k=1}^K (1-y_k) \ln(1-p(\boldsymbol{e_k})) + \lambda \|\boldsymbol{a}\|^2 \label{eq:plr}
\end{equation}
for the $K$ instances with class $y_k \in \{0,1\}$, embedding vector $\boldsymbol{e}_k$, and regularisation parameter $\lambda$ to find $a_0$ and $\boldsymbol{a}$ in the function. 
\begin{equation}
p(\boldsymbol{e}) = (1 + e^{- (a_0 + \boldsymbol{a}.\boldsymbol{e})})^{-1}
\end{equation}
$\boldsymbol{a}$ is the normal to the classification surface and $a_0 + \boldsymbol{a}.\boldsymbol{e} = \ln (p / (1-p))$ is the log-odds.

\subsubsection{Implementation} 

\textbf{Language models.  } We use the quantisation of the generative language models provided by \texttt{TheBloke} on \href{https://huggingface.co/TheBloke/}{Hugging Face}. For the Llama models we use the \textit{ggml} quantisation, for the \href{https://huggingface.co/stabilityai/stablelm-zephyr-3b}{Zephyr model}, which we tested at later point in time, we use the more recent \textit{gguf} quantisation. Additionally we tested two sentence embedding models. 
First, \texttt{bge-large-en-v1.5} \citep{bge_embedding} is a 326 million parameter $1024$-dimensional embedding model that performed best on the MTEB benchmark \citep{muennighoff2022} both across all tasks and on classification tasks specifically, as of September 9, 2023. The model can be downloaded from \href{https://huggingface.co/BAAI/bge-large-en}{Hugging Face}. Second, we tested the \texttt{ada2} embedding model, which is accessible via the OpenAI API.

\textbf{Penalised logistic regression.  } We estimate ridge regression using the R package \texttt{glmnet}. Unless stated otherwise, we use the default regularisation path (100 different values for $\lambda$ and choose the model with the lowest degree of regularisation instead of conducting a hyperparameter search. Section \ref{sec:embeddings} shows that our model is robust to the choice of $\lambda$: fixing its value makes our approach computationally cheaper and less complex, increasing its value for practitioners.

\section{Performance comparison: Small training set, large test set}  \label{sec:learning}

We begin in the usual setting in the literature, examining the learning curves as we increase the training set size, with a large pool of samples to draw the training and test set from. Specifically, we sample 20\% of the observations of a data set as the test set and sample training samples of increasing size from the remaining 80\% of observations. To obtain stable performance estimates we repeat this procedure 50 times. Confidence intervals show $\pm 1.96$ standard errors around the mean performance estimate and are estimated using bootstrapping. 

\def\ww{.3}
\begin{figure}[h!t]
\includegraphics[width=\ww\linewidth]{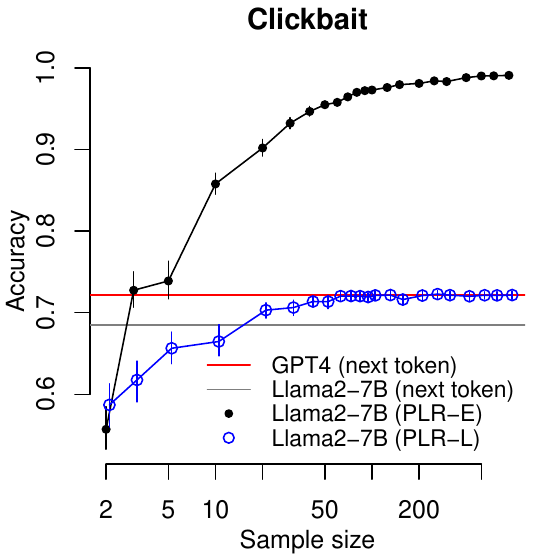}
\includegraphics[width=\ww\linewidth]{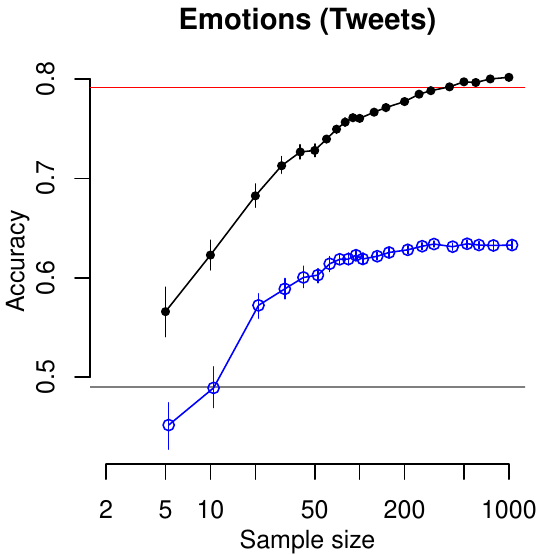}
\includegraphics[width=\ww\linewidth]{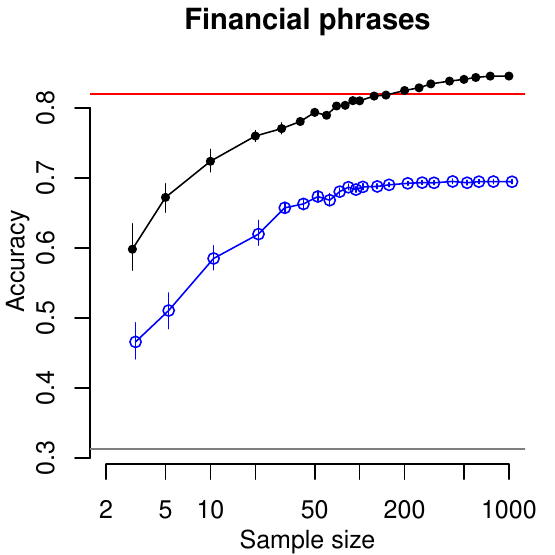}
\includegraphics[width=\ww\linewidth]{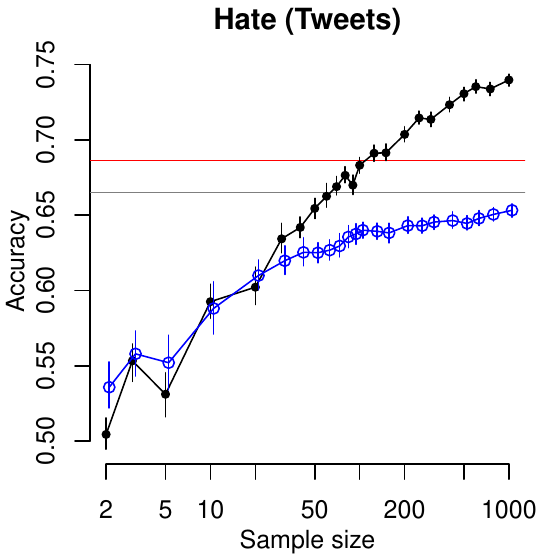}
\includegraphics[width=\ww\linewidth]{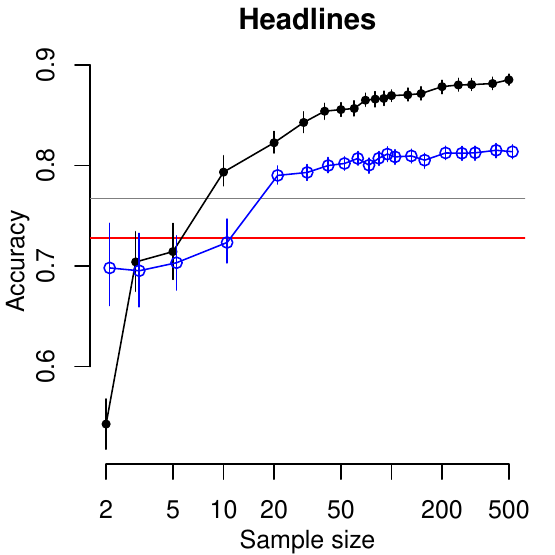}
\includegraphics[width=\ww\linewidth]{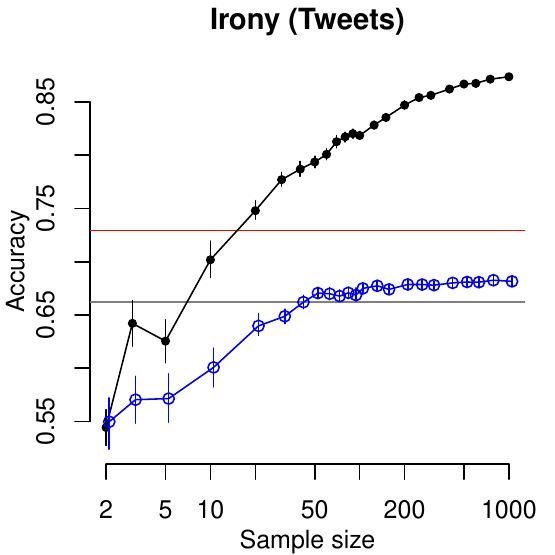}
\includegraphics[width=\ww\linewidth]{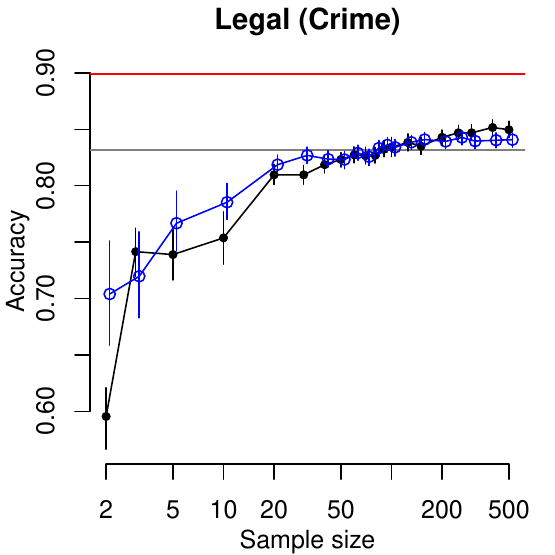}
\includegraphics[width=\ww\linewidth]{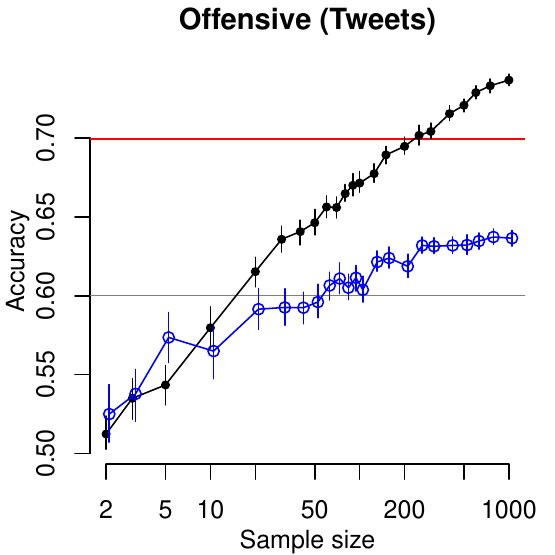}
\includegraphics[width=\ww\linewidth]{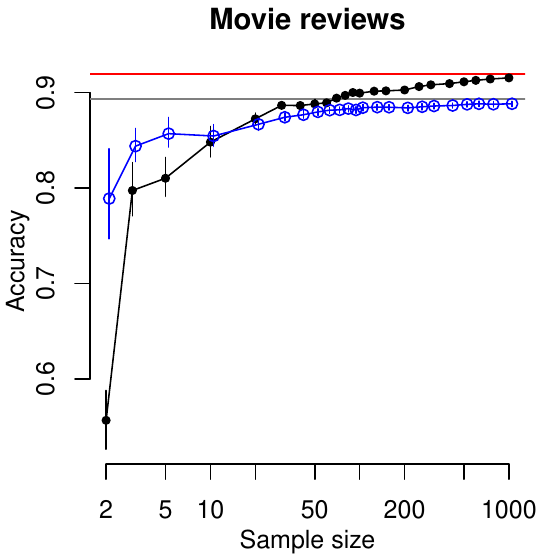}
\includegraphics[width=\ww\linewidth]{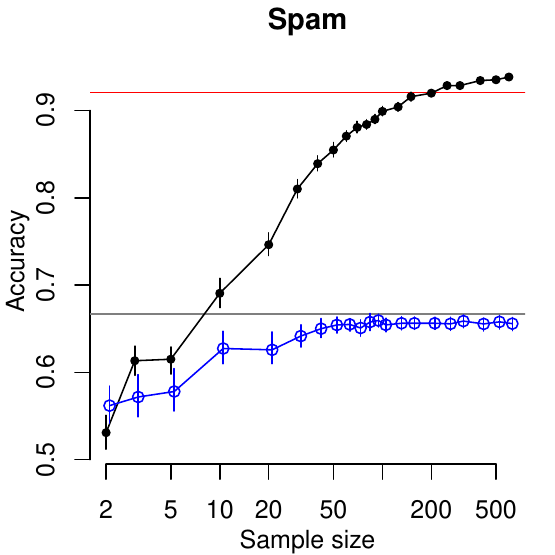} 
\includegraphics[width=\ww\linewidth]{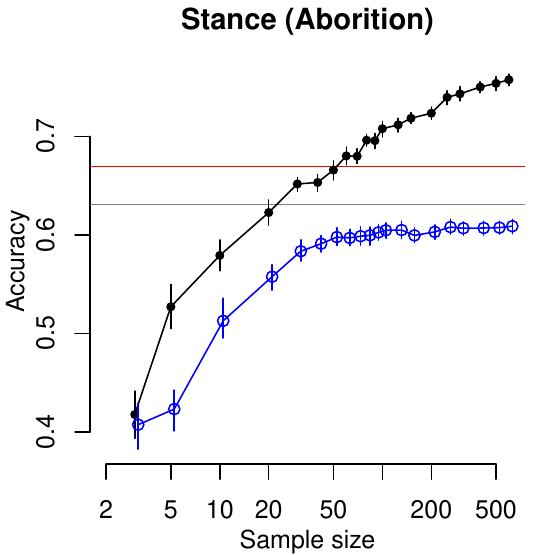}
\includegraphics[width=\ww\linewidth]{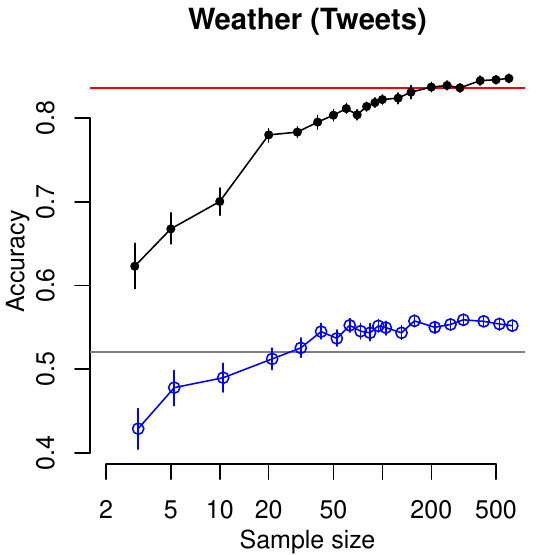}

\caption{\small The accuracies of the zero-shot next token text predictions from GPT-4 and Llama2-7B, along with with the learning curves for the PLR-L and PLR-E methods applied to our baseline model (Llama2-7B q4.0).}
\label{fig_baseline_learning}
\end{figure}

Figure \ref{fig_baseline_learning} demonstrates the main result in this section for 12 classification problems (the remaining learning curves being shown in Figure \ref{fig_baseline_learning_rest} in the Appendix). In each case we show the accuracies of zero-shot next token prediction for both GPT-4 and the baseline (Llama2 7B q4.0) model, along with penalised logistic regression models trained on the baseline model's next token logits (PLR-L) and on the baseline model's embedding (PLR-E). 

Despite the zero-shot next token accuracy of the baseline model underperforming relative to GPT-4, the accuracy of the PLR-E method becomes comparable to or exceeds that of GPT-4 as the training sample size is increased. PLR-L is inferior to PLR-E but significantly exceeds the performance of the baseline model's next token prediction in several data sets.

Even with sample sizes of $10$ observations, PLR-E outperforms the baseline model's next token prediction in 9 of the 17 data sets. This is surprising due to the very small sample size and the high dimensionality of the embedding space. 

These results also hold when using the F1 macro score as performance metric -- the analogous learning curves are shown in Figure \ref{fig_baseline_learning_f1} in the appendix.

Figure \ref{fig_vs_GPT} shows a different view of the learning curves. From left to right, it compares the accuracy GPT-4 against our baseline model using zero-shot next token prediction, PLR-L at a sample size of 100, and PLR-E at sample sizes 10 and 100.

Table \ref{tab_learning_acc} shows the performance of PLR-E at different sample sizes. The last two columns show at what sample size (divided by the number of classes) PLR-E outperforms GPT-4.  We observe that PLR-E eventually beats GPT-4 in all but four datasets.\footnote{Note that the Central Banking data set constitutes a substantially more complex classification problem than the other data sets in our pool. See data set descriptions in Appendix \ref{datasets}} The last column replicates this analysis but here the embeddings on which we train PLR-E are only based on the sentences, omitting the instructions (see Section \ref{sec:instructions}).

\addtolength{\tabcolsep}{-1.5pt}    
\begin{table}
\begin{small}
\begin{tabular}{lrrrrrrrrr}
  \hline
Dataset & Classes & \multicolumn{5}{c}{Sample size} & GPT-4 & \multicolumn{2}{c}{Min.\ sample (per class) } \\
 &  & \multicolumn{5}{c}{(PLR-E, full prompt)} &Token& \multicolumn{2}{c}{for PLR-E win} \\
& & 10 & 30 & 100 & 250 & 400 && Full prompt & Sentence\\ 
  \hline
Central banking & 3 & 0.46 & 0.53 & 0.58 & 0.62 & 0.63 & 0.67 & - & - \\ 
  Clickbait & 2 & 0.86 & 0.93 & 0.97 & 0.98 & 0.99 & 0.72 & 2 & 5 \\ 
  Emotions (Tweets) & 4 & 0.62 & 0.71 & 0.76 & 0.78 & 0.79 & 0.79 & 100 & - \\ 
  Financial phrases & 3 & 0.72 & 0.77 & 0.81 & 0.83 & 0.84 & 0.82 & 67 & - \\ 
  Hate (Tweets) & 2 & 0.59 & 0.63 & 0.68 & 0.71 & 0.72 & 0.69 & 62 & 100 \\ 
  Headlines & 2 & 0.79 & 0.84 & 0.87 & 0.88 & 0.88 & 0.73 & 5 & 10 \\ 
  Irony (Tweets) & 2 & 0.70 & 0.78 & 0.82 & 0.85 & 0.86 & 0.73 & 10 & 15 \\ 
  Legal (Crime) & 2 & 0.75 & 0.81 & 0.83 & 0.85 & 0.85 & 0.90 & - & - \\ 
  Legal (Money) & 2 & 0.70 & 0.74 & 0.80 & 0.82 & 0.83 & 0.77 & 25 & 75 \\ 
  Legal (Work) & 2 & 0.92 & 0.93 & 0.94 & 0.94 & 0.95 & 0.95 & 250 & - \\ 
  Offensive (Tweets) & 2 & 0.58 & 0.64 & 0.67 & 0.70 & 0.72 & 0.70 & 125 & - \\ 
  Movie reviews & 2 & 0.85 & 0.89 & 0.90 & 0.91 & 0.91 & 0.92 & - & - \\ 
  Spam & 2 & 0.69 & 0.81 & 0.90 & 0.93 & 0.93 & 0.92 & 125 & 100 \\ 
  Stance (Aborition) & 3 & 0.58 & 0.65 & 0.71 & 0.74 & 0.75 & 0.67 & 20 & 83 \\ 
  Stance (Atheism) & 3 & 0.67 & 0.72 & 0.77 & 0.81 & 0.82 & 0.47 & 1 & 2 \\ 
  Stance (Feminism) & 3 & 0.59 & 0.64 & 0.68 & 0.71 & 0.72 & 0.75 & - & - \\ 
  Weather (Tweets) & 3 & 0.70 & 0.78 & 0.82 & 0.84 & 0.84 & 0.84 & 67 & - \\ \hline
 Mean & 2 & 0.69 & 0.75 & 0.80 & 0.82 & 0.83 & 0.77 & &  \\ 
  Median & 2 & 0.70 & 0.77 & 0.81 & 0.83 & 0.84 & 0.75 & &  \\ \hline
\end{tabular}
\caption{\small The accuracy of PLR-E at different sample sizes, the accuracy of GPT-4's next-token prediction, and the minimum sample size (per class, for the full prompt and with the sentence only) where PLR-E wins versus GPT-4's next token prediction.}
\label{tab_learning_acc}
\end{small}
\end{table}
% \multicolumn{5}{c}{PLR-E, full prompt} 
\addtolength{\tabcolsep}{1.5pt}    

\begin{figure}[!tb]
% \rule[1ex]{\linewidth}{0.5pt}
\begin{center}
\includegraphics[width=.24\linewidth]{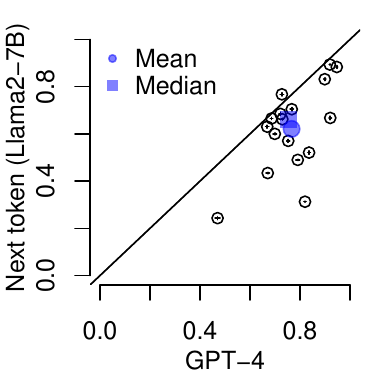}
\includegraphics[width=.24\linewidth]{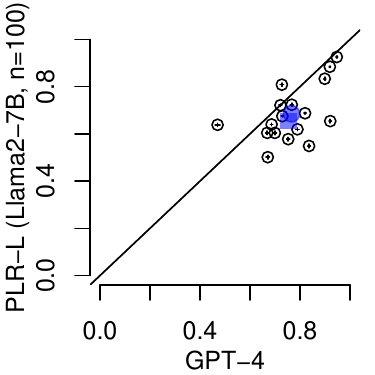} 
\includegraphics[width=.24\linewidth]{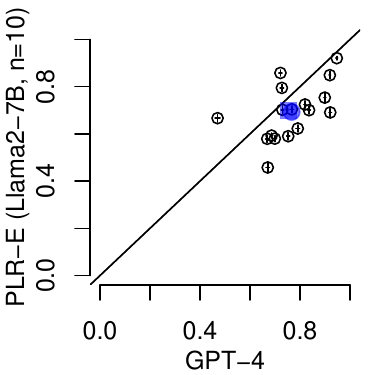} 
\includegraphics[width=.24\linewidth]{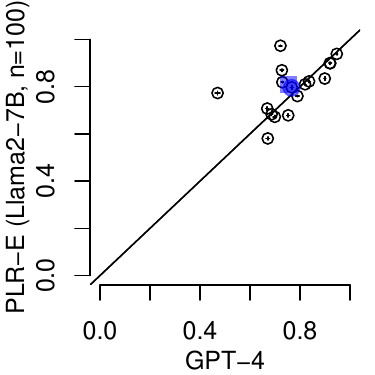} 

\caption{\small Comparing GPT-4 to our baseline model. From left, using the baseline model's next token prediction, learning from its logits (PLR-L), and learning from its embeddings (PLR-E).}
\label{fig_vs_GPT}
\end{center}
\end{figure}

\subsection{Relating next token prediction, logits, and embeddings}

Considering a two-class problem, we can decompose the performance gap between next-token prediction and the PLR-E method. We use $l_+$ and $l_-$ to denote the logits associated with the responses for the first and second class (the logits of `Yes' and `No', or `a' and `b', for example), and $\boldsymbol{e}_+$ is the vector used in the prediction head to extract the logits such that $l_+ = \boldsymbol{e} . \boldsymbol{e}_+$, and similarly for $l_-$. The log-odds can be decomposed as
\begin{align*}
    a_0 + \boldsymbol{a}.\boldsymbol{e} &= (l_+ - l_-) \\
    &+ \left[a_{\pm} + (a_+ - 1) l_+ + (a_- + 1) l_-\right] \\
   &+ \left[\boldsymbol{a}.\left(\boldsymbol{e} - \sum_{i = +,-} \frac{(a_i - \boldsymbol{a}. \boldsymbol{e}_i) + \boldsymbol{a}. \boldsymbol{e}_i}{\boldsymbol{a}. \boldsymbol{e}_i} \boldsymbol{e}_i \right) + a_0 - a_{\pm}\right]
\end{align*}

The first term is the log-odds associated with next-token prediction: If $l_+ - l_- > 0$ then $+$ is predicted, and $-$ otherwise.

The first square-bracketed term still only considers the log-odds but represents the learned correction to the position of the classification surface, which may need translating and rotating in the $\{l_+,l_-\}$-space to best fit the data. Figure \ref{fig_baseline_learning} shows that the model was well positioned in some cases (i.e. $a_{\pm}$, $a_+ - 1$, and $a_- - 1$ are all small) and so the zero-shot next-token prediction results in an accuracy very close to the large sample limit for PLR-L. PLR-L always eventually exceeds the performance of the zero-shot next-token case as the position of the classification surface is improved.

The second square bracketed term results from the PLR-E model being able to discriminate based on directions outside of the plane spanned by $\boldsymbol{e}_+$ and $\boldsymbol{e}_-$. This allows it to bring in features which discriminate between training instances but which were not projected into the logit directions, or were, but their contribution was swamped by other variance. Examining the learning curves in Figures \ref{fig_baseline_learning} and \ref{fig_baseline_learning_rest} suggests that despite not being in the logit directions these features are generally well separated, leading to the model learning rapidly with new training instances.

\subsection{Robustness: Omitting instructions from the prompt} \label{sec:instructions}

To investigate the role of the instructions for the accuracy of PLR-E we replicate the analysis, but now train PLR-E on embeddings found when omitting the surrounding instructions from the prompt. Additionally, we train PLR-E models (both on the full prompt and the sentence only) on the embeddings from two standard non-generative sentence embedding models, \texttt{bge-large-en-v1.5} and \texttt{ada-002}. Neither embedding model is instruction-tuned but both have different sizes and lineages. 

Figure \ref{fig_plr_learning} compares the different models and prompts in learning curves for 12 data sets with the remaining datasets shown in Figure \ref{fig_plr_learning_rest} in the appendix. First, we observe that in most datasets PLR-E on Llama2 embeddings is more accurate when including the instructions in the prompt. Second, we find that the sentence embedding models \texttt{bge} and \texttt{ada2} are mostly inferior to PLR-E, particularly at small sample sizes. For both sentence embedding models adding the contextualising prompt does not improve, and often hurts, the performance. 

\begin{figure}[!ht]
\rule[1ex]{\linewidth}{0.5pt}

\includegraphics[width=.325\linewidth]{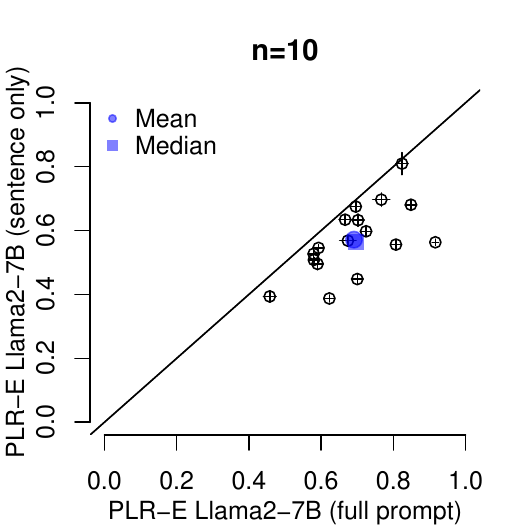}
\includegraphics[width=.325\linewidth]{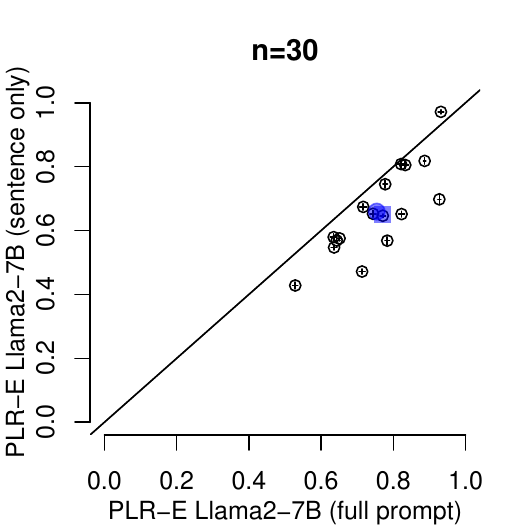}
\includegraphics[width=.325\linewidth]{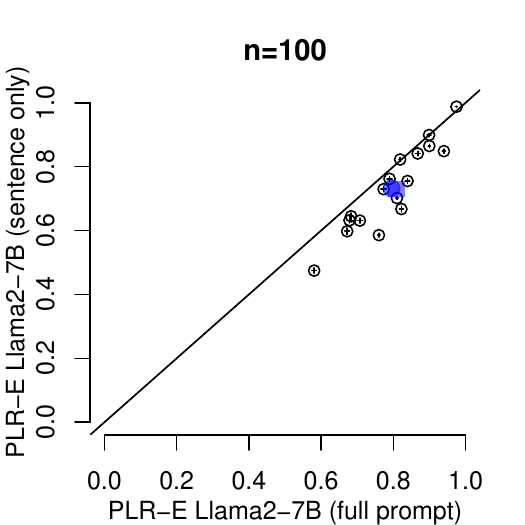}

\includegraphics[width=.325\linewidth]{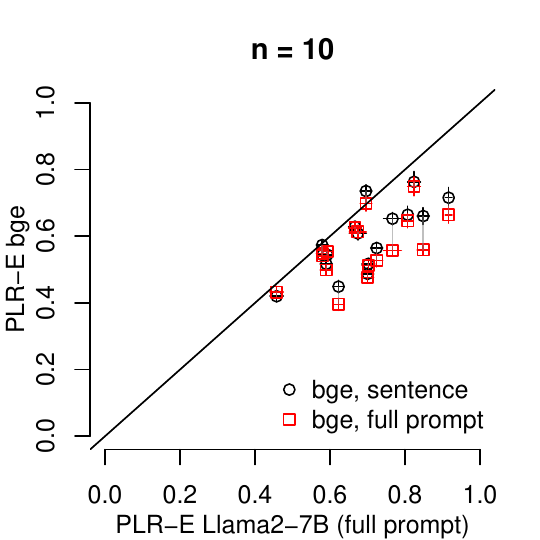}
\includegraphics[width=.325\linewidth]{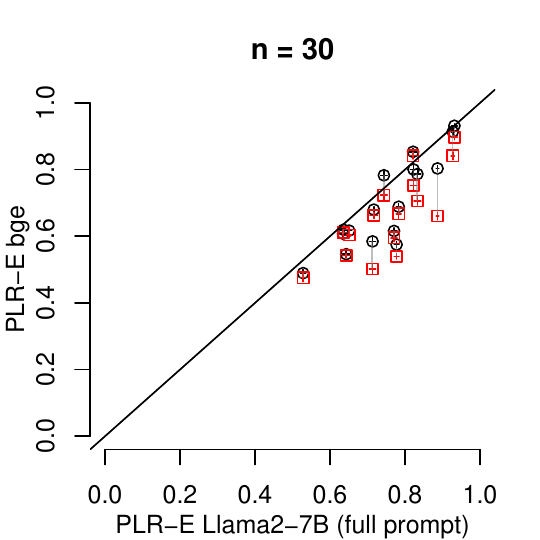}
\includegraphics[width=.325\linewidth]{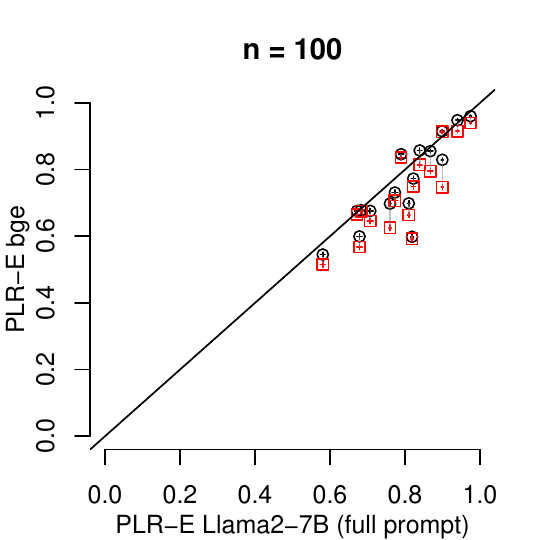}

\includegraphics[width=.325\linewidth]{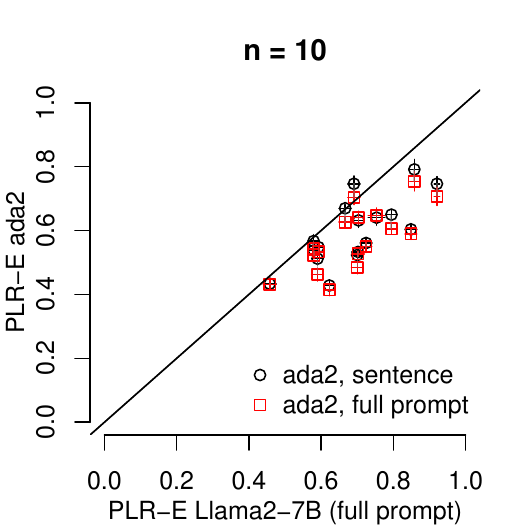}
\includegraphics[width=.325\linewidth]{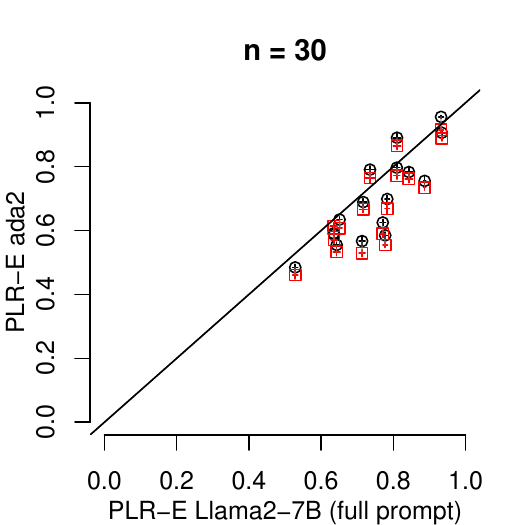}
\includegraphics[width=.325\linewidth]{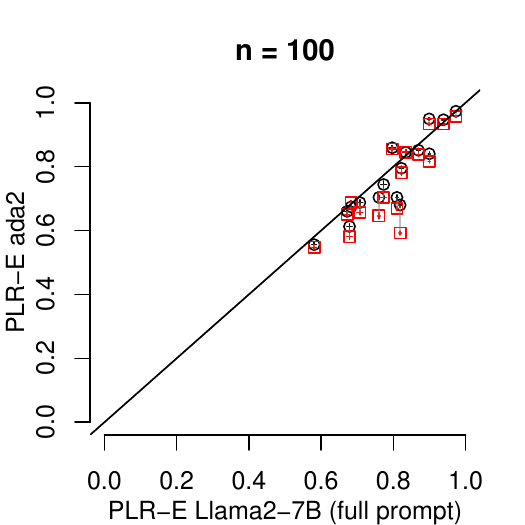}

\caption{\small Instruction prompting. Top panel: Comparing the performance when using PLR-E on our baseline model with and without surrounding instructions. Middle and bottom panels: Comparing our baseline PLR-E (with instructions) against applying PLR-E to the embeddings from two sentence embedding models with (red arrows) and without instructions (black crosses).}
\label{fig_instructions}
\end{figure}

Figure \ref{fig_instructions} summarise the findings from the learning curves. It compares our baseline approach (horizontal axis) to PLR-E trained on Llama-2 embeddings without instructions (first row) and to PLR-E trained on \texttt{bge} (second row) and \texttt{ada2} (thrid row) embeddings. 

\def\ww{.315}
\begin{figure}[h!t]
\includegraphics[width=\ww\linewidth]{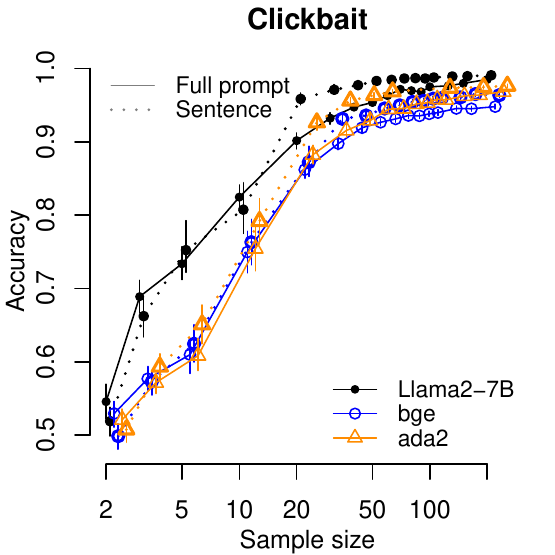}
\includegraphics[width=\ww\linewidth]{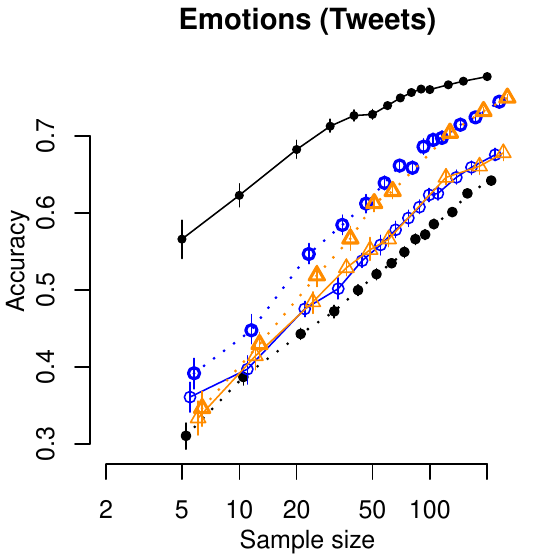}
\includegraphics[width=\ww\linewidth]{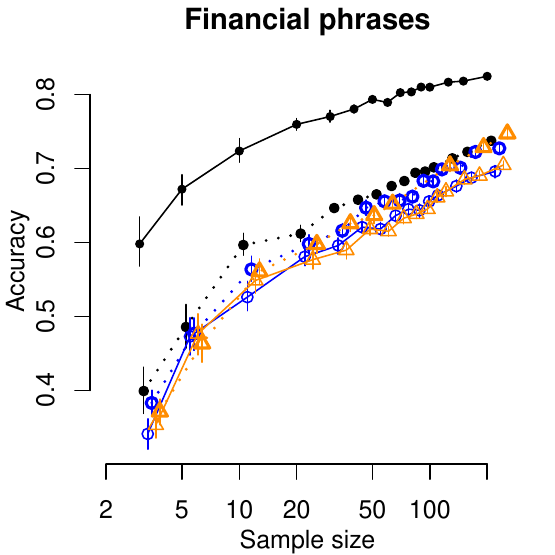}
\includegraphics[width=\ww\linewidth]{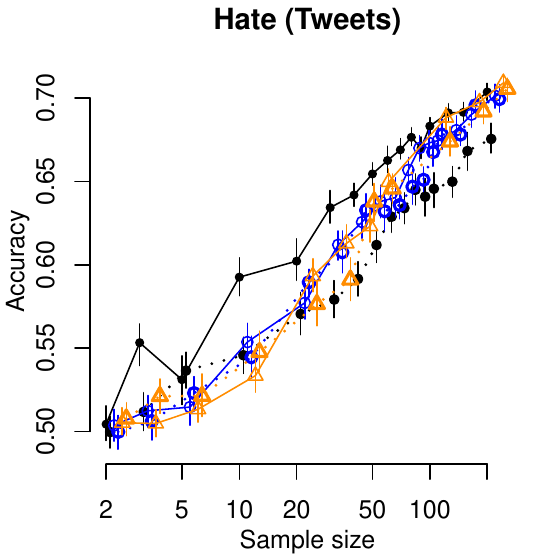}
\includegraphics[width=\ww\linewidth]{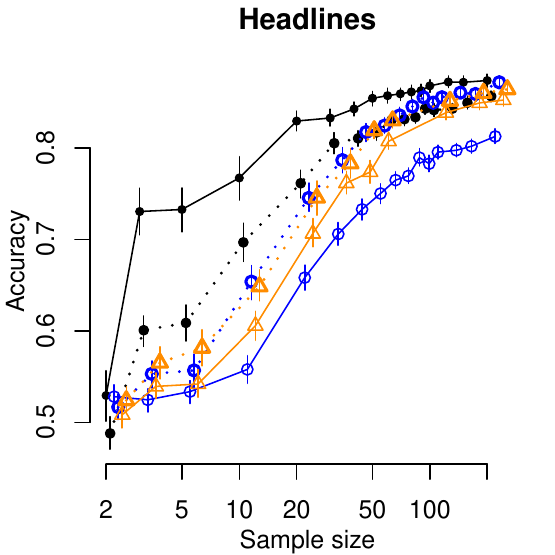}
\includegraphics[width=\ww\linewidth]{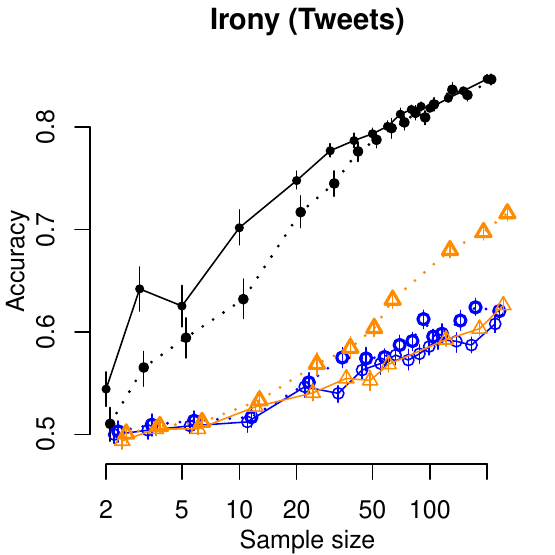}
\includegraphics[width=\ww\linewidth]{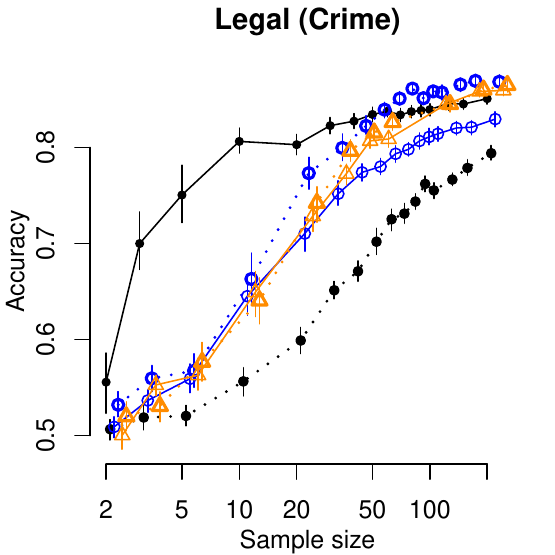}
\includegraphics[width=\ww\linewidth]{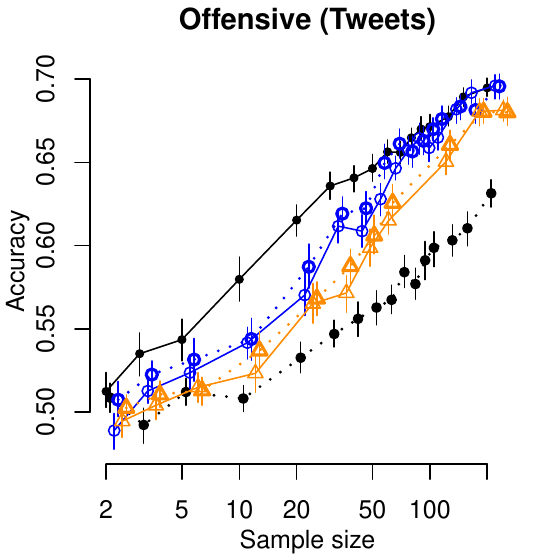}
\includegraphics[width=\ww\linewidth]{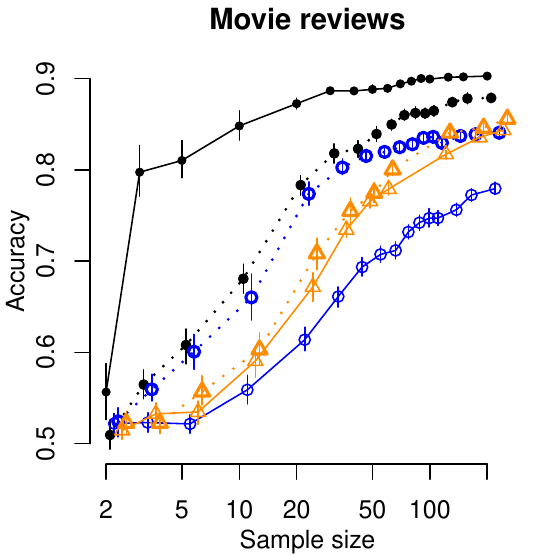}
\includegraphics[width=\ww\linewidth]{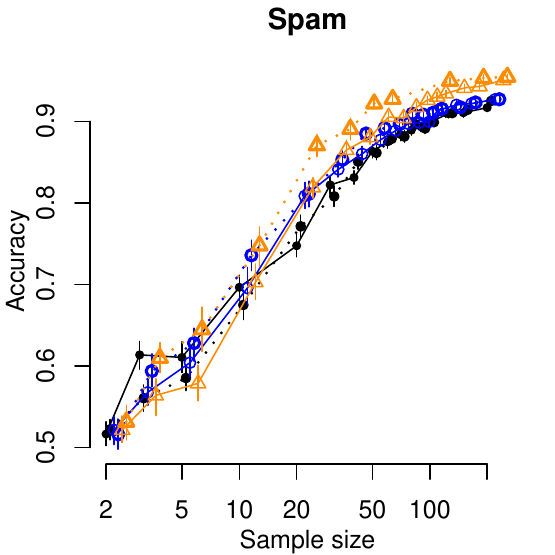} 
\includegraphics[width=\ww\linewidth]{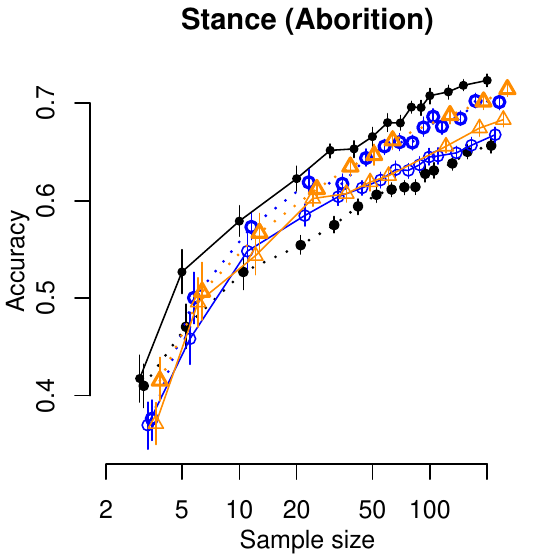}
\includegraphics[width=\ww\linewidth]{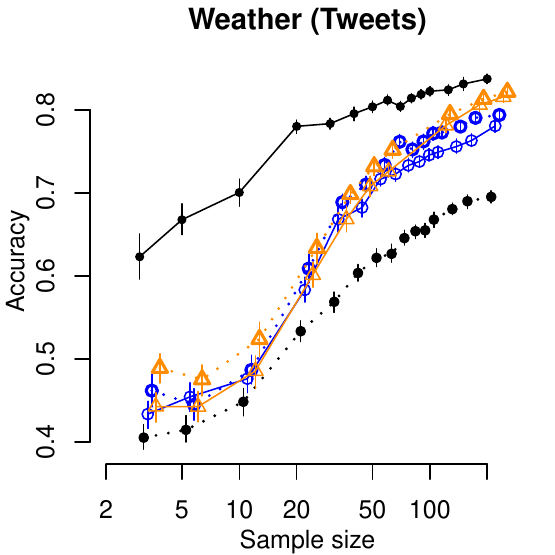}

\caption{\small The accuracy of PLR-E when trained on embeddings from different models and promptings, c.f. Figure \ref{fig_baseline_learning}.}
\label{fig_plr_learning}
\end{figure}

\subsection{Robustness: Choice of prefix and suffix}  \label{sec:rob_exact_text}

At small sample sizes, adding instructions boosts the performance of PLR-E. But are the models also sensitive to the wording of the instructions?
To test this, we attach different prefixes and suffixes, shown in Table \ref{tab_prompts}, to the sentence.

Figure \ref{fig_prompts} shows learning curves of PLR-E model for the different prompts in two data sets. In both data sets, we observe that those prompts with minimal or no instructions (curves 5 - 7) perform worse, with accuracies between 0.1 and 0.2 lower for small training set sizes. For other prompt configurations we do not observe substantial differences in their performance (with a spread in accuracies of around 0.05) suggesting that PLR-E is robust to the exact prompt specification as long as the instructions are complete.

{\renewcommand{\arraystretch}{1.5}% for the vertical padding
\begin{table}[h!]
\rule[1ex]{\linewidth}{0.5pt}
\centering
\small
\begin{tabular}{p{2.5cm} | p{5.5cm}|p{5.5cm}}
Type & Prefix & Suffix \\ 
\hline
Baseline & The following sentence contains financial news & Does the sentence have (a) positive, (b) negative, (c) neutral sentiment?\\ %baseline
No a,b,c,d & The following sentence contains financial news & Does the sentence have positive, negative, neutral sentiment? Answer with a single word.\\ % direct
No prefix & -- & Does the sentence have positive, negative, neutral sentiment?\\ %direct_no_prefix
No choices & -- & What is the sentiment of the sentence?\\ %no_option
Minimal instructions & Sentiment of & --\\ %direct_minimal
Distortions & The following sentence contains financial news & Does the sentence have (a) positive, (b) negative, (c) neutral sentiment? X9asd7bV \\ %baseline
No instructions & -- & --\\  %noq
No instructions + distortion & -- & X9asd7bV\\ % distortion
\end{tabular}
\caption{\small Different types of prompts are shown in a case where the possible answers are positive, negative and neutral. `X9asd7bV' is an example random alphanumeric string of that form which is inserted into the prompt in that position.}
\label{tab_prompts}
% \rule[1ex]{\linewidth}{0.5pt}
\end{table}
{\renewcommand{\arraystretch}{1}% for the vertical padding 

\begin{figure}[!h]
\includegraphics[width=0.5 \linewidth]{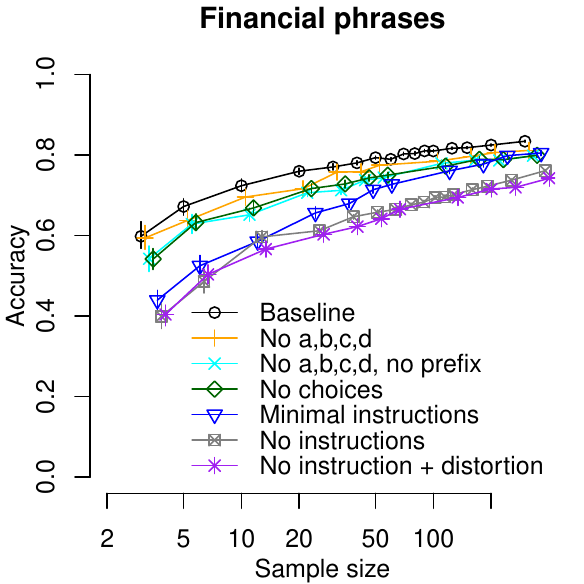}
\includegraphics[width=0.5 \linewidth]{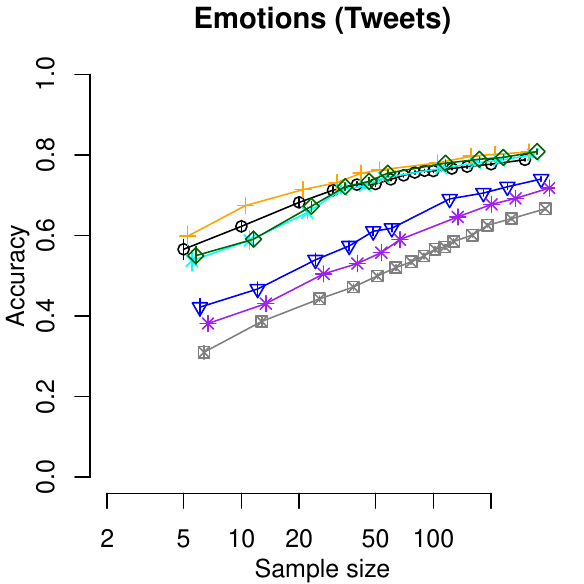}
\caption{\small Learning curves for surrounding prompts with different prefix and suffix. See Table \ref{tab_prompts} for the specifications of the prompts in the legend.}
\label{fig_prompts}
\end{figure}

\begin{figure}[h!t]

\rule[1ex]{\linewidth}{0.5pt}

\centering

\includegraphics[width=.325\linewidth]{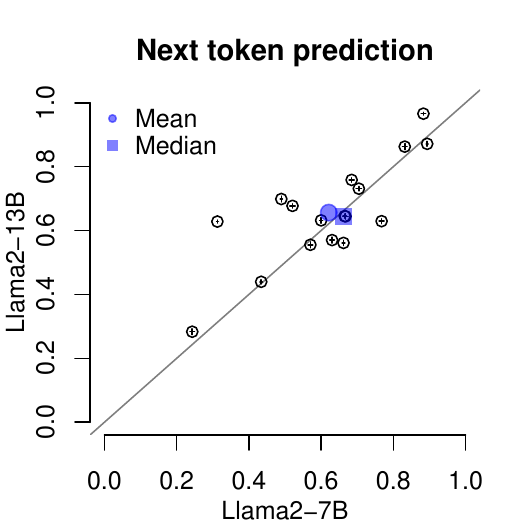}
\includegraphics[width=.325\linewidth]{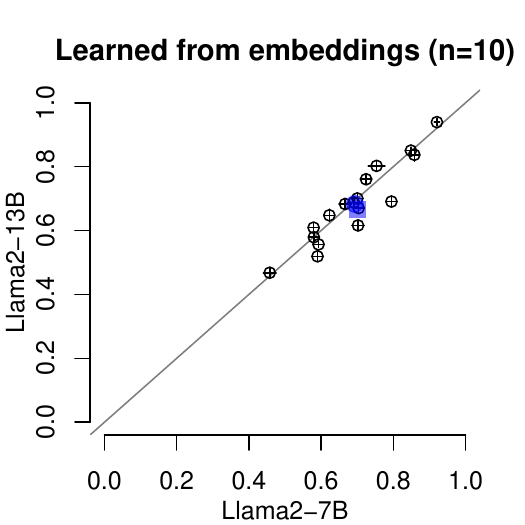}
\includegraphics[width=.325\linewidth]{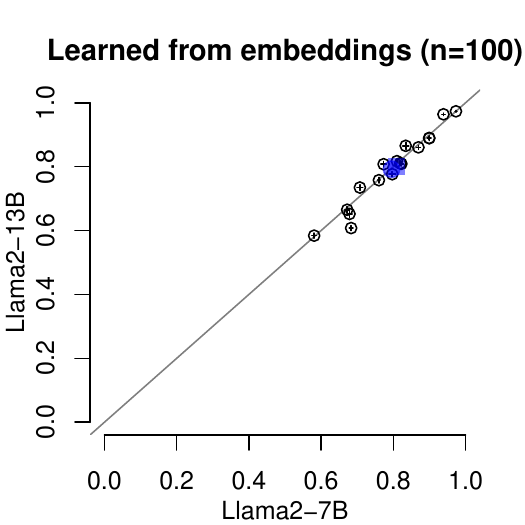}

\includegraphics[width=.325\linewidth]{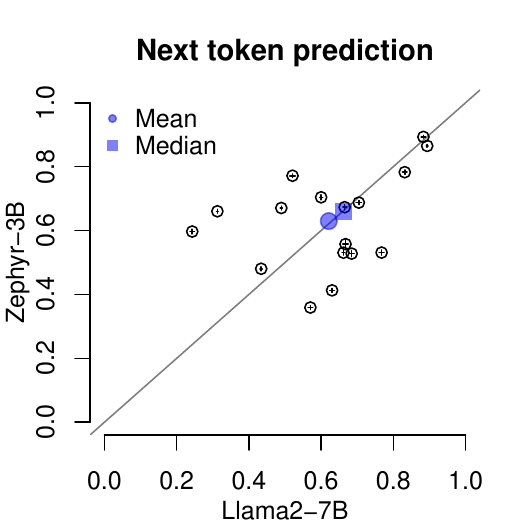}
\includegraphics[width=.325\linewidth]{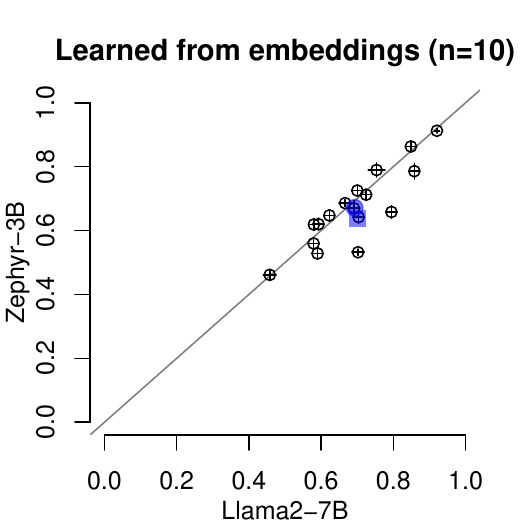}
\includegraphics[width=.325\linewidth]{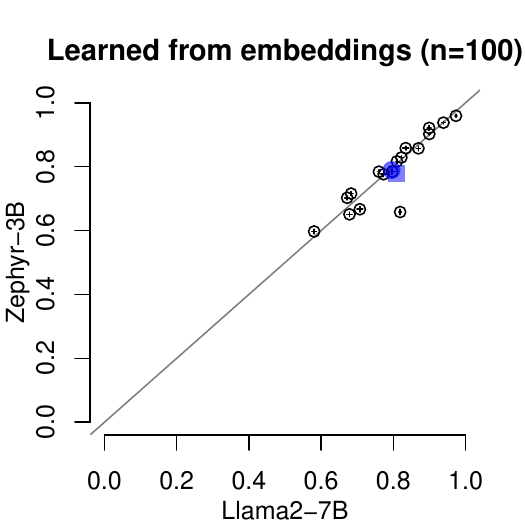}

\includegraphics[width=.325\linewidth]{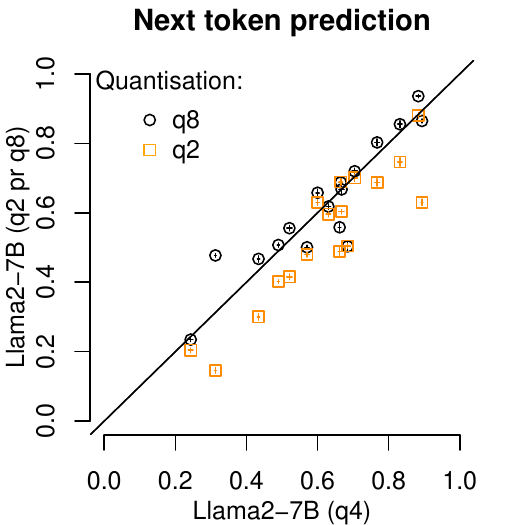}
\includegraphics[width=.325\linewidth]{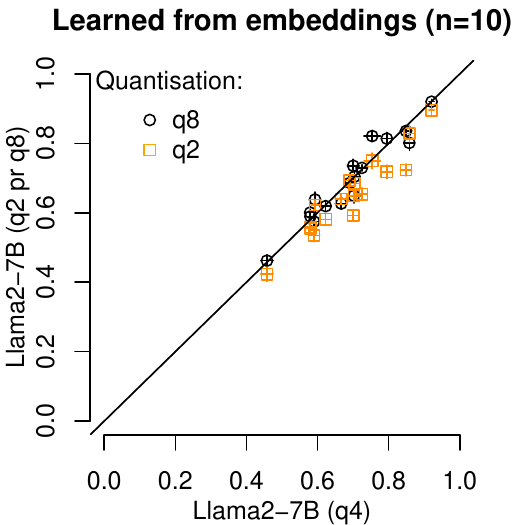}
\includegraphics[width=.325\linewidth]{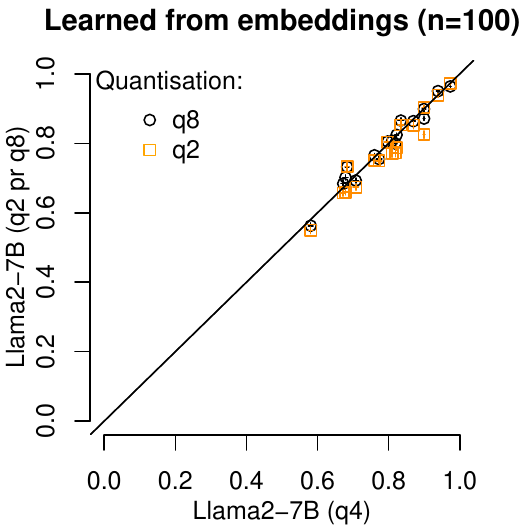}

\caption{\small The accuracy of PLR-E at different sample sizes when using the embeddings from the baseline model is compared against PLR-E using embeddings from other generative models (top two panels) or different quantisations of the baseline model (bottom panel).}
\label{fig_size_quant}
\end{figure}

\addtolength{\tabcolsep}{-2.5pt}    
\begin{table}[tb]
\begin{small}
\centering
\begin{tabular}{llllllllll}
  \hline
Datasets & \sc{GPT-4} & \multicolumn{2}{c}{\sc{Llama2 7B}} &\multicolumn{2}{c}{\sc{Llama2 13B}} &  \multicolumn{2}{c}{\sc{Zephyr 3B}} & \sc{Ada2} & \sc{BGE} \\ 
  &Token &  PRL-E &  Token &  PLR-E &  Token &  PLR-E  &  Token &  PLR-E & PLR-E \\ 

  \hline
Central banking & \underline{\textbf{0.67}} & 0.58 & 0.43 & 0.58 & 0.44 & 0.60 & 0.48 & 0.55 & 0.52 \\ 
  Clickbait & 0.72 & \underline{\textbf{0.97}} & 0.68 & \textbf{0.97} & 0.76 & 0.96 & 0.53 & 0.96 & 0.94 \\ 
  Headlines & 0.73 & \underline{\textbf{0.87}} & 0.77 & 0.86 & 0.63 & 0.86 & 0.53 & 0.84 & 0.80 \\ 
  Spam & \underline{0.92} & 0.90 & 0.67 & 0.89 & 0.64 & 0.92 & 0.56 & \textbf{0.93} & 0.91 \\ 
  Financial phrases & \underline{\textbf{0.82}} & 0.81 & 0.31 & \textbf{0.82} & 0.63 & \textbf{0.82} & 0.66 & 0.67 & 0.66 \\ 
  Weather (Tweets) & \underline{\textbf{0.84}} & 0.82 & 0.52 & 0.81 & 0.68 & 0.83 & 0.77 & 0.78 & 0.75 \\ 
  Irony (Tweets) & 0.73 & \underline{\textbf{0.82}} & 0.66 & 0.81 & 0.56 & 0.66 & 0.53 & 0.59 & 0.59 \\ 
  Emotions (Tweets) & \underline{\textbf{0.79}} & 0.76 & 0.49 & 0.76 & 0.70 & 0.78 & 0.67 & 0.65 & 0.63 \\ 
  Offensive (Tweets) & \underline{\textbf{0.70}} & 0.67 & 0.60 & 0.67 & 0.63 & \textbf{0.70} & \textbf{0.70} & 0.65 & 0.66 \\ 
  Hate (Tweets) & \underline{0.69} & 0.68 & 0.67 & 0.61 & 0.65 & \textbf{0.72} & 0.67 & 0.69 & 0.67 \\ 
  Stance (Feminism) & \underline{\textbf{0.75}} & 0.68 & 0.57 & 0.65 & 0.55 & 0.65 & 0.36 & 0.58 & 0.57 \\ 
  Stance (Aborition) & 0.67 & \underline{0.71} & 0.63 & \textbf{0.73} & 0.57 & 0.67 & 0.41 & 0.66 & 0.65 \\ 
  Stance (Atheism) & 0.47 & \underline{0.77} & 0.24 & \textbf{0.81} & 0.28 & 0.78 & 0.60 & 0.70 & 0.71 \\ 
  Movie reviews & \underline{\textbf{0.92}} & 0.90 & 0.89 & 0.89 & 0.87 & 0.90 & 0.87 & 0.82 & 0.75 \\ 
  Legal (Money) & 0.77 & \underline{0.80} & 0.70 & 0.78 & 0.73 & 0.78 & 0.69 & \textbf{0.86} & 0.84 \\ 
  Legal (Work) & \underline{0.95} & 0.94 & 0.88 & 0.96 & \textbf{0.97} & 0.94 & 0.89 & 0.93 & 0.92 \\ 
  Legal (Crime) & \underline{\textbf{0.90}} & 0.83 & 0.83 & 0.86 & 0.86 & 0.86 & 0.78 & 0.84 & 0.81 \\ \hline
  Mean & 0.77 & \underline{\textbf{0.80}} & 0.62 & 0.79 & 0.66 & 0.79 & 0.63 & 0.75 & 0.73 \\ 
  Median & 0.75 & \underline{\textbf{0.81}} & 0.66 & \textbf{0.81} & 0.64 & 0.78 & 0.66 & 0.70 & 0.71 \\ 
   \hline
\end{tabular}
\caption{\small Comparison of accuracy of different models. PLR-E methods are trained on 100 samples.}
\label{tab_perf_acc}
\end{small}
\end{table}
\addtolength{\tabcolsep}{2.5pt}    

\subsection{Robustness: Model size and quantization}  \label{sec:rob_mod_size_quant}

We test whether our results hold for different generative language models. Specifically, we compare the larger Llama2 13B-chat (q4.0) and the smaller, but more recent, Stable LM Zephyr 3B (q5.0) to our baseline model. The results are shown in the top two rows of Figure \ref{fig_size_quant}. Considering next token prediction, Llama2 13B outperforms the baseline model in most data sets. However, when evaluating the performance of PLR-E, the larger model is not superior and the performance difference decreases with increasing sample size. Comparing Zephyr 3B and Llama2 7B, we do not see a clear winner in next token prediction but also observe a decrease in the performance difference when using PLR-E. 

Table \ref{tab_perf_acc} shows the performance of PLR-E and next token prediction on the different models and compares it to the next token prediction of GPT-4. If our baseline PLR-E (Llama2 7B) beats GPT-4 the accuracy score is underlined and vice versa. The best performing model on a data set model is highlighted in bold. Table \ref{tab_perf_f1} in the appendix replicates this table using the macro F1 score as a performance metric.

Our baseline model uses 4-bit quantisation, next we test whether our results hold for 2-bit and 8-bit quantisations on our key data sets. Figure \ref{fig_size_quant} (bottom panel) shows that for the majority of the classification tasks we observe a substantial improvement in the zero-shot next token predictive accuracy when using the 8-bit model and a corresponding decline in performance when using the 2-bit model. However, when using PLR-E the performance differences between models are less pronounced, particularly at larger sample sizes. Note that the degree of quantisation influences processing speed as well as the memory footprint: going from the 4 to the 8 bit model also increases the average token generation time by 30\%.

In summary, having compared models of different lineages, sizes, and degrees of quantisation, we conclude that while these differences affect the accuracy of next token predictions they make relatively little difference to the performance of the PLR-E method.

\subsection{Robustness: In-context few-shot learning}

\begin{figure}[!h]

\includegraphics[width=.325\linewidth]{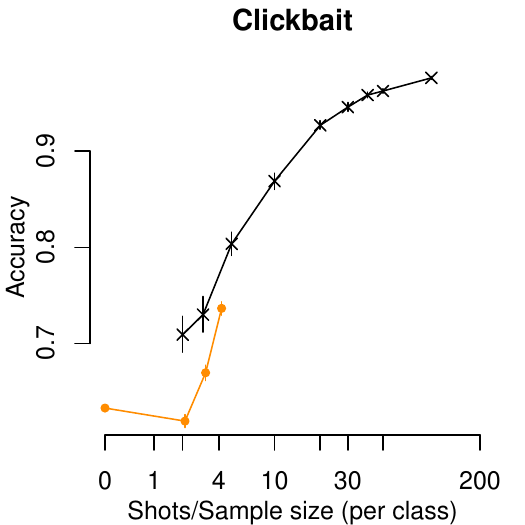}
\includegraphics[width=.325\linewidth]{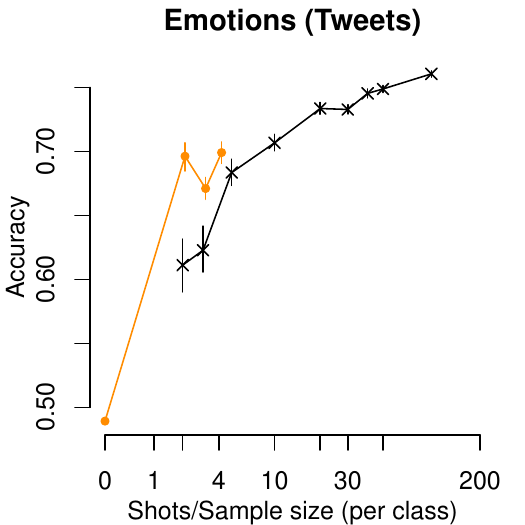}
\includegraphics[width=.325\linewidth]{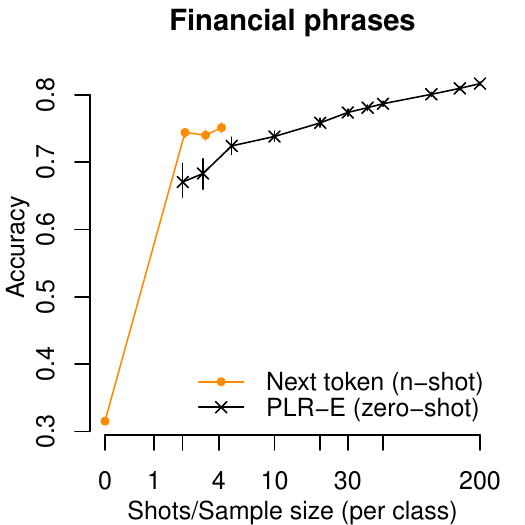}
\includegraphics[width=.325\linewidth]{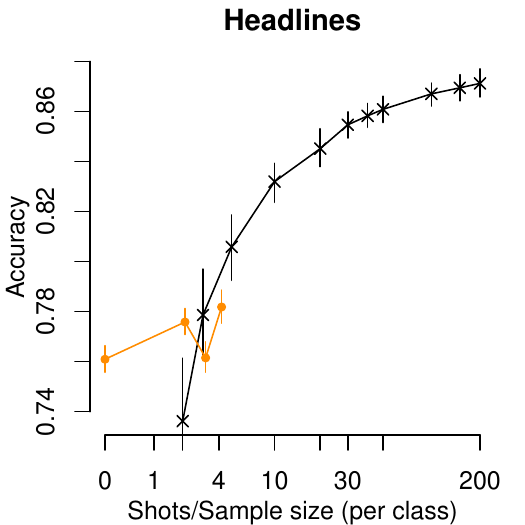}
\includegraphics[width=.325\linewidth]{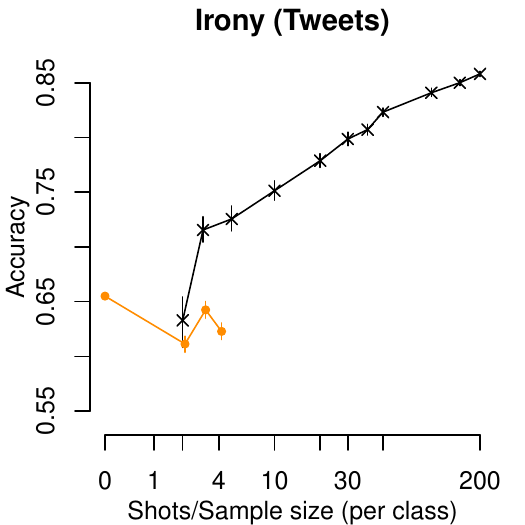}
\includegraphics[width=.325\linewidth]{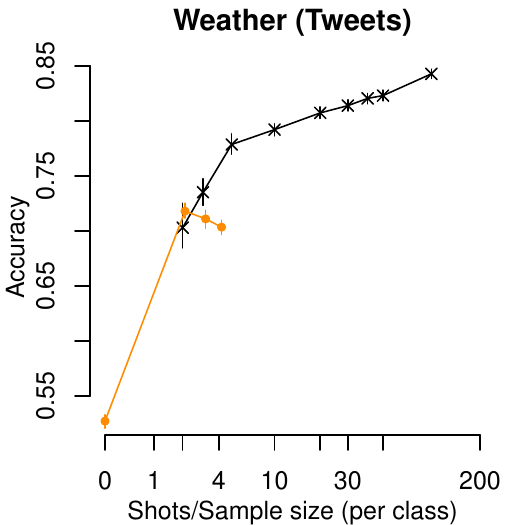}

\caption{\small Comparison of zero-shot and few-shot next token prediction of the baseline model and the PLR-E (zero-shot prompts) when calibrated on the same number of examples. }
\label{fig_learning_few_shot}

\end{figure}

In-context few-shot learning \citep{brown2020language} works by showing the generative model examples of prompts and responses. 
We test this calibration approach approach using $m \in \{2,3,5\}$ shots \textit{per class}} and compare its performance against PLR trained on exactly the same number of zero-shot text embeddings per class\footnote{Note that this differs from our other analyses, where we randomly sample observations across classes.}. Due to the high computational cost of few-shot learning, we only conduct the experiments on six of our data sets. Figure \ref{fig_learning_few_shot} shows that few-shot learning  leads to substantial performance improvements over the zero shot case in four of the six data sets (orange line). Comparing few shot learning against PLR-E zero-shot (black curve), we only observe  a consistently better performance of few-shot learning in two data sets. In both datasets, the performance of few shot learning seems to have saturated at five shots, suggesting PLR-E will perform better with more training samples. 

In Figure \ref{fig_learning_few_shot_regression} in the appendix we combine the in-context and PLR-E approaches by calibrating PLR-E on the embeddings produced from few-shot prompts. While few-shot prompts rarely lead to better performance at any point along the learning curves, they introduce a substantial computational overhead at inference time, since in a $k-$class, $m$-shot-per-class case, the prompt will typically be $km$ times longer, making tens-of-shot cases very costly. 

\section{Performance comparison: ``tens-of-shot" labelled data -- small training and test sets} \label{sec:uncertainty}

\begin{figure}[!ht]
\centering
\includegraphics[width=.95\linewidth]{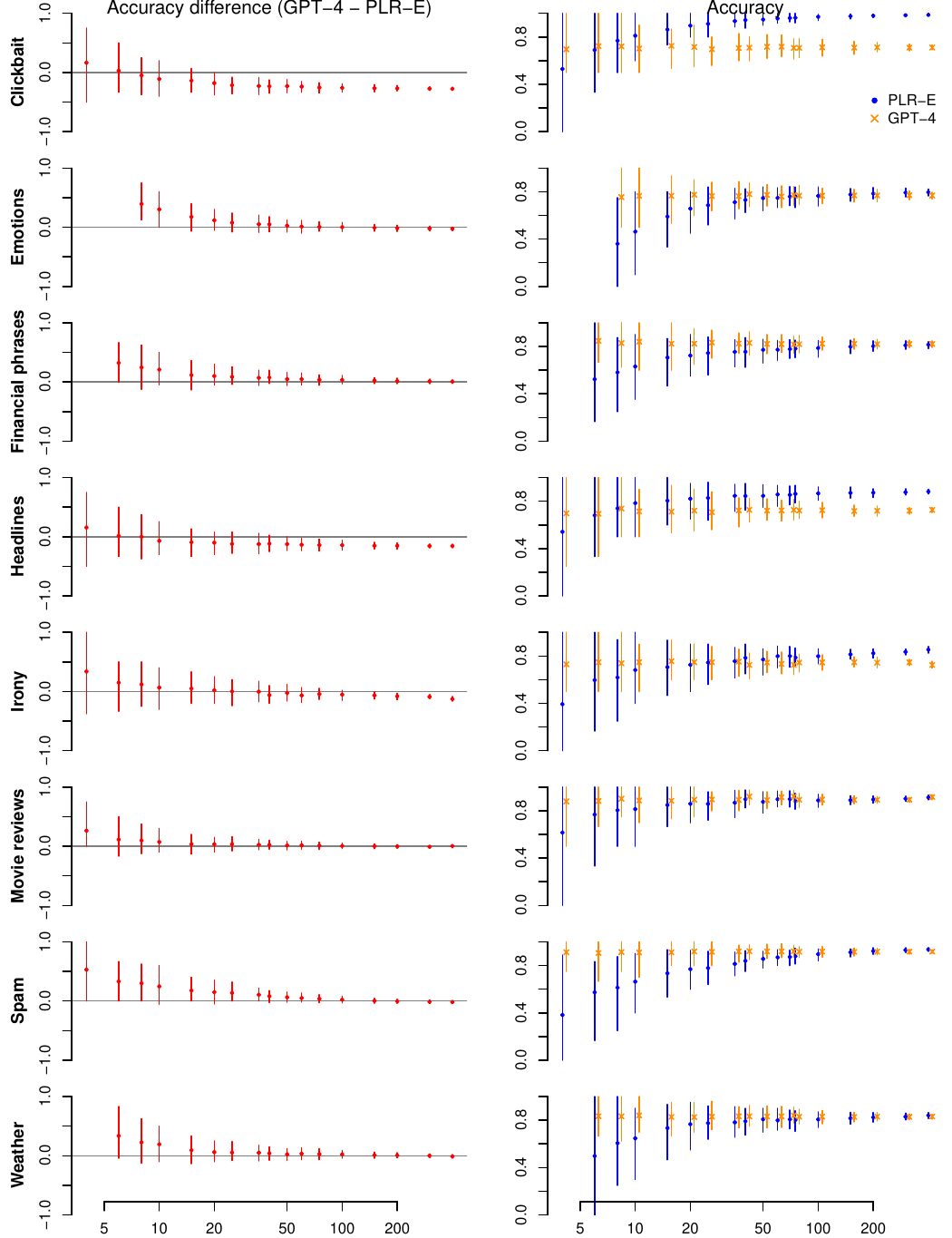}

\caption{\small Uncertainty of accuracy estimates for GPT-4 and PLR-E (right panel) and their difference (left panel) as a function of the training set size. We show the 5$^{th}$ to 95$^{th}$ percentile range based on 250 randomly drawn samples. }
\label{fig_uncertainty}
\end{figure}

The previous section showed that the PLR-E method can equal or better the performance of a flagship language model (GPT-4) in the ``tens-of-shot" limit, whereas the zero-shot next-token performance of the baseline model is, on average, significantly worse than GPT-4's zero-shot performance.

For ad hoc analysis, simply using GPT-4 will usually be an accurate and low-effort approach. But in any setting where the performance of the model has some value attached to it, the user needs a number of labelled instances to show empirically that their chosen model is indeed a good classifier.

In this section we show that the number of labelled instances needed as a test set to validate the performance GPT-4 (or any other zero-shot model) is large enough that, if those instances are used instead to train and test the PLR-E method, the methods are then in ``competition" (where we define competition as there being no statistically significant difference in their performance). We then show that by the time we have provided enough labelled instances to resolve that competition, that the number of instances is large enough that the PLR-E method equals or betters GPT-4's performance in all but one data set.

\subsection{Validating zero-shot performance}

Even if no actual competitor model exists, there is a natural limiting classification behaviour against which we can evaluate a model: the random baseline accuracy $a_r$, which is $0.5$ in a balanced, 2-class case. Depending on the accuracy of our model, we need some number of labelled data points to statistically show that the model performs better than random. 
To illustrate this, we model the true accuracy of the classifier $\hat{a}$ as a binomial variable with an unknown success probability $\hat{a}$. Our point estimate for $\hat{a}$ is the observed accuracy in the labelled sample. By estimating the binomial proportion confidence interval using the Wilson score method with continuity correction \citep{newcombe1998} we estimate how many labelled observations are needed so that the confidence interval of the estimate $\hat{a}$ does not overlap with the random classifier. For example, at a sample size 10, we need to observe an accuracy of at least 0.8 to reject the hypothesis that the classifier is not better than random (at $\alpha = 0.1$). At a sample size of 20, the minimum accuracy required to reject the null hypothesis is 0.67. 

In practice, the user might have some further operational or regulatory reasons for imposing a tight bound on the maximum uncertainty around the observed model performance, which requires a sizeable set of labelled instances. For example, with 25, 50, 100, and 250 labelled data points, the width of the 95\% confidence interval around an observed accuracy of 0.75 is 0.36, 0.25, 0.18, and 0.11, respectively. 

In this scenario we only need to estimate $\hat{a}$ and its uncertainty but if we compare two classifiers the sample size requirements increase due to two unknown parameters. Additionally, one classifier being trained on the sample introduces a further, and not estimable, variance in its accuracy across samples: In the following, we therefore measure the empirically observed uncertainty of the point estimates of the performance of GPT-4 and PLR-E by repeated sampling.

\subsection{Statistical comparison of baseline model and GPT-4}

For a given training sample of size $n$, we use  $k$-fold cross-validation to obtain a point estimate of the performance of PLR-E, where $k = \min(20, n)$. For GPT-4, we just measure the accuracy across all $n$ observations. To estimate the variance of these point estimates we replicate this procedure on 250 randomly drawn training sets. We reject training samples that do not have at least two data points of each class in the sample, ensuring at least one observation per class in each training set.

The left-hand side of Figure \ref{fig_uncertainty} shows the mean and the 5$^{th}$ and 95$^{th}$ percentiles of the difference in accuracy of GPT-4 and PLR-E. The right-hand side shows the same statistics for the levels of accuracy of the two models. The maximum sample size tested is 400. While GPT-4 generally performs better in the mean than PLR-E at the smallest sample size, that mean cannot be observed in practice. The 5$^{th}$ quantile of the difference in performance is lower or equal to zero in all data sets, telling us that we cannot reliably state in practice that GPT-4 outperforms PLR-E.\footnote{Note that at small sample sizes the uncertainty interval around the PLR-E performance estimate is generally larger than that of GPT-4 due to the additional uncertainty created by the model estimation.} With increasing sample size, the uncertainty interval around the performance difference shrinks but so does the GPT-4's advantage. GPT-4 is only the ``statistical winner" ( i.e.\ the uncertainty interval of the performance difference is strictly positive at any point of the learning curve) in three of the 17 data sets. The opposite -- PLR-E being the statistical winner over GPT-4 -- can be observed in seven data sets by the end of the learning curve. 

\section{Analysis of embedding space} \label{sec:embeddings}

In this section, we aim to better understand the embedding vectors and their usefulness for prediction. To simplify this analysis, we avoid multinomial models and transform all multi-class prediction problems into a binary prediction problem by training PLR-E to differentiate the majority class from all other classes.

The high performance of the regression, albeit penalised, on a 4096-dimensional space with tens of training points, is surprising. 

Consider a 2-class data set with $n$ observations per class. With non-duplicated values on each dimension, there are $(2n)!$ possible permutations of values of which $ 2(n!)^2$ separate the classes linearly. Thus, the expected number of embedding dimensions in which the classes separate linearly by chance is $4096\times 2(n!)^2/(2n)!$ which is above $1$ for $n\leq 7$. Therefore, for small training samples, the chance of the model focussing on arbitrary, non-predictive dimensions is high. The main reason for the strong performance of PLR-E on small sample size is the high correlation between the embedding vectors, which implies that the model does not need to find a sparse solution -- a difficult endeavour in the high dimensional space.

We show the high degree of collinearity of the embedding space using a principal component analysis (PCA). Figure \ref{pca_explained} depicts the cumulative explained variance when using between 1 and 100 components. With the contextualising prefix and suffix (left panel), the first component explains 18--56 (median = 23) of the variance  and the first ten components explain 58--79 (median 63). Without prefix and suffix, the explained variance decreases substantially to 8 and 29 (median = 10) percent and 24 and 50 (median = 39) percent, when using the first and the first 10 components, respectively. 

\begin{figure}[!ht]
\centering
\includegraphics[width=.8\linewidth]{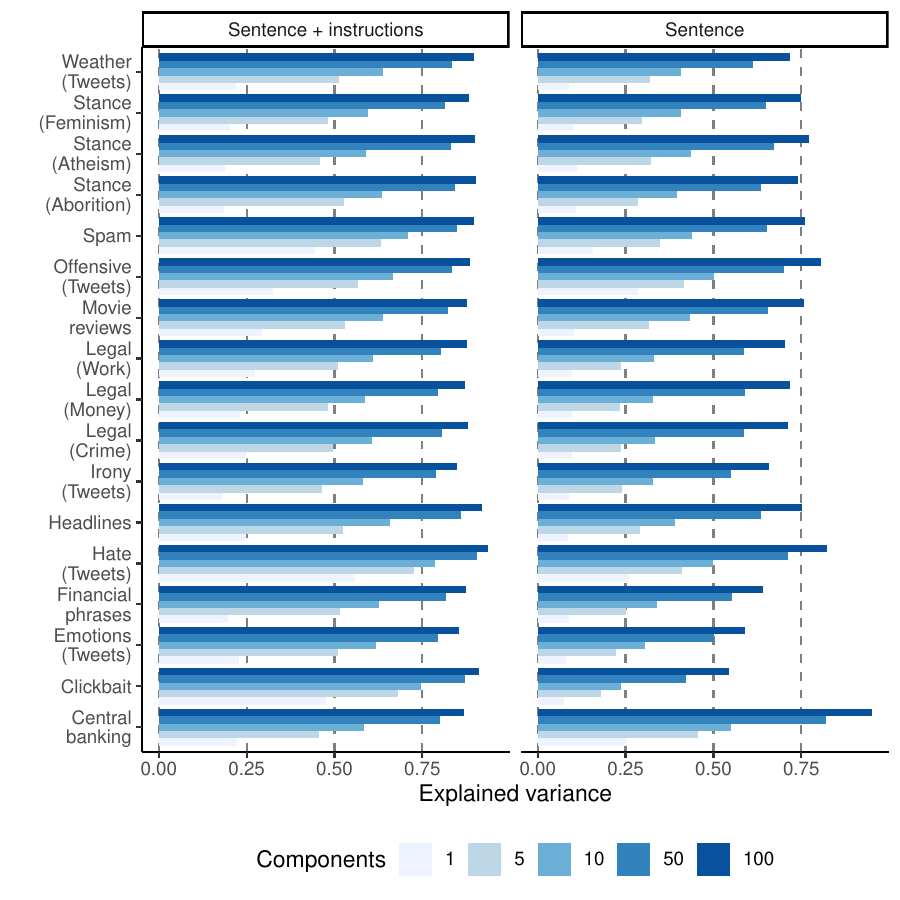}
\caption{\small The cumulative explained variance of PCAs on the embeddings produced by our baseline model with (left panel), and without (right panel), a surrounding prompt.}
\label{pca_explained}
\end{figure}

Figure \ref{fig_pca_learning} shows the predictive accuracy as a function of the number of principle components for a given sample size (colour and symbol). The PCA is fitted to all data points rather than only to a smaller, labelled, training set. Furthermore, we do not normalise the components before we train the ridge regression. In this way, the regularisation implicitly penalises those components more that explain less variance, as these tend to have a higher coefficients. For small training samples, a low dimension PCA can equal (or even slightly exceed) the performance of the baseline approach using all dimensions (shown as dotted horizontal lines). When increasing the training sample, we often observe that the maximum performance is reached when using 30--50 components. 

In some datasets we find that the performance deteriorates with a large number of components. This is much more pronounced, with an earlier onset, when we normalise the PCA components to have unit variance (see Figure \ref{fig_pca_learning_scaled} in the appendix). This is expected since the linear model will give weight to spurious correlations in the training data by the permutation-based argument at the beginning of this section. In contrast to the collinear embedding matrix, most of the orthogonal PCA components explain only little variance in the embeddings and have a small correlation with the class labels. Therefore, `accidentally' giving them weight in the regression model can have a strong negative impact on performance.

\begin{figure}[!ht]

\includegraphics[width=\ww\linewidth]{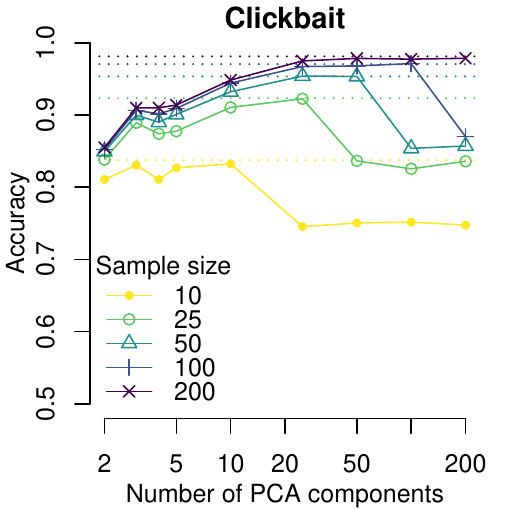}
\includegraphics[width=\ww\linewidth]{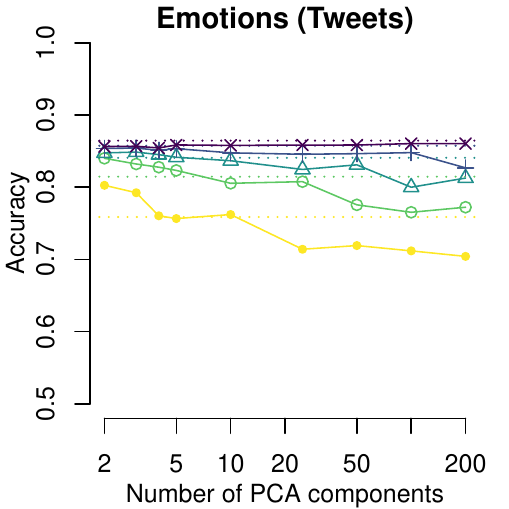}
\includegraphics[width=\ww\linewidth]{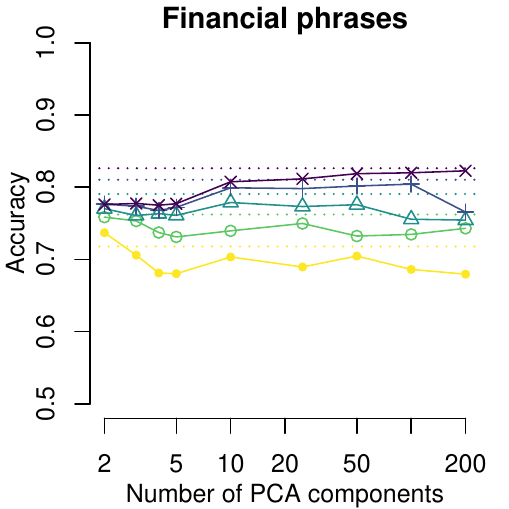}
\includegraphics[width=\ww\linewidth]{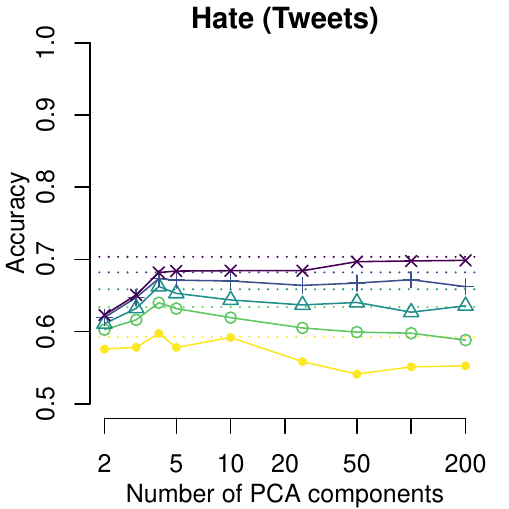}
\includegraphics[width=\ww\linewidth]{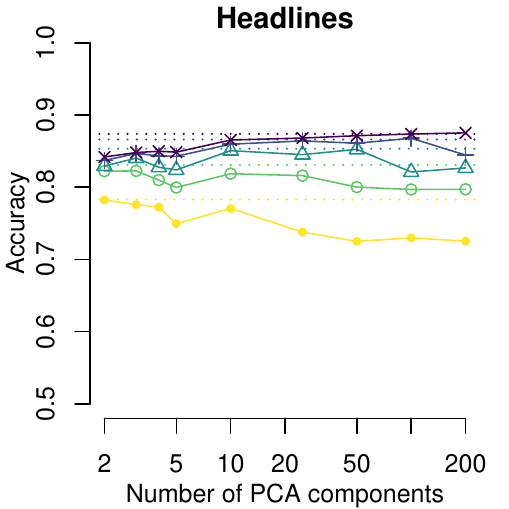}
\includegraphics[width=\ww\linewidth]{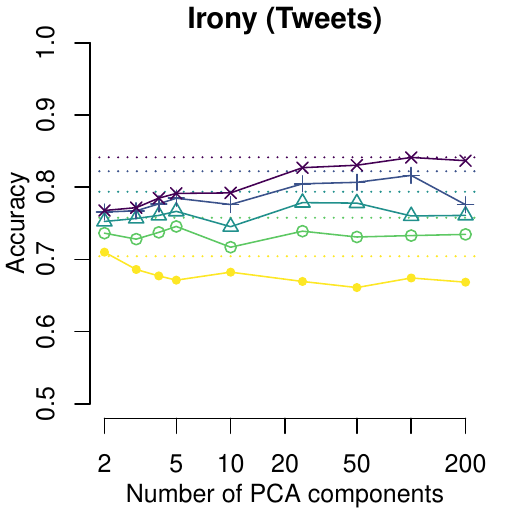}
\includegraphics[width=\ww\linewidth]{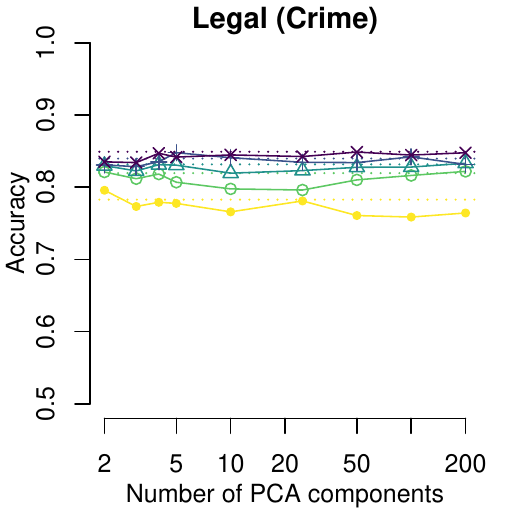}
\includegraphics[width=\ww\linewidth]{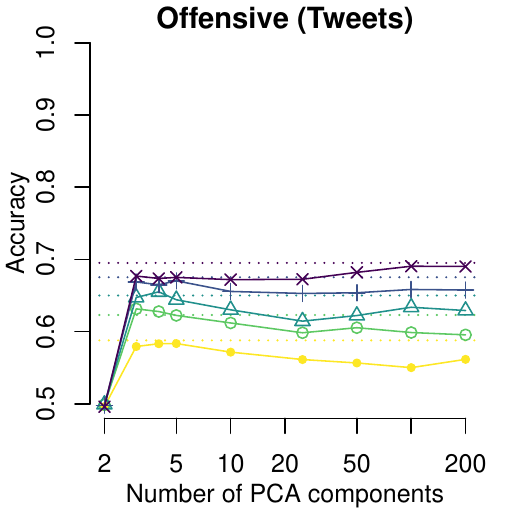}
\includegraphics[width=\ww\linewidth]{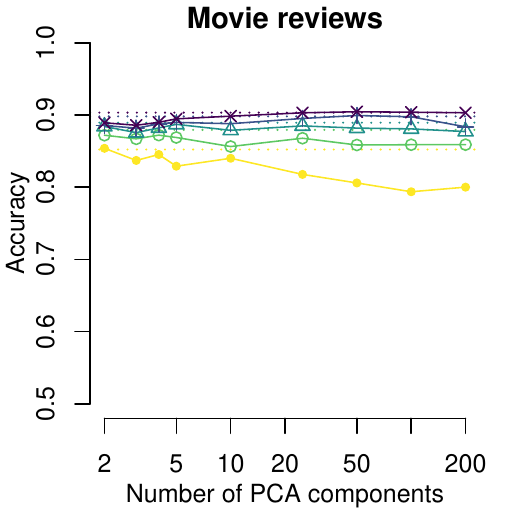}
\includegraphics[width=\ww\linewidth]{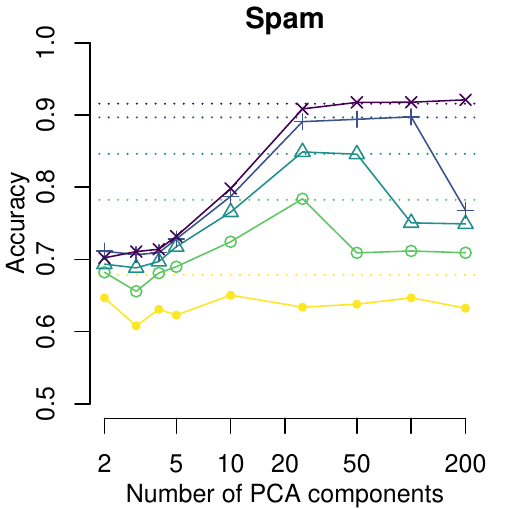} 
\includegraphics[width=\ww\linewidth]{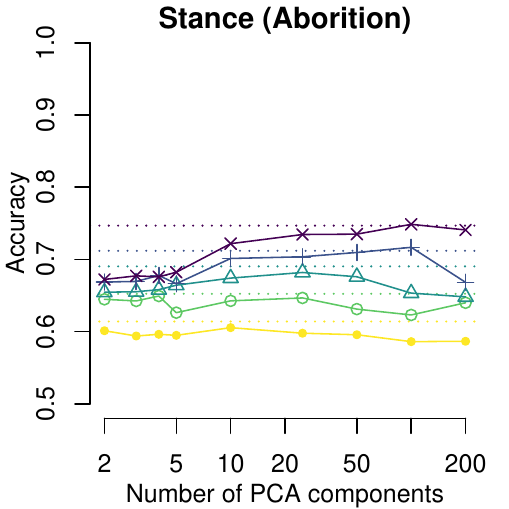}
\includegraphics[width=\ww\linewidth]{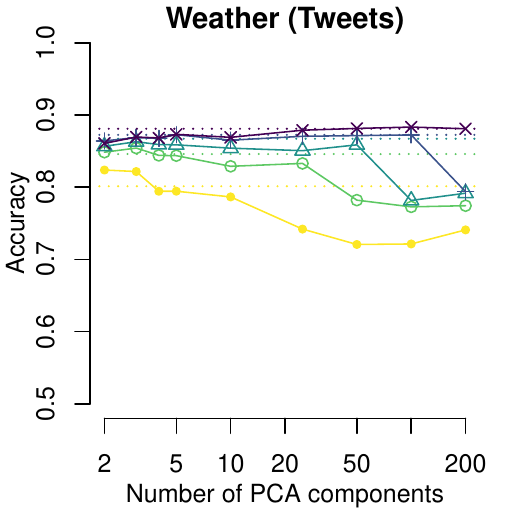}

\caption{\small Accuracy as a function of the number of (unnormalised) principle components for a given sample size (colour and symbol).}
\label{fig_pca_learning}
\end{figure}

We also test whether we can use sparse models directly on the embedding space. Specifically, we train a Lasso regression to select at most $n$ features with non-zero coefficients and then train an unregularised regression model on that subset of features to undo the shrinkage of the parameters due to the $L_1$ penalty \citep[see][]{hastie2017extended}.

Figure \ref{sparse_learning} shows the accuracy of the Lasso regression as a function of $n$ for different sample sizes (colour and symbol). The performance of the baseline ridge model is shown as dotted horizontal lines.

When the training sample is small, lasso generally falls behind the ridge baseline model. But on larger sample sizes, the Lasso regression's performance often approaches or equals that of the ridge regression Surprisingly, this is often achieved with very sparse models that use less than 5 of the 4096 embedding dimensions for prediction. In line with our other results, sparse models are less accurate when we remove the context from the prompt (see Figure \ref{sparse_learning_noq} in the appendix). 

\begin{figure}[!ht]

\includegraphics[width=\ww\linewidth]{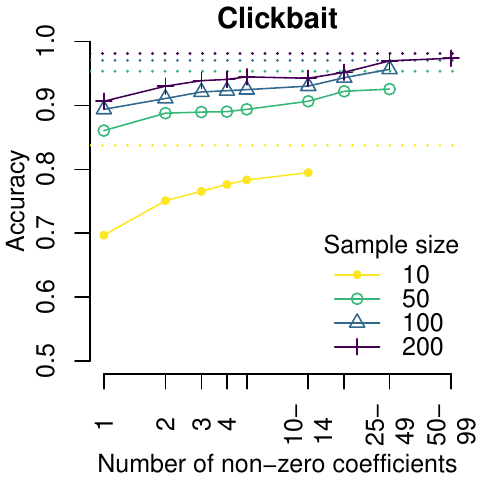}
\includegraphics[width=\ww\linewidth]{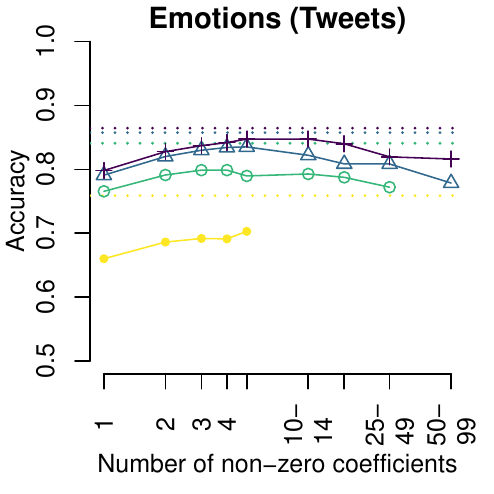}
\includegraphics[width=\ww\linewidth]{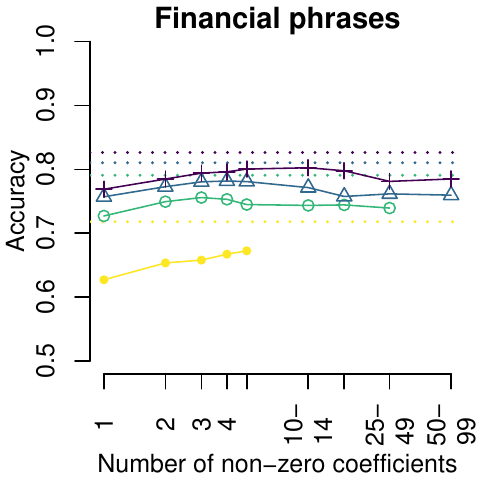}
\includegraphics[width=\ww\linewidth]{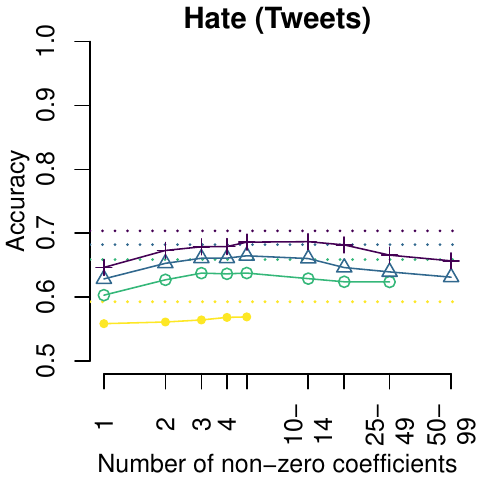}
\includegraphics[width=\ww\linewidth]{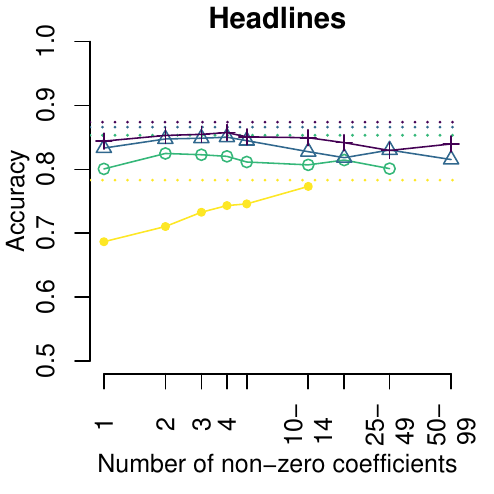}
\includegraphics[width=\ww\linewidth]{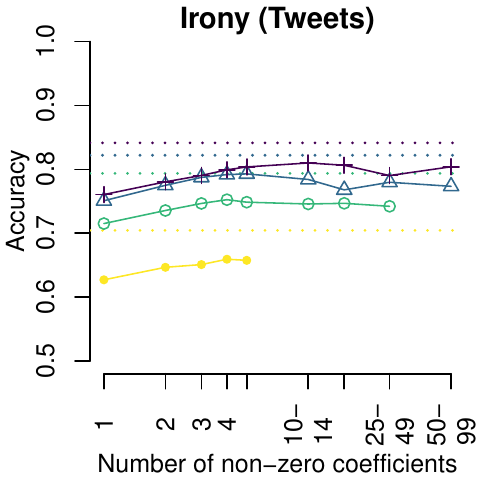}
\includegraphics[width=\ww\linewidth]{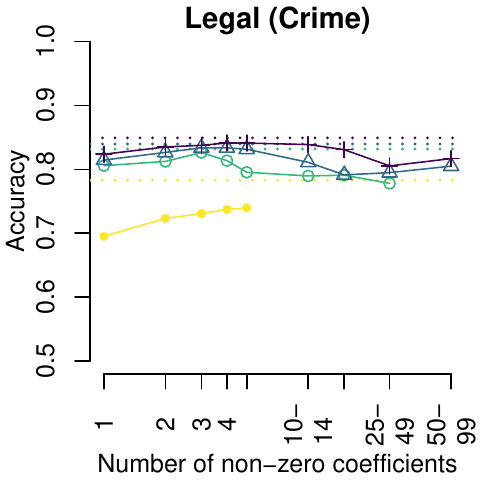}
\includegraphics[width=\ww\linewidth]{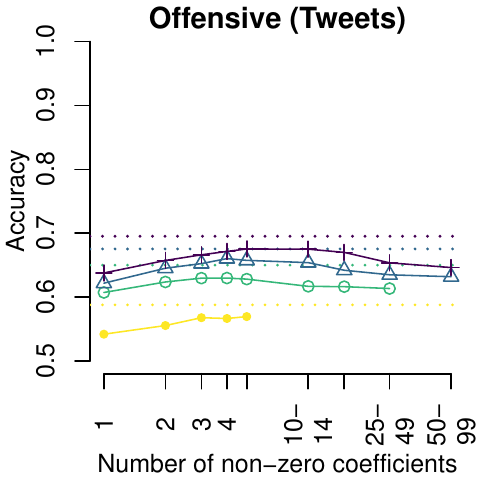}
\includegraphics[width=\ww\linewidth]{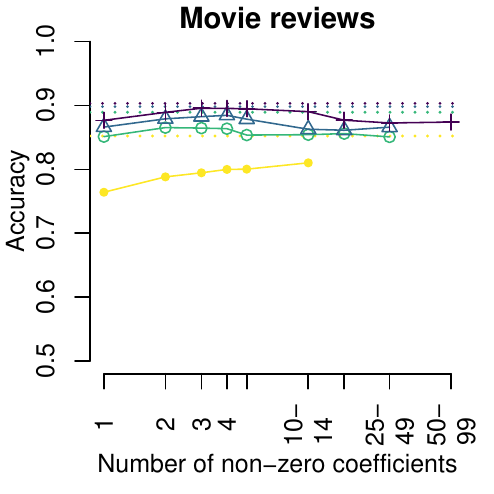}
\includegraphics[width=\ww\linewidth]{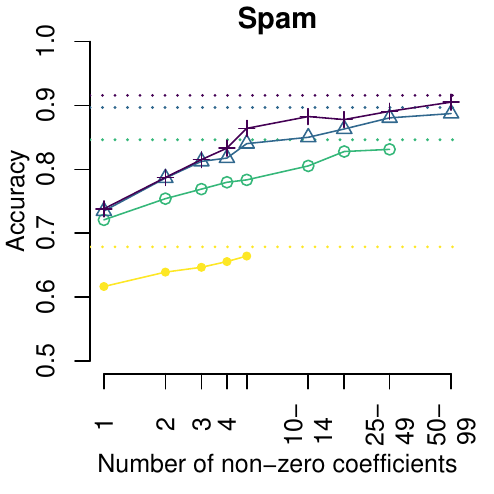} 
\includegraphics[width=\ww\linewidth]{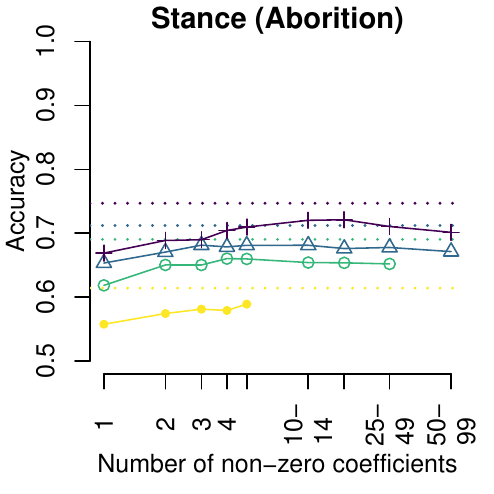}
\includegraphics[width=\ww\linewidth]{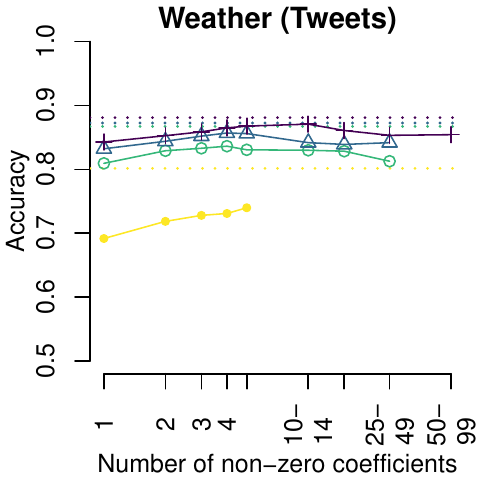}

\caption{\small The accuracy of the Lasso regression as a function of the number of non-zero coefficients for different sample sizes (colour and symbol).}
\label{sparse_learning}
\end{figure}

Finally, we test the regularisation paths which show how sensitive the performance of our PLR-E model is to the regularisation parameter $\lambda$ (Equation \ref{eq:plr}). In Figure \ref{regularisation}, each line corresponds to a different dataset and shows how the accuracy of PLR-E, trained on 10 instances (left panel) and 100 instances (right panel), changes with the regularisation parameter\footnote{From the 100 models on the  regularisation path, we omit the model with the highest degree of regularisation as it shrinks all coefficients to 0.}. At both sample sizes, the performance is insensitive to the regularisation parameter chosen and the lowest degree of regularisation (our baseline) produces the most accurate models.

\begin{figure}[!ht]
\centering
\includegraphics[width=.49\linewidth]{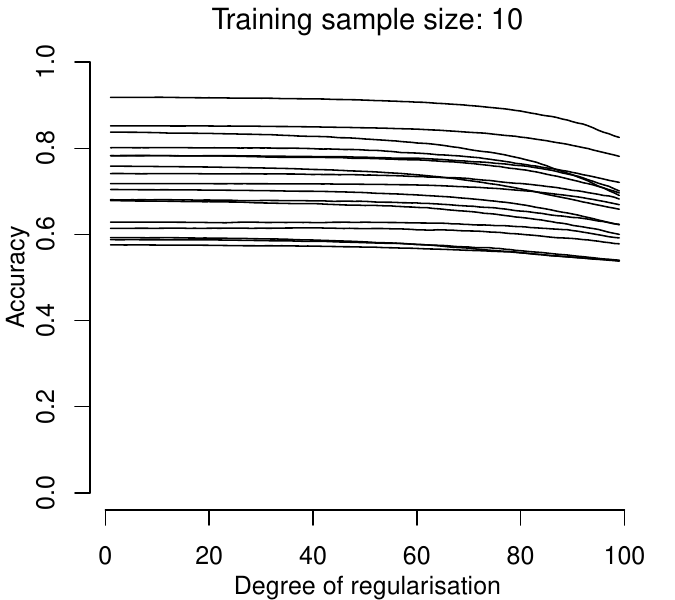}
\includegraphics[width=.49\linewidth]{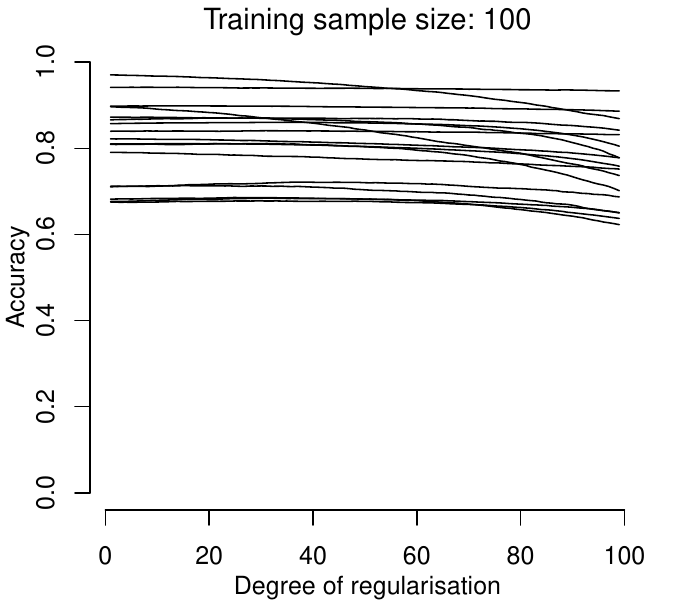}
\caption{\small The dependence of the accuracy of PLR-E, trained on 10 instances (left panel) and 100 instances (right panel), on the regularisation parameter.}
\label{regularisation}
\end{figure}

\section{Explainability} \label{sec:explain}

Explainability is important for our model. In a in commercial context, it is required for robust decision making, and for fulfilling legal and ethical responsibilities. And, given that our model works by classifying based on training data, we need to check that the model is not attaching importance to features unrelated to the task, for example grammatical or formatting differences between different classes \citep{du2023shortcut}.

Using a combination of de-dimensionalisation by unnormalised PCA and $L_2$ regularisation (as in Figure \ref{fig_pca_learning}), we can obtain models which have both high performance and produce explanations which are (a) stable between models created using training sets of the same size, and (b) converge quickly to the explanations given in the large training set limit. We focus on the Financial Phrases dataset and manually annotate a set of $30$ examples ($15$ positive and $15$ negative, listed in Appendix \ref{sentences}) with our perception of the positivity or negativity of parts of the example at the word level. We quantify the agreement between our annotations and the feature importances produced by the model to show that the model is attaching importance to the `correct' parts of the phrases.

The model used in this section uses 10-dimensional unnormalised PCA, and a regularisation constant $\lambda = 0.01$. With $30$ training instances (an average of 10-per-class) it has a test accuracy which equals GPT-4's. (In the large sample limit, the model then underperforms relative to the baseline model by $1\%$.) 

We take an example phrase, $e$, and tokenise it into words and symbols (i.e.\ not the tokenisation used in an LLM, but in the more conventional meaning). We define feature importance using the decision function ($\boldsymbol{a}.\boldsymbol{e}$ above). We calculate the feature importance of the k$^{th}$ token by deleting it and finding the decision function $d_m(k, e, c)$ for a given class, $c$ and model $m(t)$, a model with training data size $t$. Then, with $d_m(, e, c)$ being the decision function for the complete phrase, the feature importance of token $k$ is $f_m(k, e, c) = d_m(, e, c) - d_m(k, e, c)$. 

The values reported in the figures are found by first constructing all the feature importances for a given example and class $F_m(e,c) = \{f_m(k, e, c) | k \in 1, ..., |e|\}$ and then normalising the standard deviation of the members of $F_m(e,c)$ to $1$, obtaining $\tilde{F}_m(e,c)$.

\subsection{Stability and convergence}

We take the average of feature importances over $20$ independent models with large ($200$ instances) training set size as our baseline, $\tilde{F}^{\infty}(e,c)$. The error for a single model with a smaller training sample size $t$ can be decomposed as $\tilde{F}_{m(t)}(e,c) = [\tilde{F}_{m(t)}(e,c) - \tilde{F}^t(e,c)] + [\tilde{F}^t(e,c) - \tilde{F}^\infty(e,c)]$, where addition and subraction are element-wise in $k$. The first term is the shift in the feature importances for a particular model $m(t)$ trained on a training set of size $t$ versus the average feature importances for all models with a training set of size $t$, and the second term is the shift in that average relative to the average feature importance in the large training set limit. 

We report the mean absolute values of both the first and second term in Figure \ref{spread_shift_explain}. The first term then describes the spread amongst models for a given training set size and the second their shift relative to the average over models trained on many training instances ($\tilde{F}^{\infty}(e,c)$). Even at the smallest training sample size the deviations are acceptable for important features which empirically have sizes in $\tilde{F}$ greater than $1$. This can be seen in four random examples in Figure \ref{word_level_explain} (the results from 16 other examples are shown in Figures \ref{word_level_explain_appx_1}--\ref{word_level_explain_appx_3} in the appendix). The models are trained on $30$ training instances, which as noted, gives models with a test accuracy on average equal to GPT-4's. The spreads in the feature importances from models trained on different random $30$-instance training sets are shown by the black bars and are small in fractional terms, particularly for the important features. The shifts relative to the large training set limit, which are not represented in the figures, are around $3$ times smaller than the spreads.

\begin{figure}[!tb]
\centering
\includegraphics[width=0.85\linewidth]{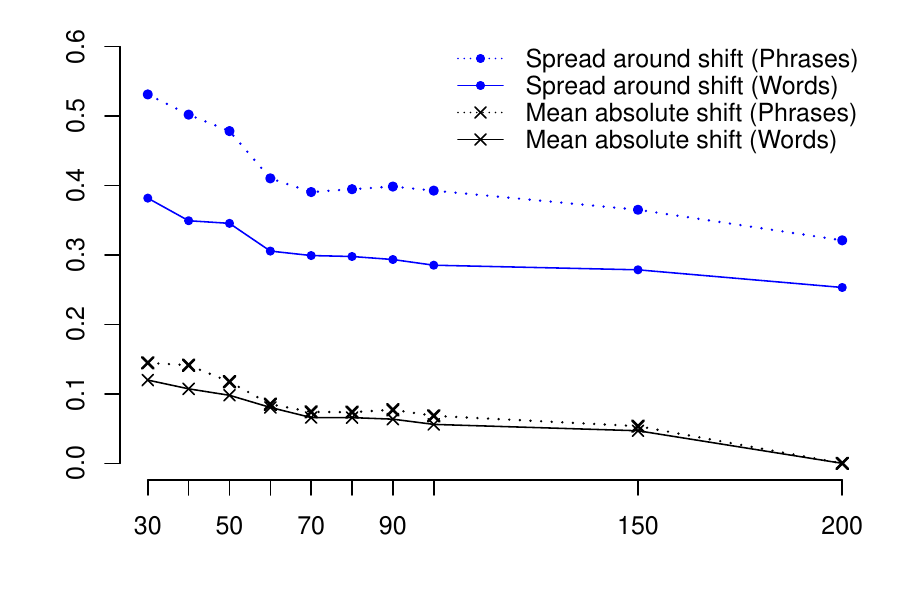}
\caption{\small The decomposition of the average absolute deviation from a ground truth explanation of a model trained on different sized training samples.}
\label{spread_shift_explain}
\end{figure}

\subsection{Correspondence with human annotations}

\def\captionexplain{The word-level feature importances according to the PLR-E model (black) are compared against annotated importances by the authors (blue, $+$ and $\times$ represent the two annotators).}

\begin{figure}[h!t]
\centering
\includegraphics[width=0.49\linewidth]{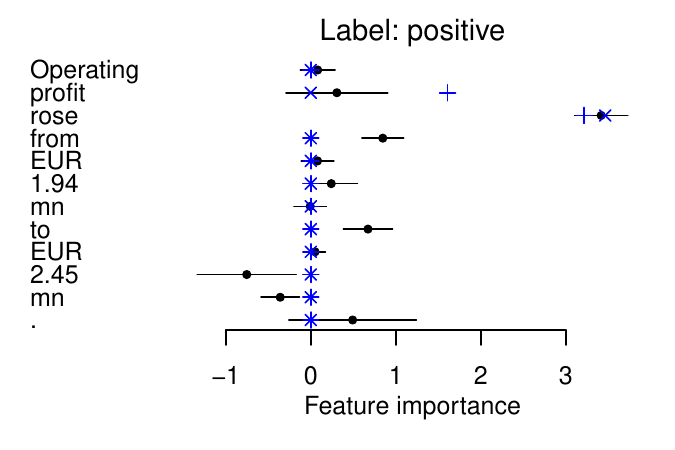}
\includegraphics[width=0.49\linewidth]{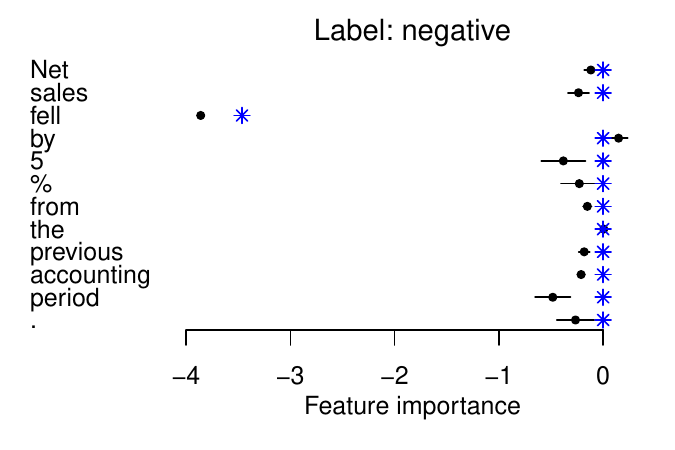}
\includegraphics[width=0.49\linewidth]{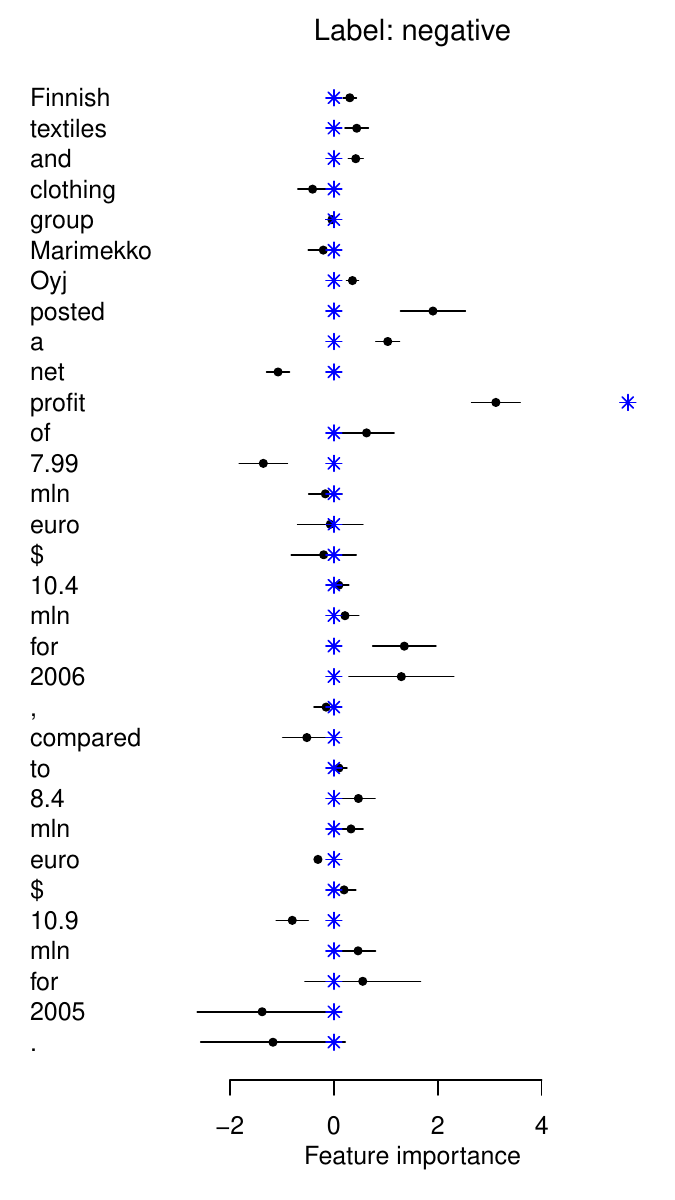}
\includegraphics[width=0.49\linewidth]{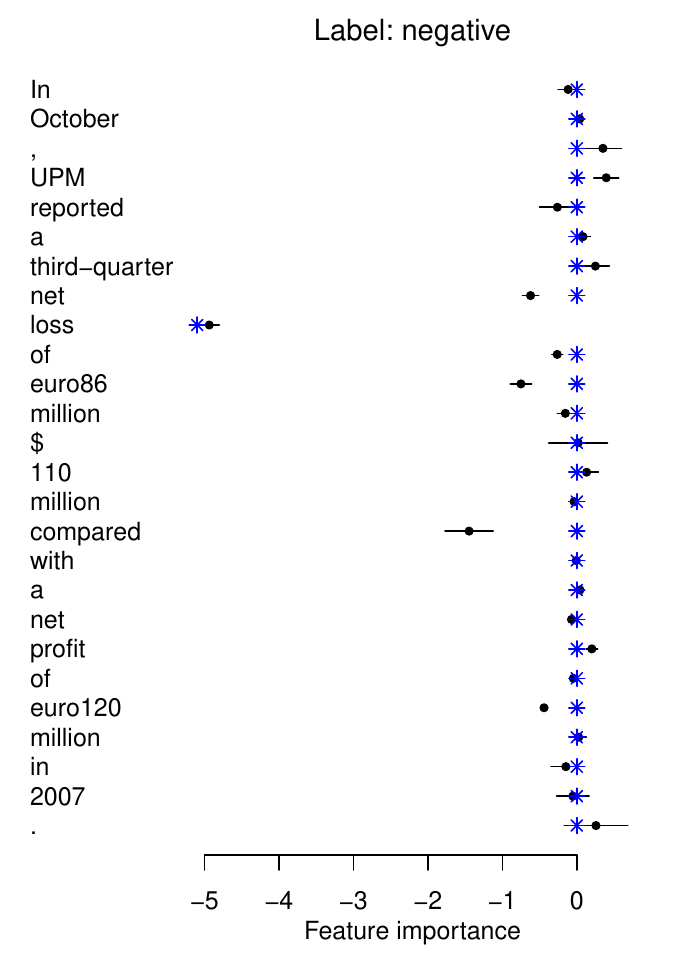}
\caption{\small \captionexplain}
\label{word_level_explain}
\end{figure}

The two authors independently labelled the positive and negative sentiments of individual words and phrases in the set of $30$ examples. 

Figure \ref{word_level_explain} shows the word-level results in four random examples (the results from 16 other examples are shown in Figures \ref{word_level_explain_appx_1}--\ref{word_level_explain_appx_3} in the appendix). Visually, it is clear that the model is attaching importance to the correct words (and indeed can outperform the human annotators, for example in attaching a negative sentiment to the word `compared' in the bottom right panel of Figure \ref{word_level_explain}. There is a scatter in the points around $0$ even for words to which the annotators attached no importance; sometimes there is a `leakage' effect where words within the same phrase as an important word have additional importance; and finally, there are some anomalies around the end of the example. All are likely due to the effect of removing words agrammatically from the phrase.

To investigate the effect of agrammatical word removal, we also implemented a simple algorithm based on a constituency parse produced by Stanza \citep{qi2020stanza} to produce grammatical (though not necessarily sensible) phrases by removing entire grammatical units rather than individual words. The mean absolute spreads are about $30\%$ larger and the shifts $10\%$ larger, but the scatter in the points decreases: quantitatively the Spearman rank correlation of the first annotator's feature importances with those from the model is $0.31$ for the word-level explanations and $0.66$ for phrase-level explanations. We do not discuss these results further here due to the difficultly in labelling and interpreting these phrase-level explanations, that is, determining a ground-truth as to which points should be non-zero. A number of the perturbations in each case are not sensible, or are sensible and have a sentiment but no longer relate to the original topic. This makes labelling very challenging and uncertain, but is a broader problem with explainability in language tasks not specific to our work.

As regards this work, we have shown that the PLR-E method can provide explanations which are stable with respect to model and training set size. In this case, at the cost of 10 labelled training instances per class, the model achieves the accuracy of a flagship model, but also benefits from stable and sensible explanations.

\section{Conclusion} \label{conclusion}

We have shown that a penalised regression on embeddings allows local, generative language models to achieve comparable performance to the flagship GPT-4 model in text classification problems. In fact, no more labelled instances are required than are needed for statistically validating the performance of GPT-4. A large number of experiments demonstrated the robustness of our results. An analysis of the embedding space reveals that a handful of the $4096$ embedding vectors often suffice to train an accurate linear model on the embeddings. In addition to general advantages of locally hosted models such as privacy, availability, low cost, we show that our approach enables stable and sensible explanations.

\bibliography{bib}

\appendix
\counterwithin{figure}{section}

\section*{Appendix}
\renewcommand{\thesection}{\Alph{section}}
\section{Datasets}\label{datasets}

\paragraph{Financial Phrases.}

The dataset contains 4,843 English financial news sentences categorised into positive, negative, and neutral sentiment by 16 annotators. The data set was assembled by \citet{Malo2014GoodDO} and can be downloaded from \href{https://huggingface.co/datasets/financial_phrasebank}{Huggingface}.  We sample of random subset of 4,838 of the sentences of this dataset.

\paragraph{Central Banking.}  % https://github.com/gtfintechlab/fomc-hawkish-dovish

This dataset contains 2,480 statements of the Federal Open Market Committee based on speeches, meeting minutes and press conferences. The statements have been labelled as having a hawkish, negative or a neutral monetary policy stance. The annotators were given a detailed guide that defined hawkishness and dovishness for eight different categories (economic status, dollar value change, energy/house prices, foreign nations. Fed expectations/actions/assets, money supply, keywords/phrases, labour), which illustrates the complexity of the classification problem. This dataset has been assembled by \citet{shah2023trillion} and is made available on Github \href{https://github.com/gtfintechlab/fomc-hawkish-dovish}{Github}. 

\paragraph{Clickbait.} This data set contains headlines of 32,000 articles. The prediction task is to predict whether a headline is \textit{clickbait} or not, where clickbait is defined as a catchy headline that lures readers to click on a link. This dataset has been assembled by \citet{chakraborty2016} and is made available from the one of the author's \href{https://github.com/bhargaviparanjape/clickbait}{Github} page. We sample of random subset of 800 of the sentences of this dataset, balancing the two classes.

\paragraph{Weather (Tweets).}

The dataset contains 1000 tweets that were assigned to the labels \textit{positive sentiment, negative sentiment, neutral sentiment, I can't tell, tweet not related to weather} by 20 human annotators. The data can be downloaded from \href{https://data.world/crowdflower/weather-sentiment}{data.world}. We removed the 238 tweets in the categories \textit{I can't tell, tweet not related to weather} from the data set.

\paragraph{Headlines.}

The dataset contains 10,570 news headlines on gold in its role as a commodity. Human annotators have annotated the price sentiment (\textit{positive, negative, neural, none}) and six boolean variables reflecting the content of the tweet: \textit{price direction up, price direction constant, price direction down, asset comparison, past information, future information}.
The dataset has been assembled by \citet{sinha2021impact} and can be downloaded from \href{https://www.kaggle.com/datasets/ankurzing/sentiment-analysis-in-commodity-market-gold}{Kaggle}. We use the mutally exclusive variables \textit{past information} and \textit{future information} as our binary classes. We randomly sample a subset of 639 headlines, balancing the two classes.

\paragraph{Tweet Eval: Emotions, Irony, Hate, Offensive, and Stance on feminism, atheism and abortion.}

Tweet Eval \citep{barbieri2020tweeteval} is the \href{https://github.com/cardiffnlp/tweeteval?tab=readme-ov-file}{repository} of a benchmark study and  contains seven heterogeneous tweet classification tasks from which we used five in our study, all of which can be downloaded from \href{https://github.com/cardiffnlp/tweeteval/tree/main/datasets}{Github}.

\paragraph{Emotions.} 
The dataset contains 5,052 tweets, each expressing one of four emotions: \textit{anger, joy, sadness, optimism}. The data was assembled by \citet{mohammad2018semeval}. We randomly sample 4,653 tweets from the dataset.

\paragraph{Irony.}
The dataset contains 4,601 tweets which are either ironic or not. The dataset was assembled by \citet{van2018semeval}. We randomly sample 2,526 tweets from the data,  balancing the two classes.

\paragraph{Stance (Abortion, Atheism, Feminism).} The dataset contains 4,870 annotated tweets that express a stance (favour, against, neutral) towards six targets in the United States: atheism, feminism, abortion, climate change, Hillary Clinton. We use the first three in our analysis, respectively subsampling 867, 681, and 881 tweets. The dataset was assembled by \citet{mohammad-etal-2016-semeval} 

\paragraph{Offensive.}
The dataset contains 14,100 tweets which human annotators have labelled as offensive or not. The dataset was assembled by \citet{zampieri2019semeval}. We randomly sampled 2,014 sentences, balancing the classes.

\paragraph{Hate.}
The dataset contains 13,000 tweets which human annotators have labelled as hate speech or not. The dataset was assembled by \citet{basile2019semeval}. We randomly sampled 2,007 sentences, balancing the classes.

\paragraph{Spam.} This dataset contains 1,956 comments under YouTube videos. The task is to identify whether these are spam or not. The data set was assembled by T.C. Alberto and J. V. Lochter and is available from the UCI machine learning repository \citep{uci_repo} under the name \textit{YouTube Spam Collection}. We randomly sampled 743 sentences, balancing the classes. 

\paragraph{Movie reviews.} This dataset contains 10,662 sentences from Rotten Tomatoes movie reviews with either positive or negative sentiment. The dataset was assembled by \citet{Pang+Lee:05a} and can be downloaded from \href{https://huggingface.co/datasets/rotten_tomatoes}{Huggingface}. We randomly sampled 2,000 sentences, balancing the classes.

\paragraph{Legal.} The dataset contains 2,811 legal questions by laypeople that have been assigned to non-mutually-exclusive labels (drawn from the \href{https://taxonomy.legal/}{Legal Issues Taxonomy}) using crowdsourcing.  We create three binary classification tasks based on the labels: payment and debt (Money), employment and job (Work) and criminal issues (Crime). We respectively sub-sample 728, 724, and 648 legal questions for the three tasks, balancing the classes.
The legal questions are often long, which is why we only use the first few sentences until 100 tokens are reached.
The dataset is named \textit{Learned Hands Data} and has been assembled by the the Legal Innovation \& Technology Lab' of Suffolk Law School and Stanford's Legal Design Lab, with the former institute providing a \href{https://spot.suffolklitlab.org/data/}{download link}.

\section{Sentences classified for explainability}\label{sentences}

These are the sentences which were classified for the explainability section. The first 15 have positive sentiment, and the rest have negative sentiment. Please contact the authors to discuss the segmentation and labelling, and the data produced from that exercise.

\begin{enumerate}
\small
\item The company 's scheduled traffic , measured in revenue passenger kilometres RPK , grew by just over 2 \% and nearly 3 \% more passengers were carried on scheduled flights than in February 2009 .
\item Finnish pharmaceuticals company Orion reports profit before taxes of EUR 70.0 mn in the third quarter of 2010 , up from EUR 54.9 mn in the corresponding period in 2009 .
\item O'Leary 's Material Handling Services , located in Perth , is the leading company in Western Australia that supplies , installs and provides service for tail lifts .
\item Net cash flow from operations is expected to remain positive .
\item In the reporting period , the company 's operating profit grew by 43.2 \% to EUR 6 million .
\item Finnish Metso Paper has been awarded a contract for the rebuild of Sabah Forest Industries ' ( SFI ) pulp mill in Sabah , Malaysia .
\item The agreement strengthens our long-term partnership with Nokia Siemens Networks .
\item The diluted loss per share narrowed to EUR 0.27 from EUR 0.86 .
\item Operating profit rose from EUR 1.94 mn to EUR 2.45 mn .
\item Nokia bought Chicago-based Navteq in 2008 , acquiring a maps database to compete with Google s maps as well as with navigation device companies such as TomTom NV and Garmin Ltd. .
\item Sales for the Department Store Division increased by 15 \% and sales for the clothing store subsidiary Seppala increased by 8 \% Meanwhile sales for Hobby Hall decreased by 12 \% .
\item However , the total orders received will still be above last year s levels .
\item We are very proud to be able to use this kind of innovative mobile service for voting in elections .
\item Finnish electronics manufacturer PKC Group Oyj ( OMX Helsinki : PKC1V ) said on Wednesday ( 31 December ) that it has completed the acquisition of MAN Nutzfahrzeuge AG 's cable harness business from MAN Star Trucks \& Buses Spolka zoo in Poland .
\item ` For Nordea , moving into the new headquarters signifies the beginning of a new era .
\item However , the suspect stole his burgundy Nissan Altima .
\item Finnish Bank of +àland reports its operating profit fell to EUR 4.9 mn in the third quarter of 2007 from EUR 5.6 mn in the third quarter of 2006 .
\item stores 16 March 2010 - Finnish stationery and gift retailer Tiimari HEL : TII1V said yesterday that it will cut a total of 28 jobs in its units Tiimari Retail Ltd and Gallerix Finland Ltd as a result of the closure of shops .
\item In October , UPM reported a third-quarter net loss of euro86 million \$ 110 million compared with a net profit of euro120 million in 2007 .
\item Employing 112 in Finland and 280 abroad , the unit recorded first-quarter 2007 sales of 8.6 mln eur , with an operating loss of 1.6 mln eur .
\item Net sales fell by 5 \% from the previous accounting period .
\item The company plans to close two of the three lines at the plant , where some 450 jobs are under threat .
\item Operating profit , excluding non-recurring items , totaled EUR 0.2 mn , down from EUR 0.8 mn in the corresponding period in 2006 .
\item The Elcoteq group recently announced that the last three months of the previous year brought to it a major loss of more than half a billion kroons ( EUR 32 mln ) for the fifth quarter running .
\item More than 14,000 customers were left powerless .
\item Operating profits in the half were 0.8 m , down from 0.9 m as Glisten invested in the brand and the management team .
\item Cash flow after investments amounted to EUR45m , down from EUR46m .
\item The majority of the company 's personnel in Finland is temporarily laid off from one to six weeks in the period from February to June 2009 period .
\item Finnish textiles and clothing group Marimekko Oyj posted a net profit of 7.99 mln euro \$ 10.4 mln for 2006 , compared to 8.4 mln euro \$ 10.9 mln for 2005 .
\item The operating loss amounted to EUR 0.8 mn , compared to a profit of EUR 3.9 mn a year earlier .
\end{enumerate}

\section{Additional results}\label{appx:results}

% Learning curves of remaining data sets
\def\ww{.325}

\begin{figure}[h!t]
 \includegraphics[width=\ww\linewidth]{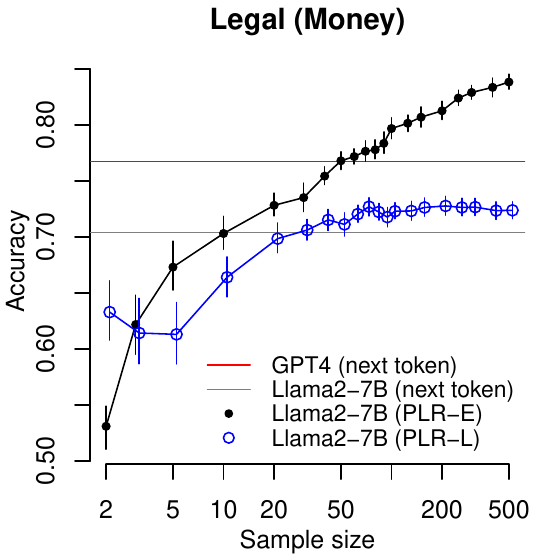}
\includegraphics[width=\ww\linewidth]{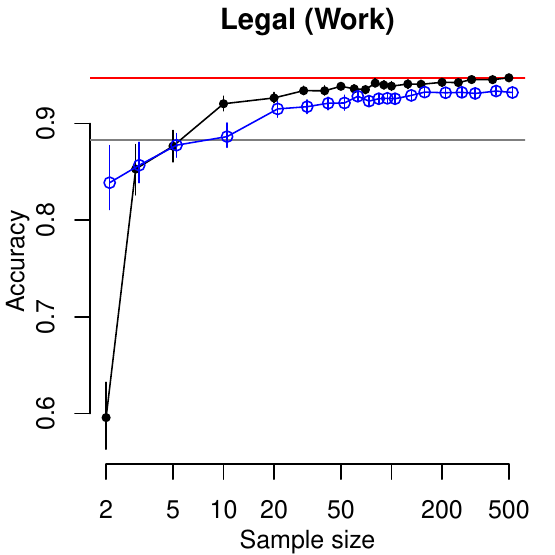}
\includegraphics[width=\ww\linewidth]{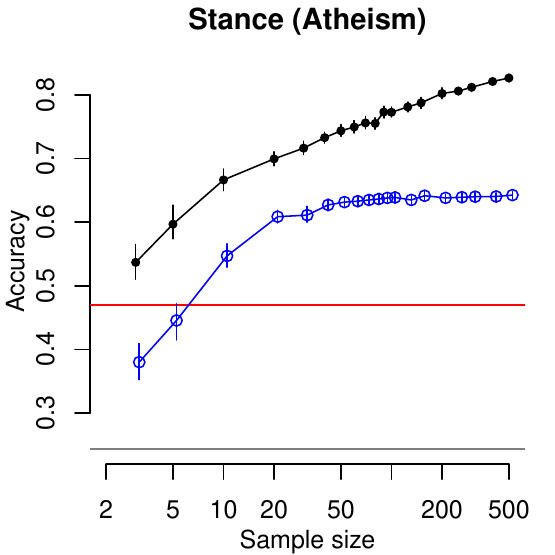}
 \includegraphics[width=\ww\linewidth]{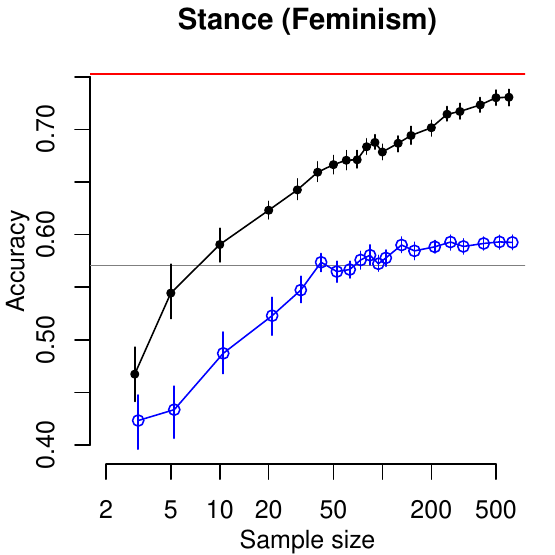}
 \includegraphics[width=\ww\linewidth]{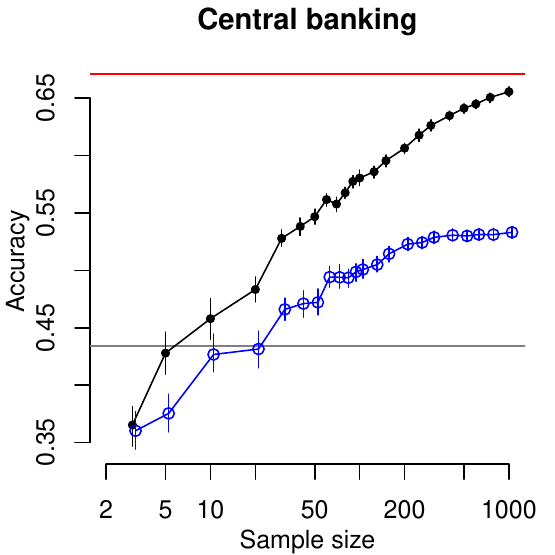}
\caption{\small Continues Figure \ref{fig_baseline_learning}. The accuracies of the zero-shot next token text predictions from GPT-4 and Llama2-7B, along with with the learning curves for the PLR-L and PLR-E methods applied to our baseline model (Llama2-7B q4.0).}
\label{fig_baseline_learning_rest}
\end{figure}
\def\ww{.315}
\begin{figure}[h!t]

 \includegraphics[width=\ww\linewidth]{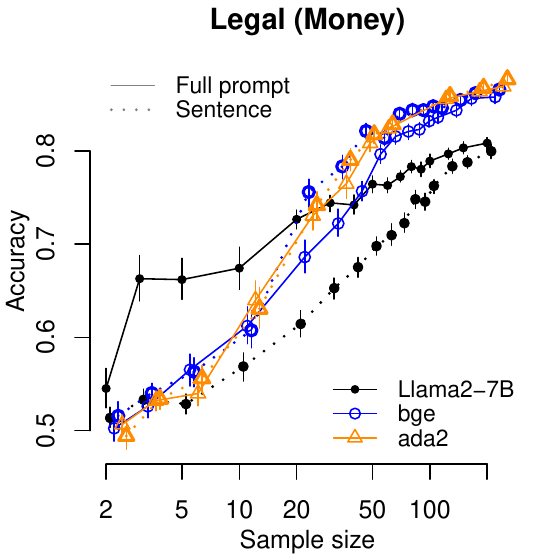}
\includegraphics[width=\ww\linewidth]{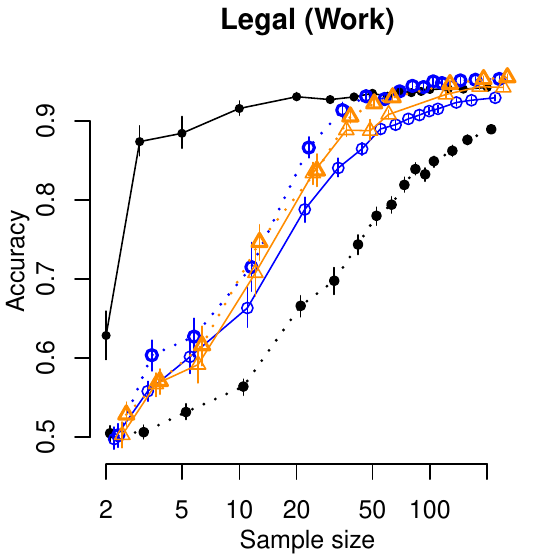}
\includegraphics[width=\ww\linewidth]{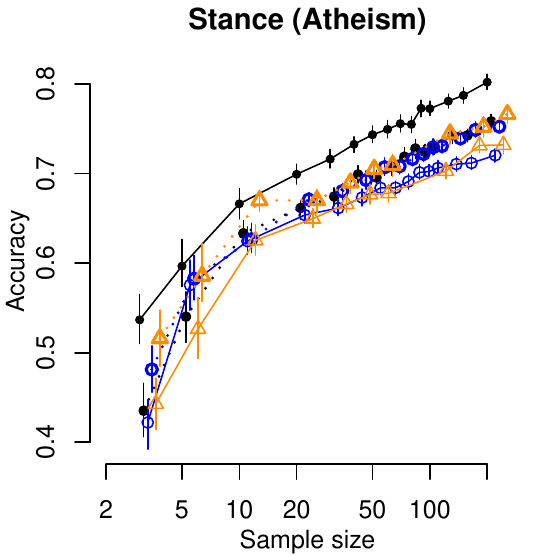}
 \includegraphics[width=\ww\linewidth]{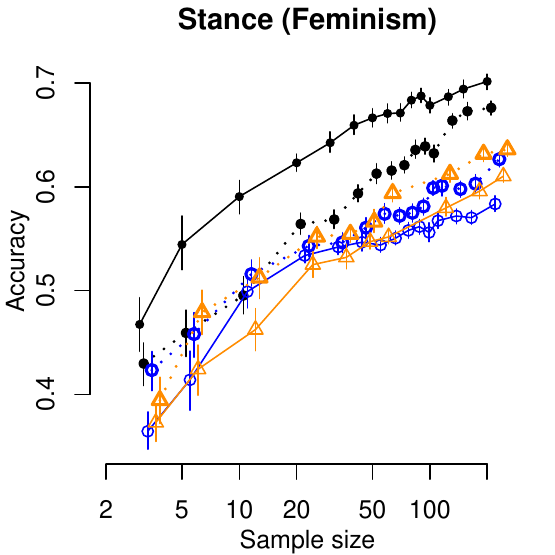}
 \includegraphics[width=\ww\linewidth]{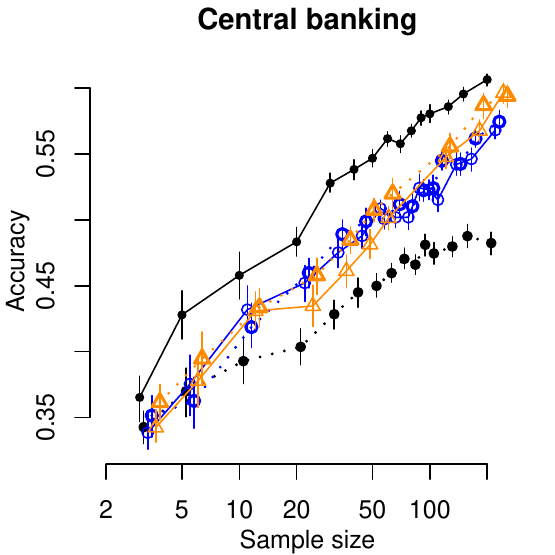}

\caption{\small Continues Figure \ref{fig_plr_learning}. The accuracy of PLR-E when trained on embeddings from different models and promptings.}
\label{fig_plr_learning_rest}
\end{figure}
\def\ww{.245}

\begin{figure}[h!t]

\includegraphics[width=\ww\linewidth]{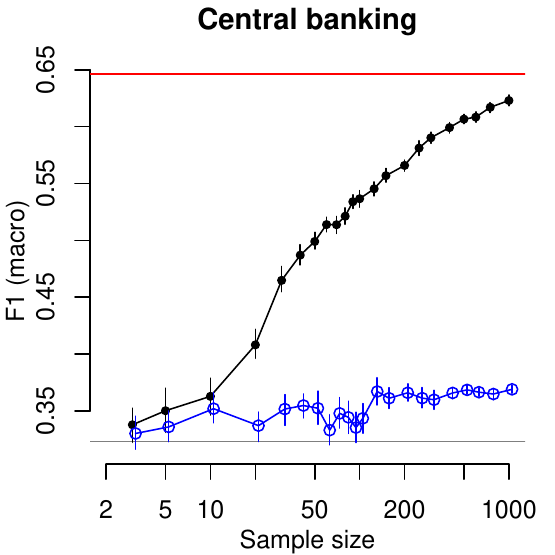}
\includegraphics[width=\ww\linewidth]{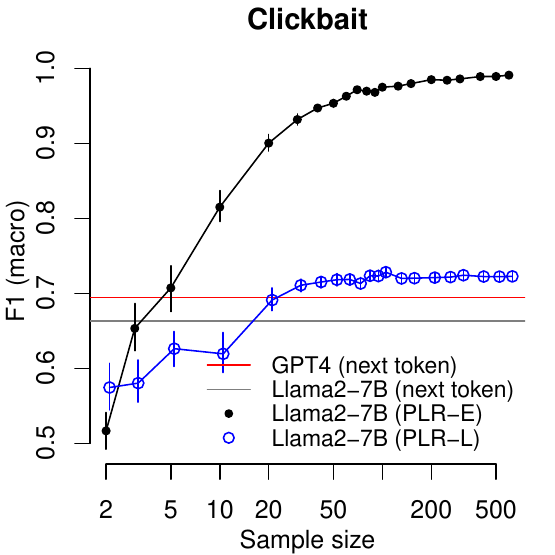}
\includegraphics[width=\ww\linewidth]{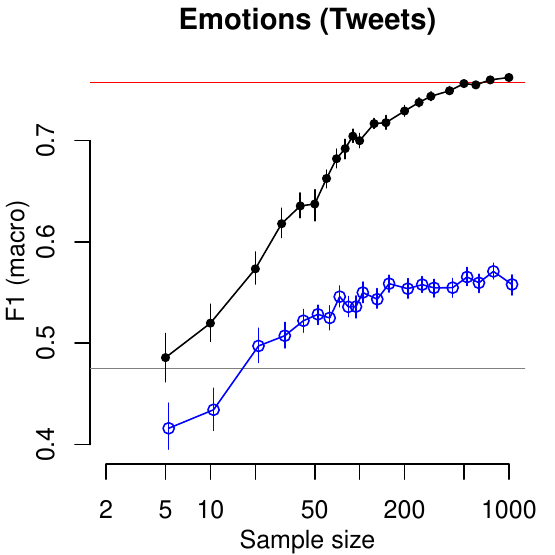}
\includegraphics[width=\ww\linewidth]{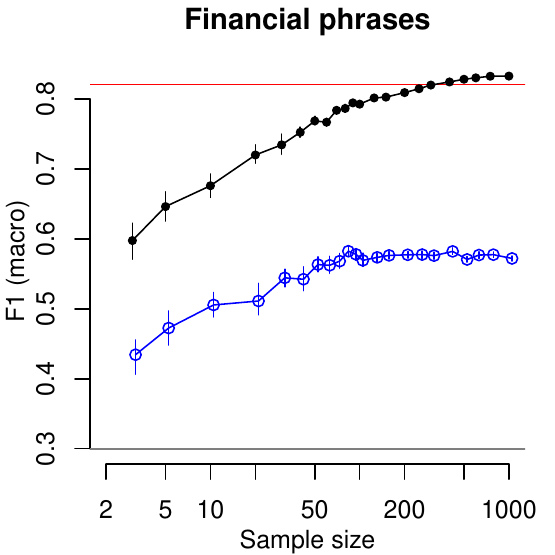}
\includegraphics[width=\ww\linewidth]{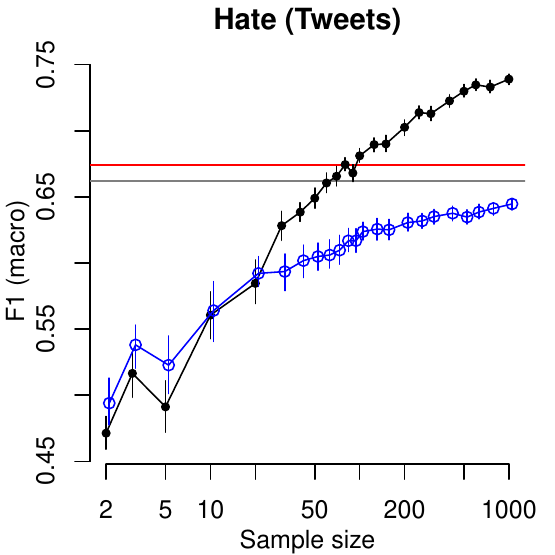}
\includegraphics[width=\ww\linewidth]{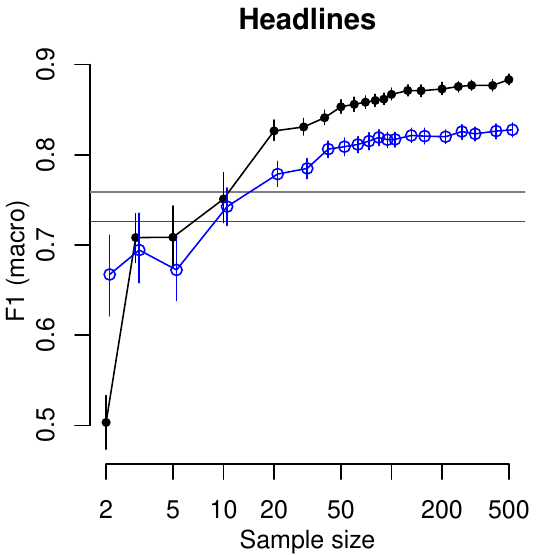}
\includegraphics[width=\ww\linewidth]{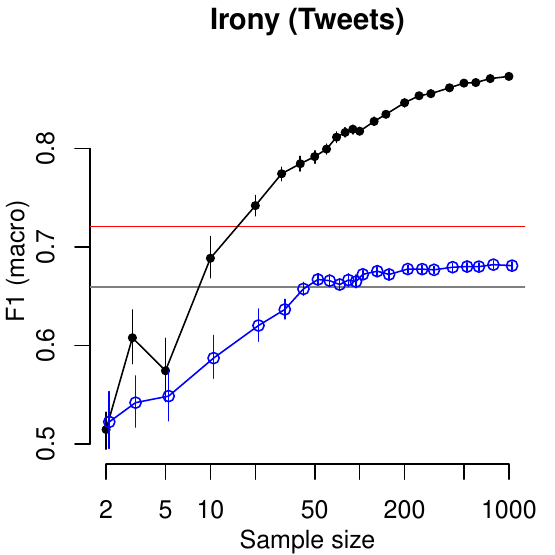}
\includegraphics[width=\ww\linewidth]{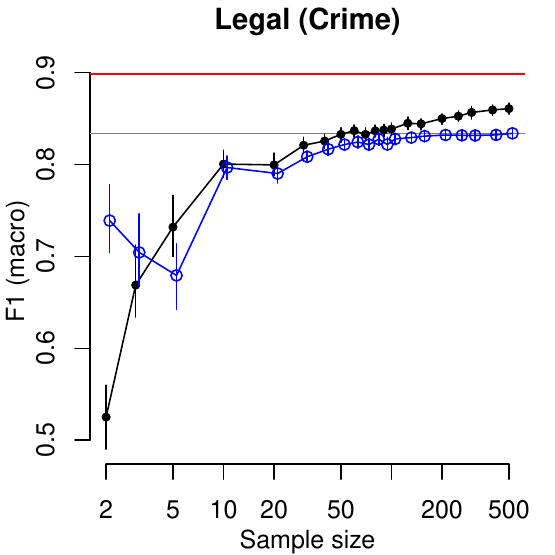}
\includegraphics[width=\ww\linewidth]{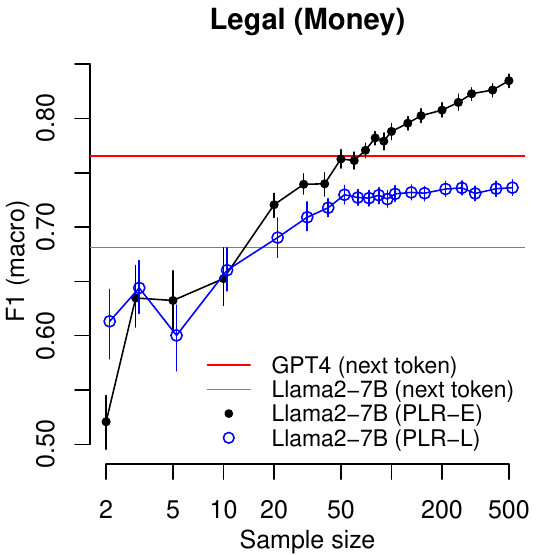}
\includegraphics[width=\ww\linewidth]{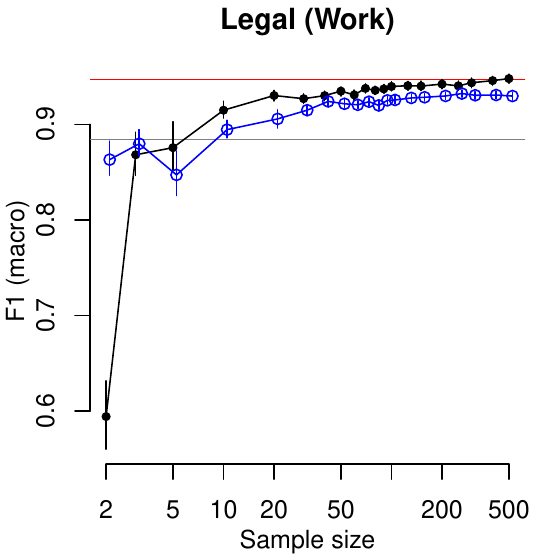}
\includegraphics[width=\ww\linewidth]{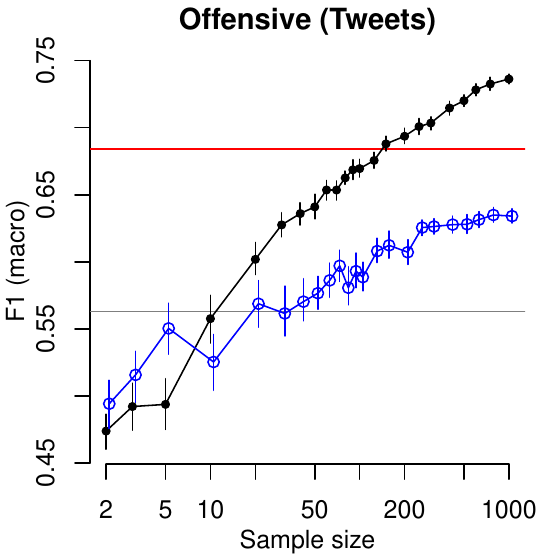}
\includegraphics[width=\ww\linewidth]{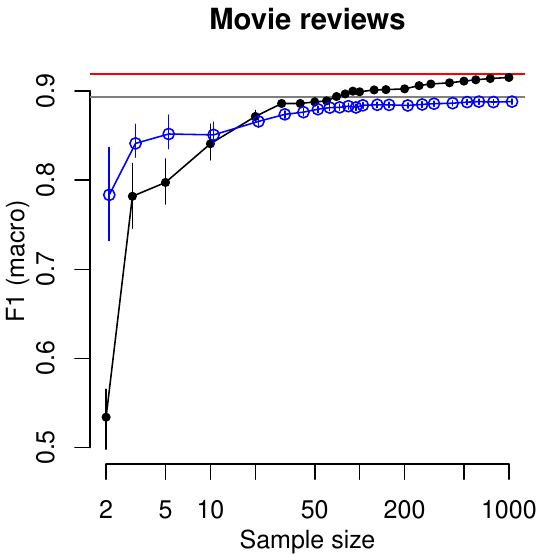}
\includegraphics[width=\ww\linewidth]{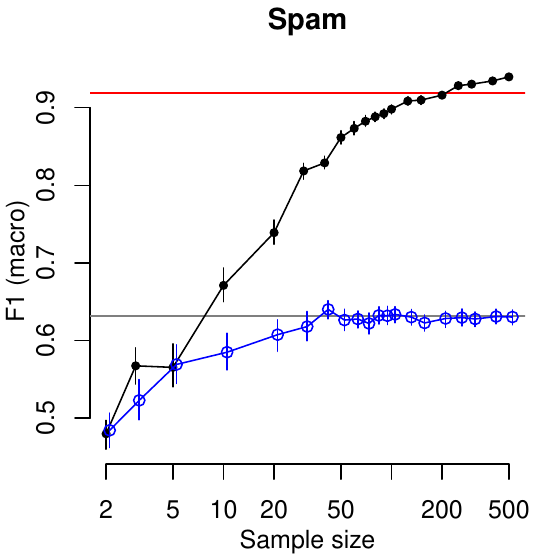} 
\includegraphics[width=\ww\linewidth]{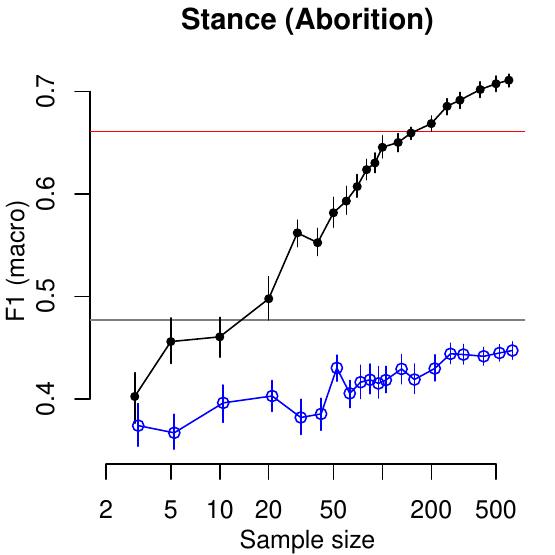}
\includegraphics[width=\ww\linewidth]{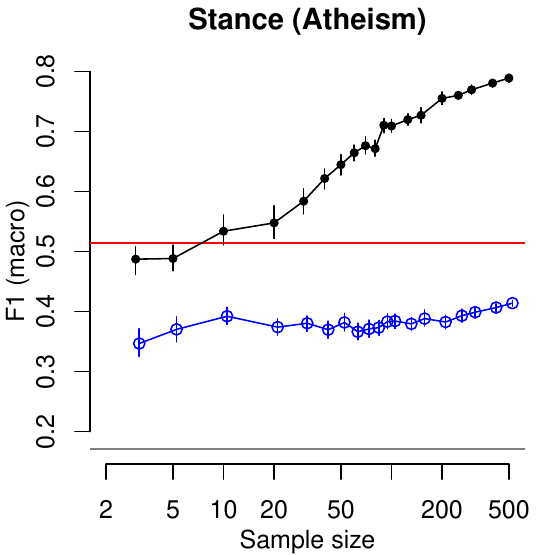}
\includegraphics[width=\ww\linewidth]{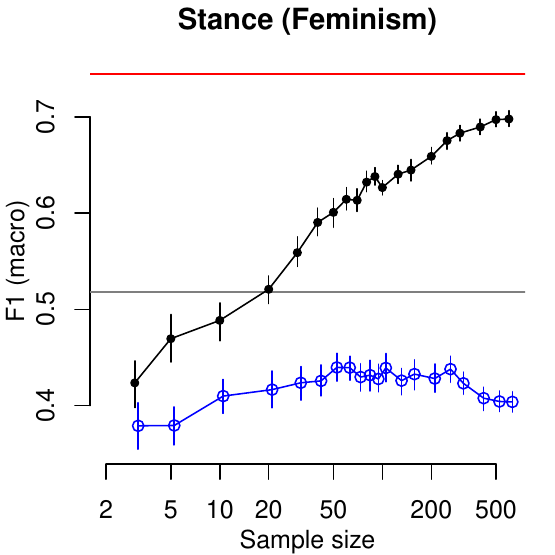}
\includegraphics[width=\ww\linewidth]{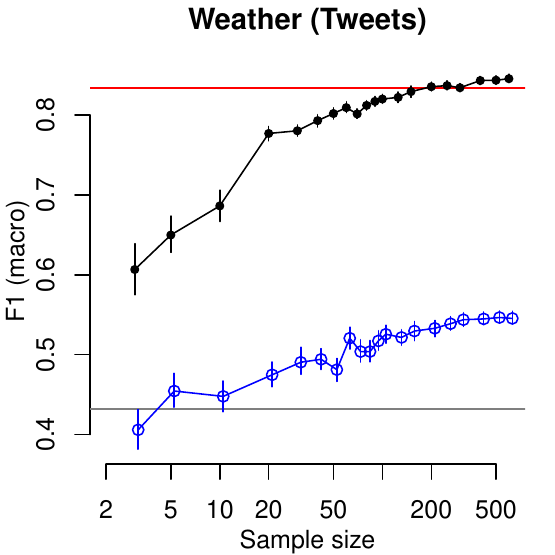}

\caption{\small The F1 macro score of the zero-shot next token text predictions from GPT-4 and Llama2-7B, along with with the learning curves for the PLR-L and PLR-E methods applied to our baseline model (Llama2-7B q4.0). This is the analogue, using F1 instead of accuracy, of Figures \ref{fig_baseline_learning} and \ref{fig_baseline_learning_rest}. }
\label{fig_baseline_learning_f1}
\end{figure}
%
%% F1 performance table

\addtolength{\tabcolsep}{-3pt}    
\begin{table}[ht]
\begin{small}
\centering
\begin{tabular}{llllllllll}
  \hline
Datasets & \sc{GPT-4} & \multicolumn{2}{c}{\sc{Llama2 7B}} &\multicolumn{2}{c}{\sc{Llama2 13B}} &  \multicolumn{2}{c}{\sc{Zephyr 3B}} & \sc{Ada2} & \sc{BGE} \\ 
  &Token &  PRL-E &  Token &  PLR-E &  Token &  PLR-E  &  Token &  PLR-E & PLR-E \\ 

  \hline
Central banking & \underline{\textbf{0.65}} & 0.54 & 0.32 & 0.56 & 0.42 & 0.56 & 0.24 & 0.50 & 0.45 \\ 
  Clickbait & 0.69 & \underline{\textbf{0.97}} & 0.67 & \textbf{0.97} & 0.75 & 0.96 & 0.40 & 0.96 & 0.94 \\ 
  Headlines & 0.73 & \underline{\textbf{0.87}} & 0.76 & 0.86 & 0.62 & 0.86 & 0.45 & 0.84 & 0.79 \\ 
  Spam & \underline{0.92} & 0.90 & 0.64 & 0.89 & 0.64 & 0.92 & 0.55 & \textbf{0.93} & 0.91 \\ 
  Financial phrases & \underline{\textbf{0.82}} & 0.79 & 0.30 & 0.80 & 0.60 & 0.81 & 0.68 & 0.54 & 0.52 \\ 
  Weather (Tweets) & \underline{\textbf{0.83}} & 0.82 & 0.43 & 0.81 & 0.67 & \textbf{0.83} & 0.77 & 0.78 & 0.75 \\ 
  Irony (Tweets) & 0.72 & \underline{\textbf{0.82}} & 0.66 & 0.81 & 0.50 & 0.66 & 0.52 & 0.59 & 0.59 \\ 
  Emotions (Tweets) & \underline{\textbf{0.76}} & 0.70 & 0.48 & 0.68 & 0.61 & 0.74 & 0.65 & 0.53 & 0.52 \\ 
  Offensive (Tweets) & \underline{0.68} & 0.67 & 0.56 & 0.66 & 0.62 & \textbf{0.70} & \textbf{0.70} & 0.65 & 0.66 \\ 
  Hate (Tweets) & 0.67 & \underline{0.68} & 0.66 & 0.60 & 0.64 & \textbf{0.71} & 0.67 & 0.69 & 0.67 \\ 
  Stance (Feminism) & \underline{\textbf{0.74}} & 0.63 & 0.52 & 0.61 & 0.45 & 0.58 & 0.36 & 0.44 & 0.43 \\ 
  Stance (Aborition) & \underline{0.66} & 0.65 & 0.48 & \textbf{0.70} & 0.51 & 0.55 & 0.37 & 0.52 & 0.51 \\ 
  Stance (Atheism) & 0.51 & \underline{0.71} & 0.17 & \textbf{0.77} & 0.28 & 0.70 & 0.48 & 0.52 & 0.54 \\ 
  Movie reviews & \underline{\textbf{0.92}} & 0.90 & 0.89 & 0.89 & 0.87 & 0.90 & 0.86 & 0.82 & 0.75 \\ 
  Legal (Money) & 0.77 & \underline{0.80} & 0.69 & 0.77 & 0.73 & 0.78 & 0.68 & \textbf{0.85} & 0.83 \\ 
  Legal (Work) & \underline{0.95} & 0.94 & 0.88 & 0.96 & \textbf{0.97} & 0.94 & 0.89 & 0.93 & 0.91 \\ 
  Legal (Crime) & \underline{\textbf{0.90}} & 0.83 & 0.83 & 0.86 & 0.86 & 0.86 & 0.77 & 0.84 & 0.81 \\ \hline
Mean & 0.76 & \underline{\textbf{0.78}} & 0.59 & \textbf{0.78} & 0.63 & 0.77 & 0.59 & 0.70 & 0.68 \\ 
  Median & 0.74 & \underline{\textbf{0.80}} & 0.64 & \textbf{0.80} & 0.62 & 0.78 & 0.65 & 0.69 & 0.67 \\
   \hline
\end{tabular}
\caption{\small Comparison of the F1 macro score of different models. PLR-E methods are trained on 100 samples. This is the analogue, using F1 instead of accuracy, of Table  \ref{tab_perf_acc}.}
\label{tab_perf_f1}
\end{small}
\end{table}
\addtolength{\tabcolsep}{3pt}    
%
%% Learning curves of all data sets (F1-macro)
%
%
%% Few shott-learning using few-shot embedding
\begin{figure}[!h]

\includegraphics[width=.325\linewidth]{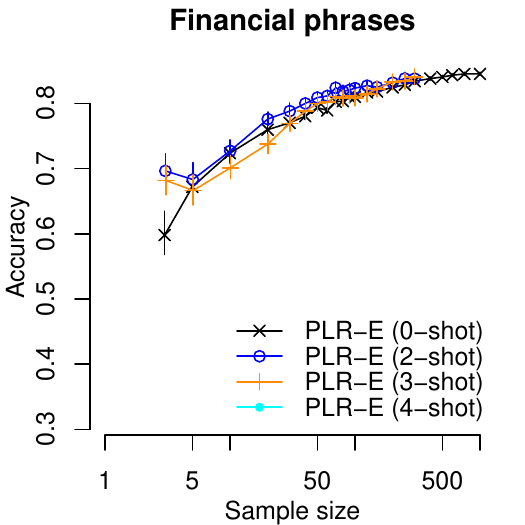}
\includegraphics[width=.325\linewidth]{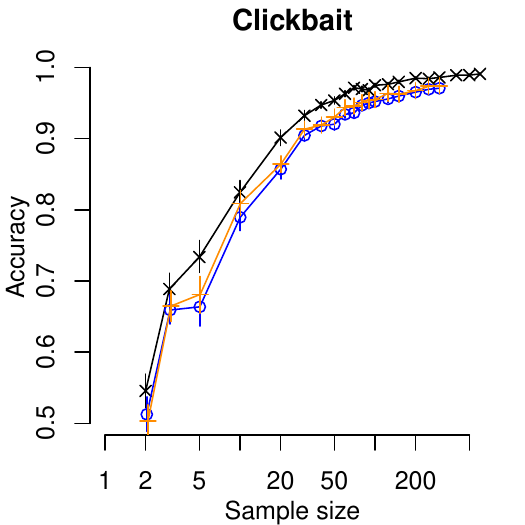}
\includegraphics[width=.325\linewidth]{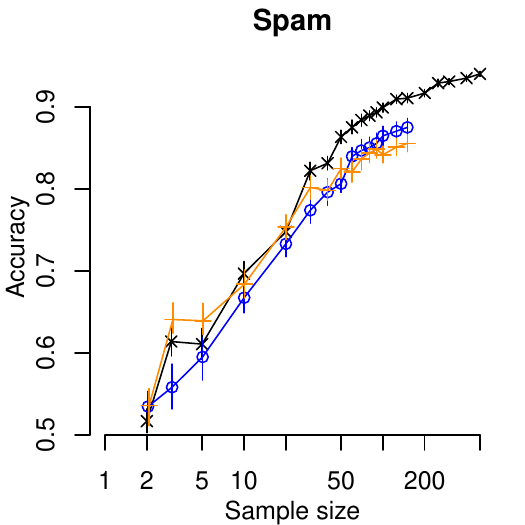} \\
\includegraphics[width=.325\linewidth]{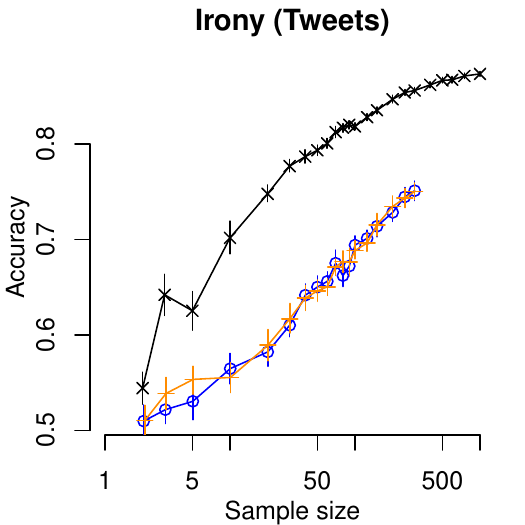}
\includegraphics[width=.325\linewidth]{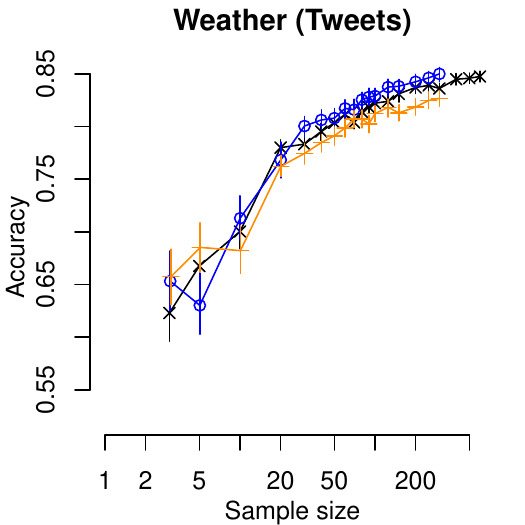}
\includegraphics[width=.325\linewidth]{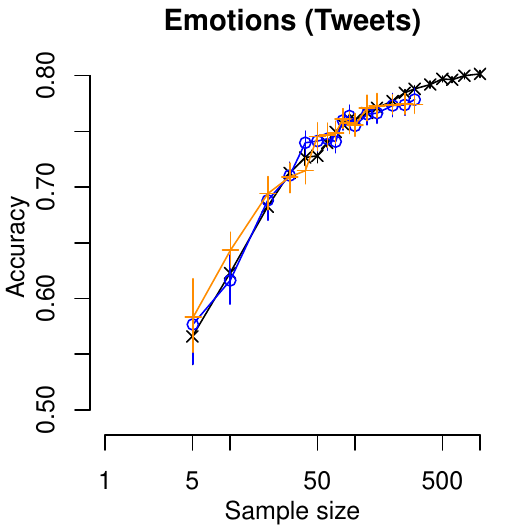}
\includegraphics[width=.325\linewidth]{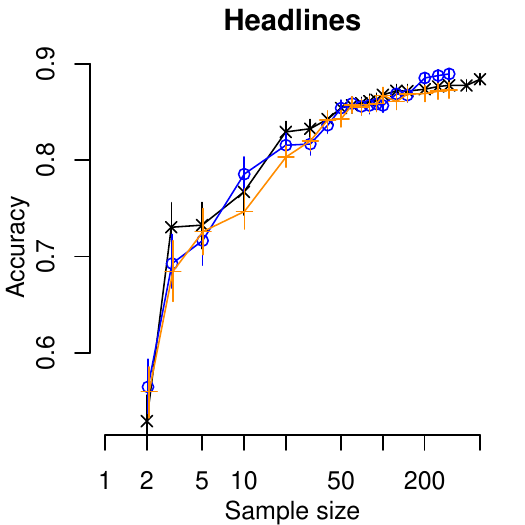}
\includegraphics[width=.325\linewidth]{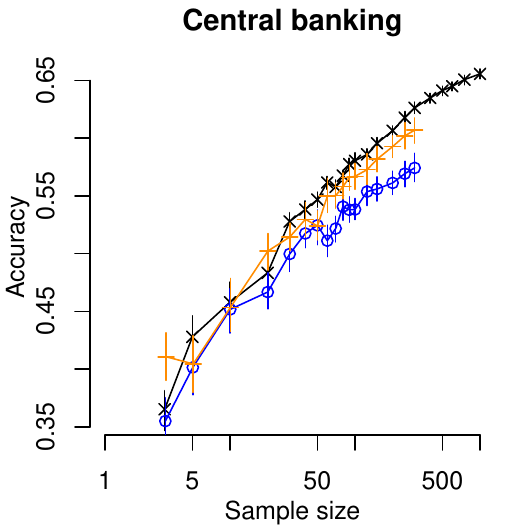}

\caption{\small Using PLR-E on the embeddings from zero- and few-shot prompting of our baseline model.}
\label{fig_learning_few_shot_regression}

\end{figure}
\def\ww{.325}
\begin{figure}[!ht]

\includegraphics[width=\ww\linewidth]{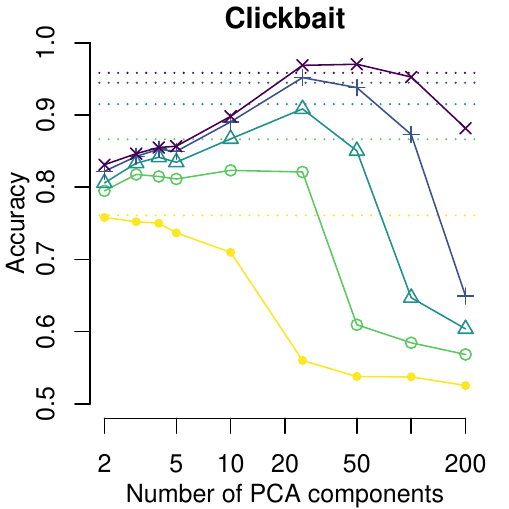}
\includegraphics[width=\ww\linewidth]{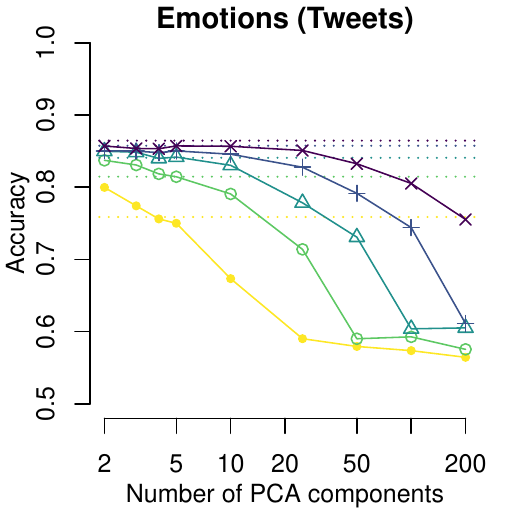}
\includegraphics[width=\ww\linewidth]{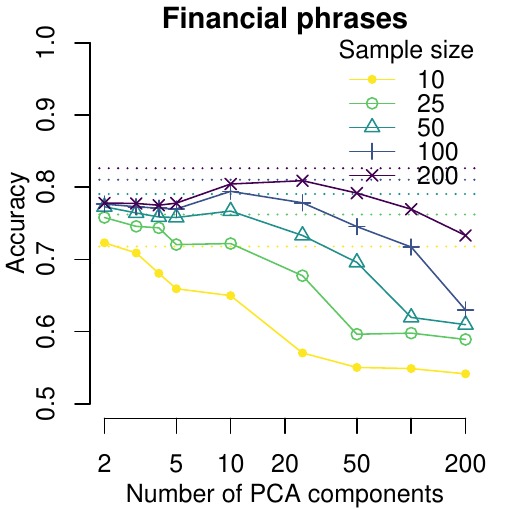}
\includegraphics[width=\ww\linewidth]{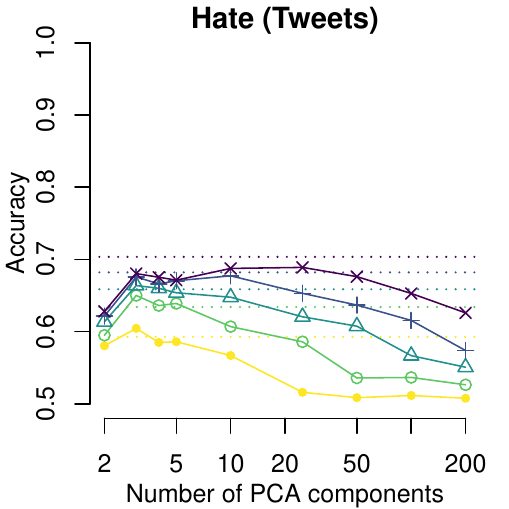}
\includegraphics[width=\ww\linewidth]{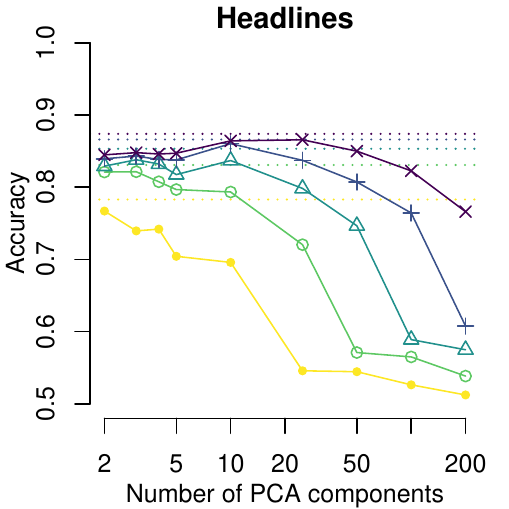}
\includegraphics[width=\ww\linewidth]{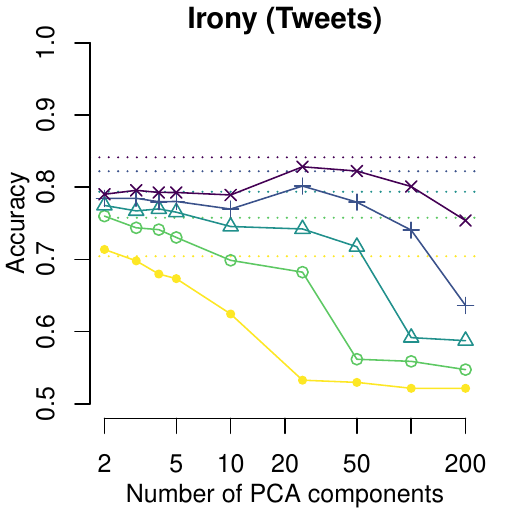}
\includegraphics[width=\ww\linewidth]{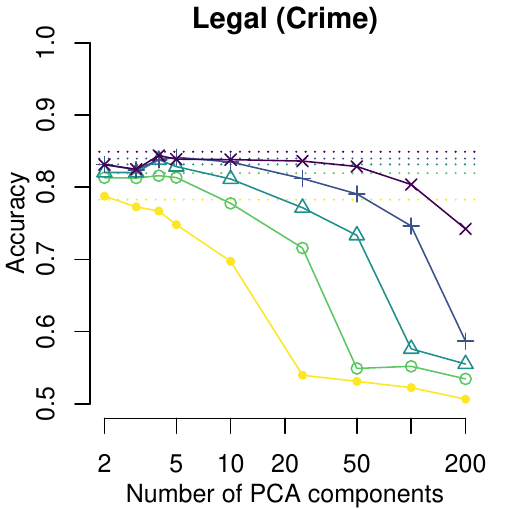}
\includegraphics[width=\ww\linewidth]{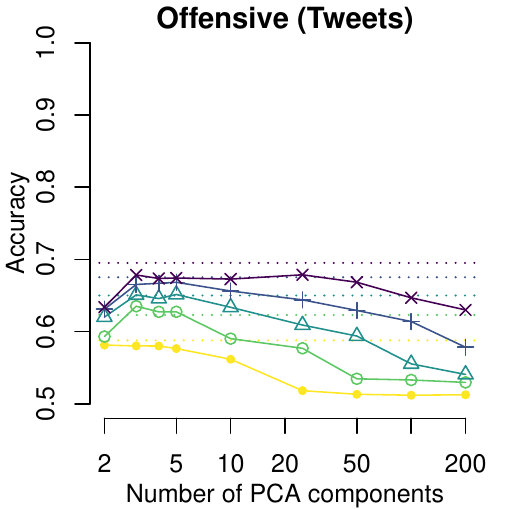}
\includegraphics[width=\ww\linewidth]{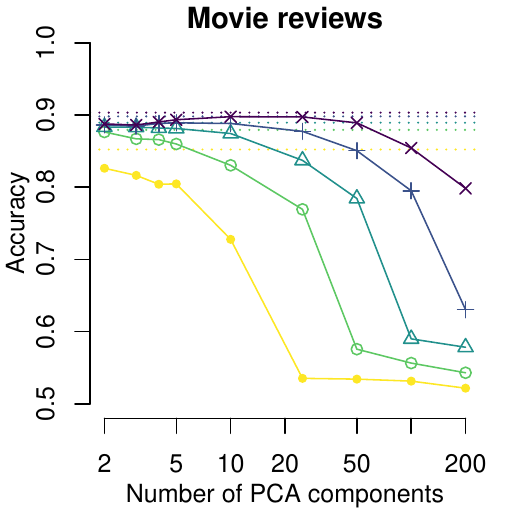}
\includegraphics[width=\ww\linewidth]{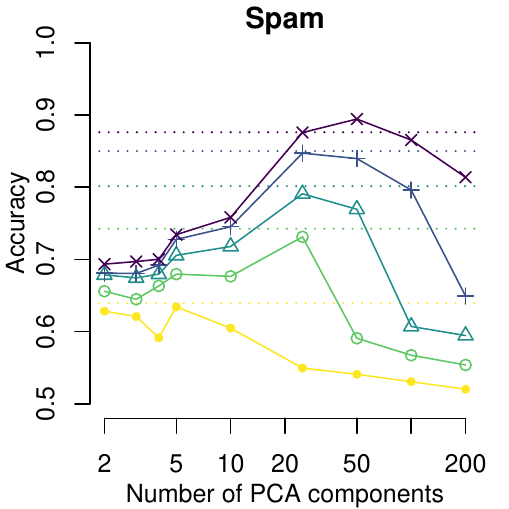} 
\includegraphics[width=\ww\linewidth]{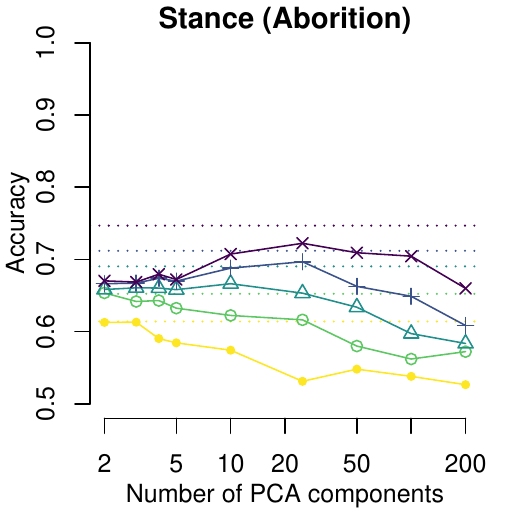}
\includegraphics[width=\ww\linewidth]{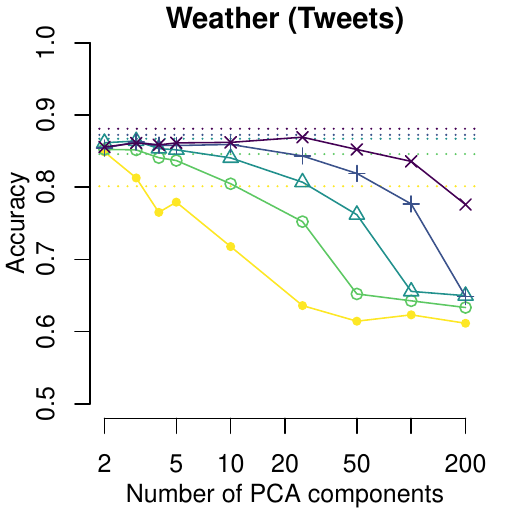}

\caption{\small Accuracy as a function of the number of (normalised) principle components for a given sample size (colour and symbol), c.f. Figure \ref{fig_pca_learning}.}
\label{fig_pca_learning_scaled}
\end{figure}
\def\ww{.325}
\begin{figure}[!ht]
\includegraphics[width=\ww\linewidth]{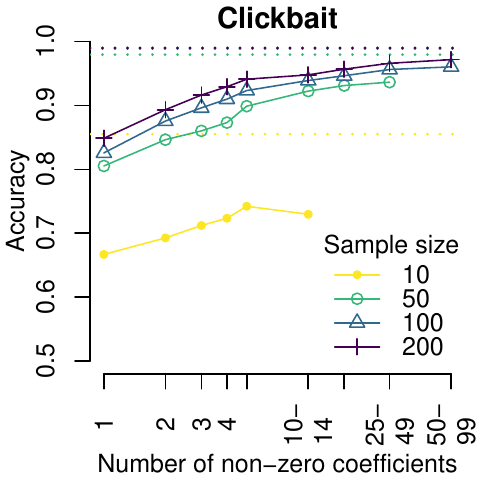}
\includegraphics[width=\ww\linewidth]{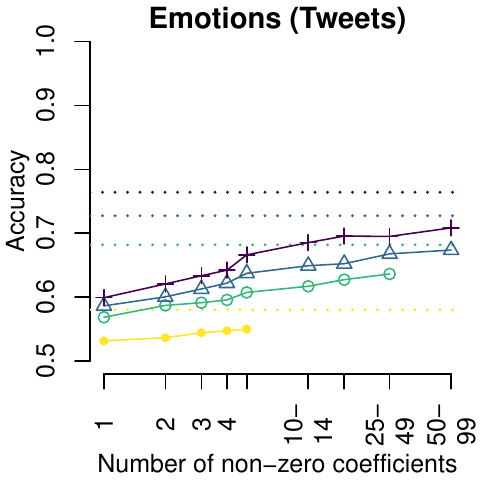}
\includegraphics[width=\ww\linewidth]{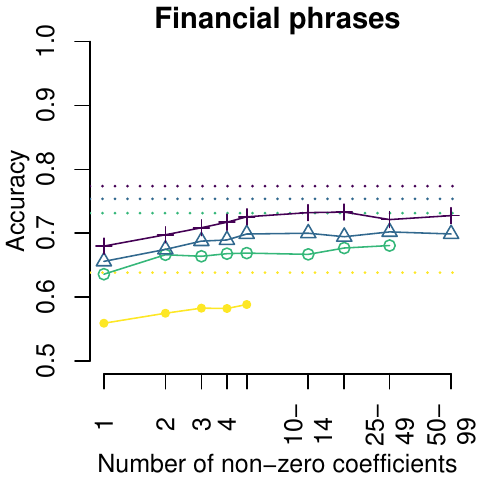}
\includegraphics[width=\ww\linewidth]{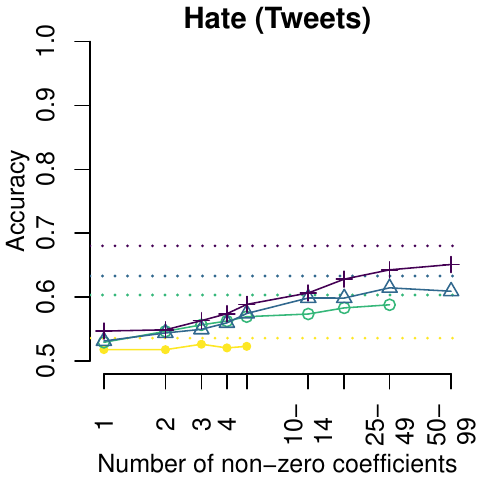}
\includegraphics[width=\ww\linewidth]{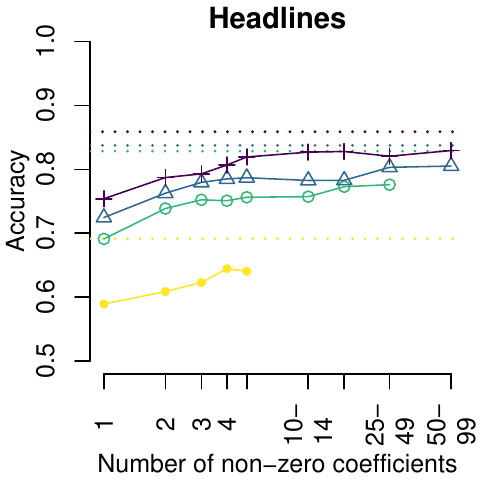}
\includegraphics[width=\ww\linewidth]{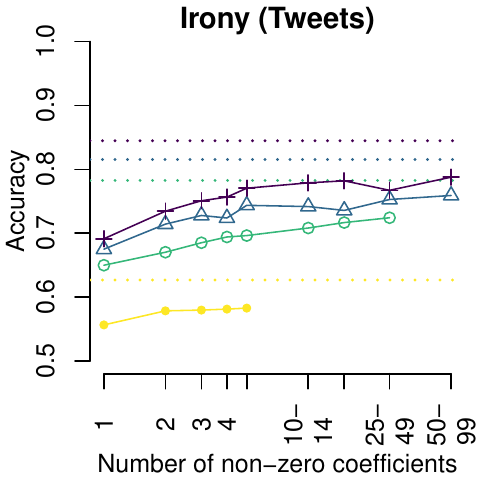}
\includegraphics[width=\ww\linewidth]{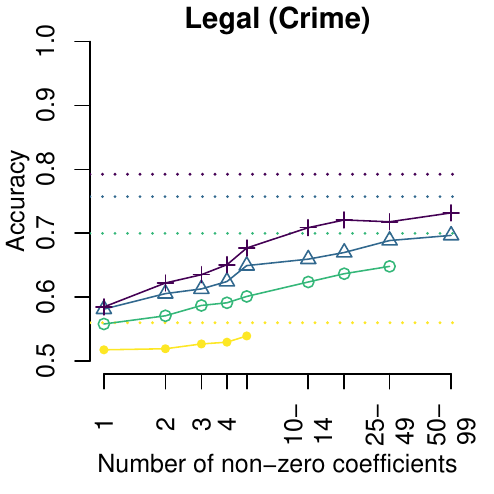}
\includegraphics[width=\ww\linewidth]{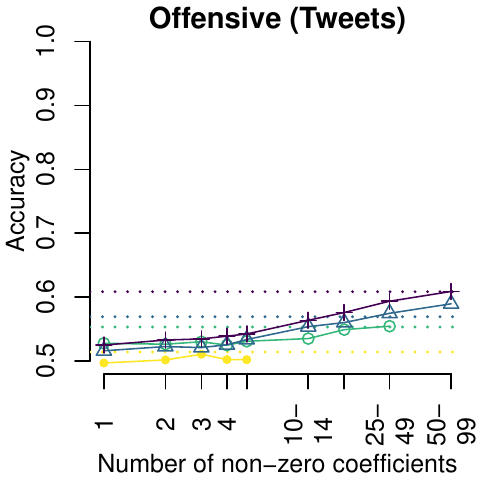}
\includegraphics[width=\ww\linewidth]{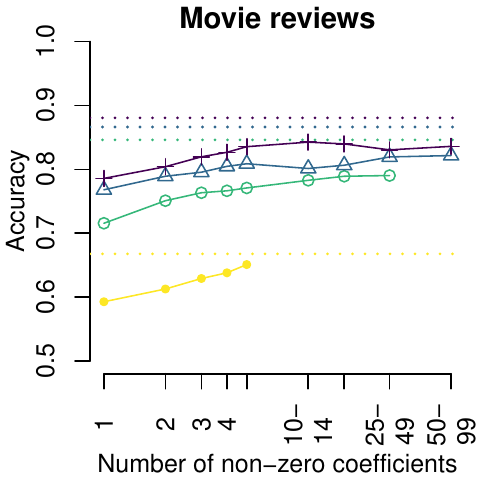}
\includegraphics[width=\ww\linewidth]{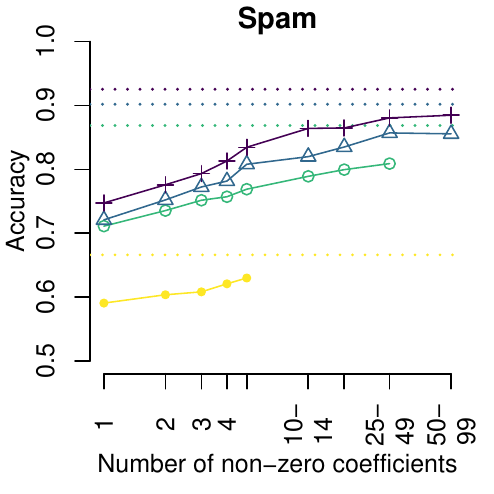} 
\includegraphics[width=\ww\linewidth]{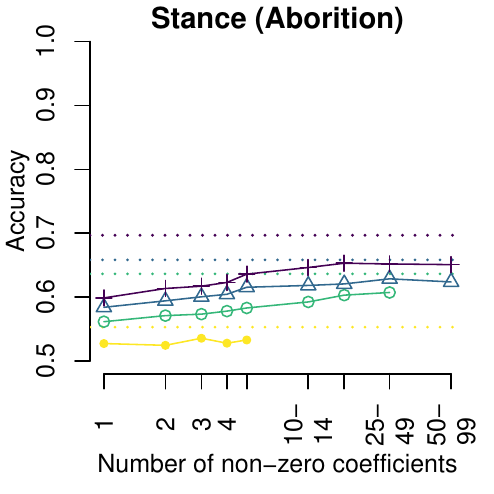}
\includegraphics[width=\ww\linewidth]{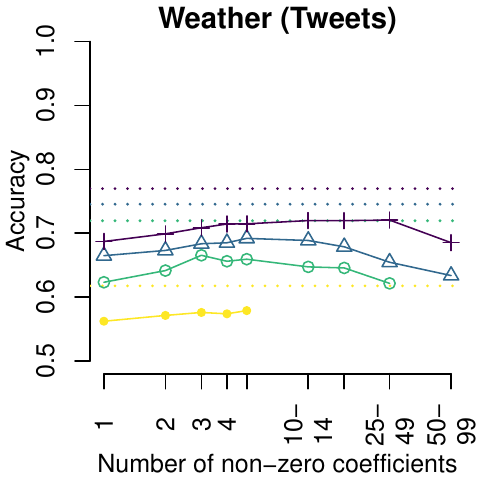}
\caption{\small The accuracy of the Lasso regression as a function of the number of non-zero coefficients for different sample sizes (colour and symbol). The embeddings are produced without the surrounding prompt, c.f. Figure \ref{sparse_learning}.}
\label{sparse_learning_noq}
\end{figure}

\def\ww{.32}
\begin{figure}[h!t]
\centering
\includegraphics[width=\ww\linewidth]{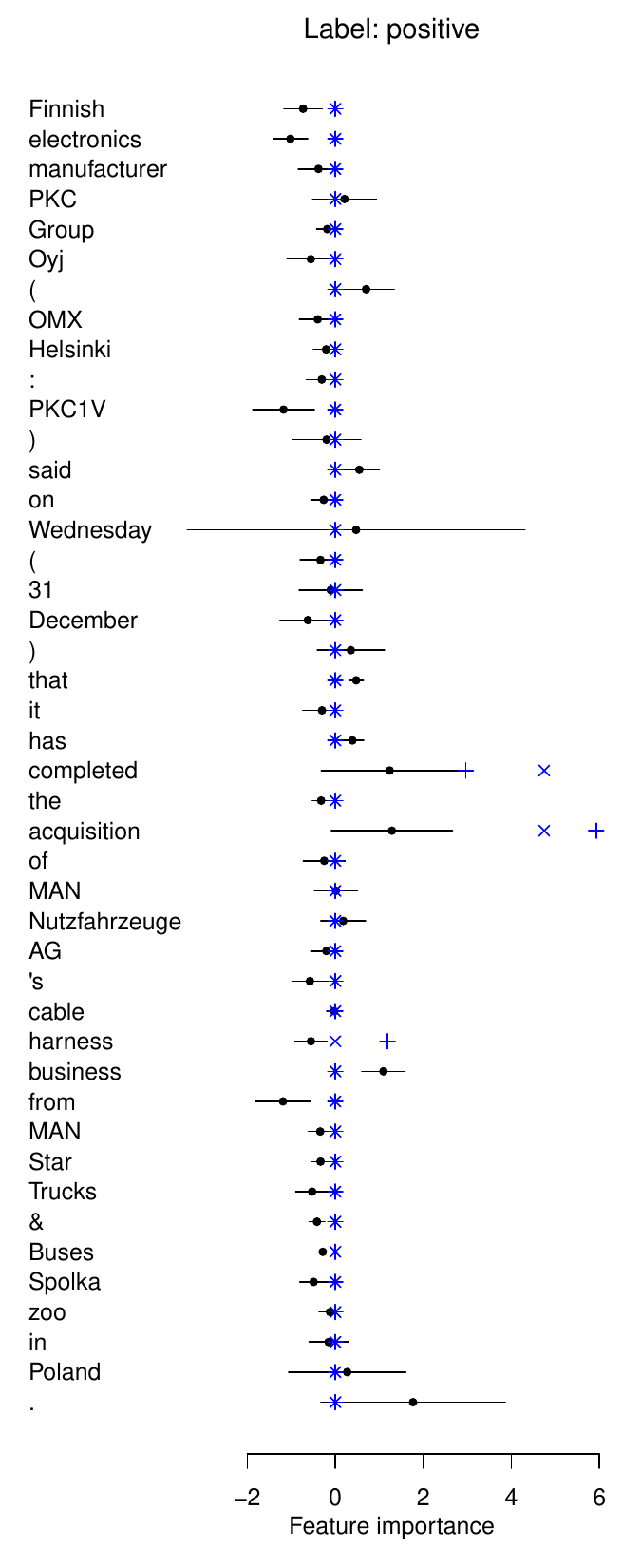}
\includegraphics[width=\ww\linewidth]{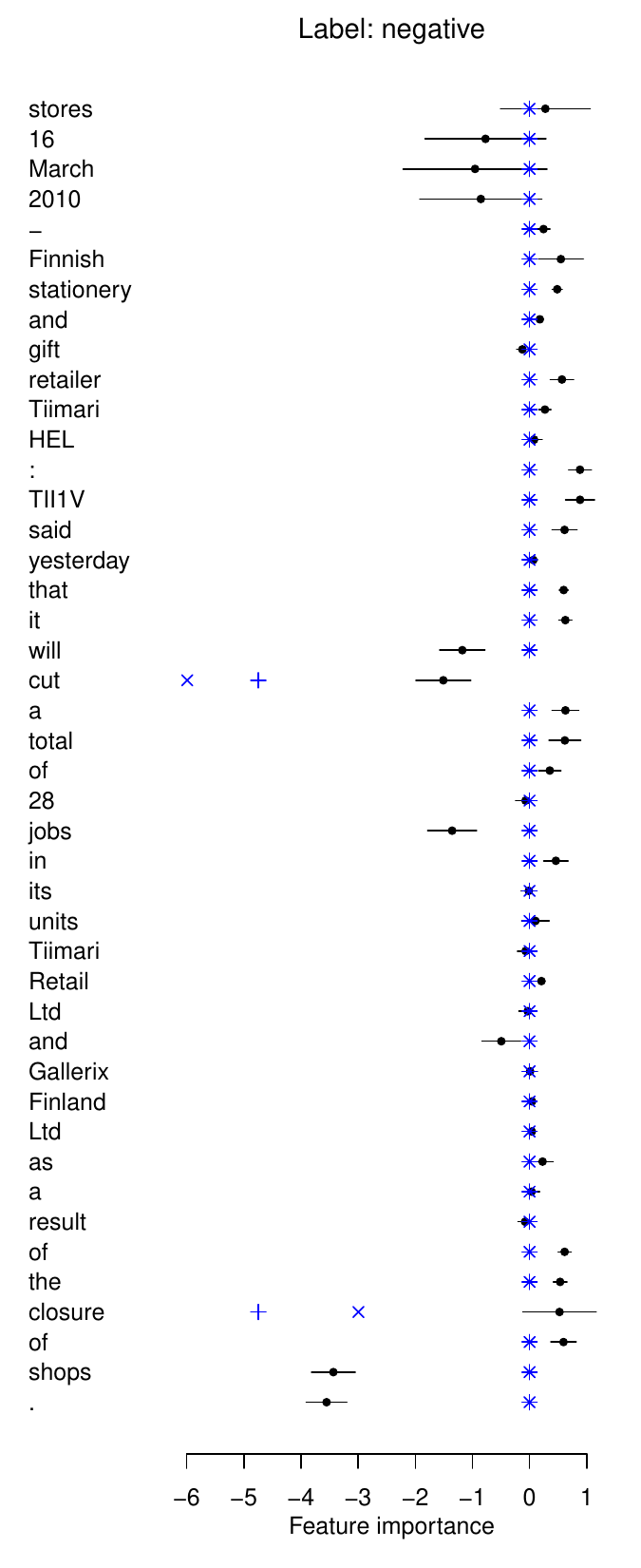}
\includegraphics[width=\ww\linewidth]{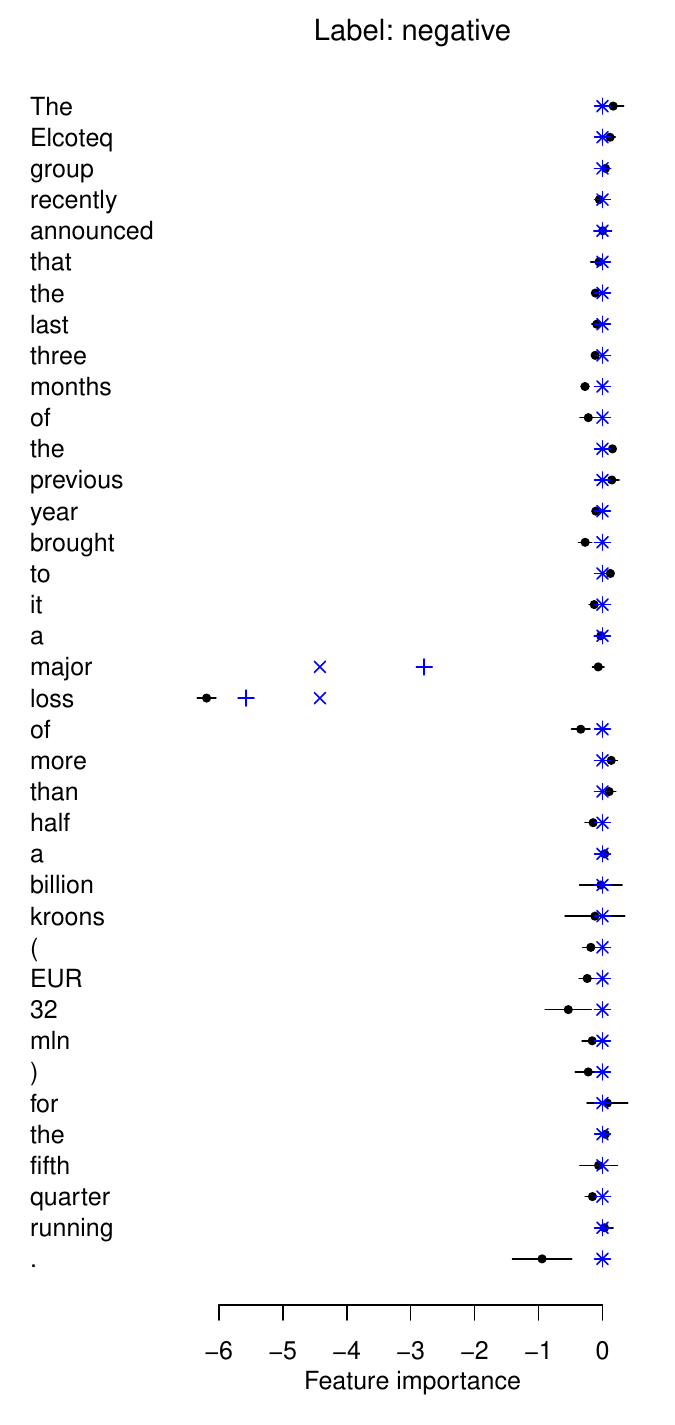}
\includegraphics[width=\ww\linewidth]{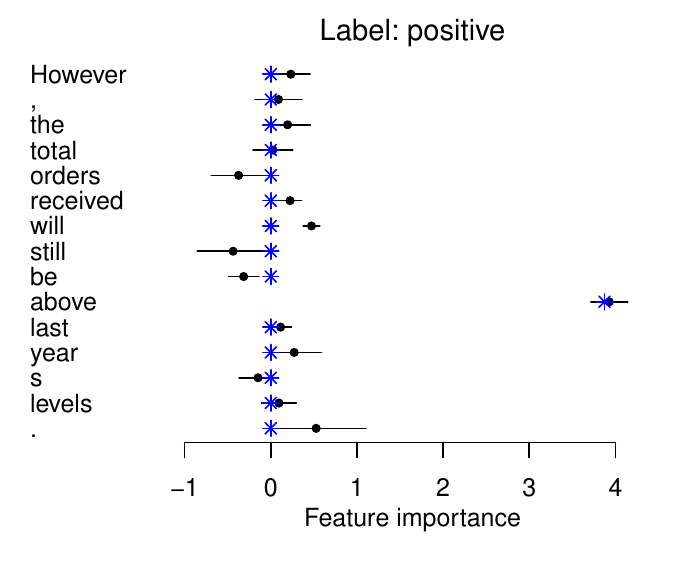}
\includegraphics[width=\ww\linewidth]{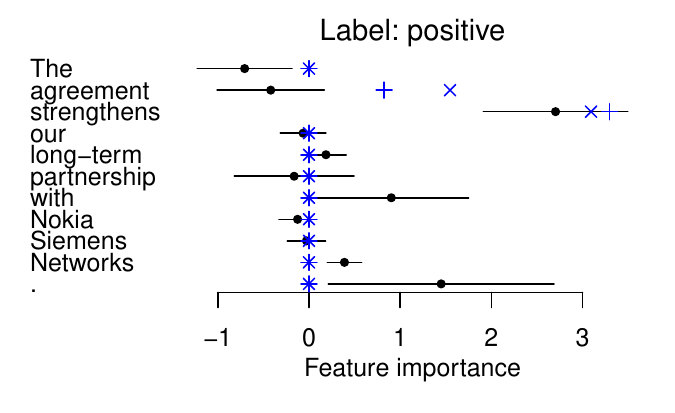}
\includegraphics[width=\ww\linewidth]{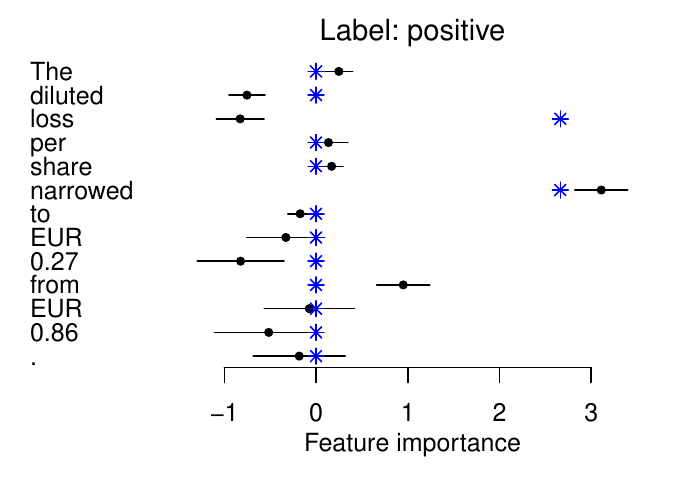}

\caption{\small \captionexplain}
\label{word_level_explain_appx_1}
\end{figure}

\begin{figure}[h!t]
\centering
\includegraphics[width=\ww\linewidth]{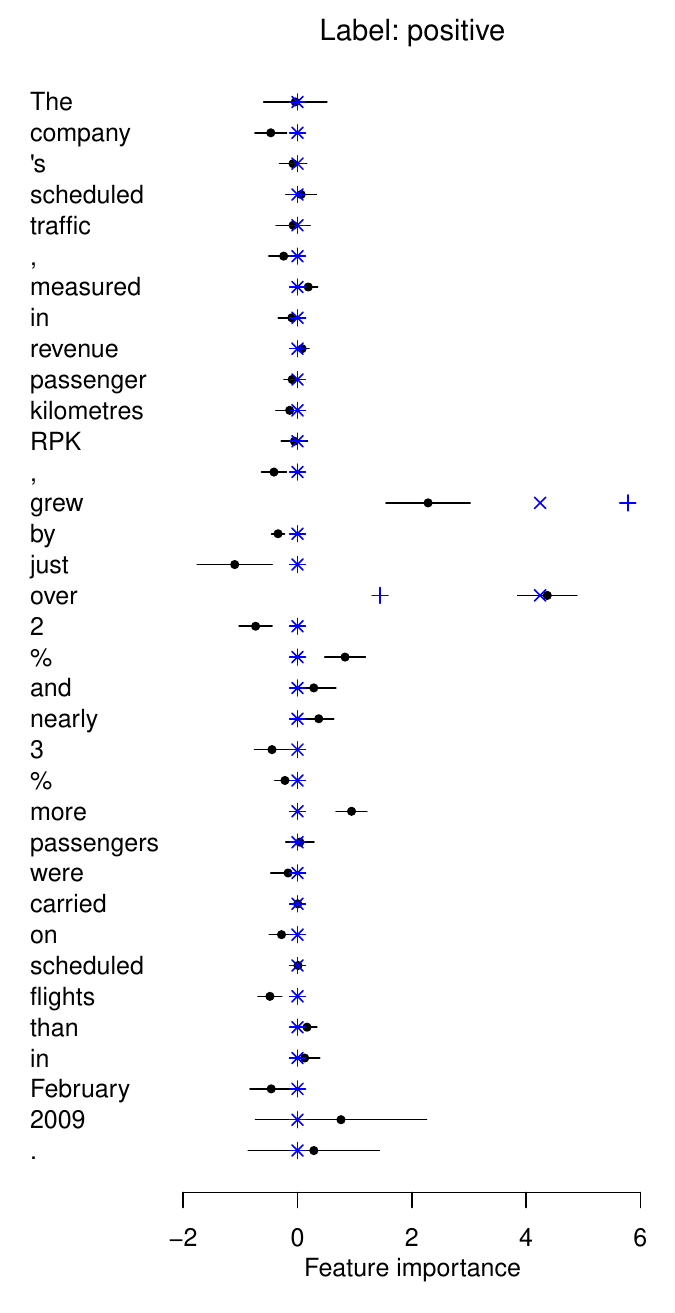}
\includegraphics[width=\ww\linewidth]{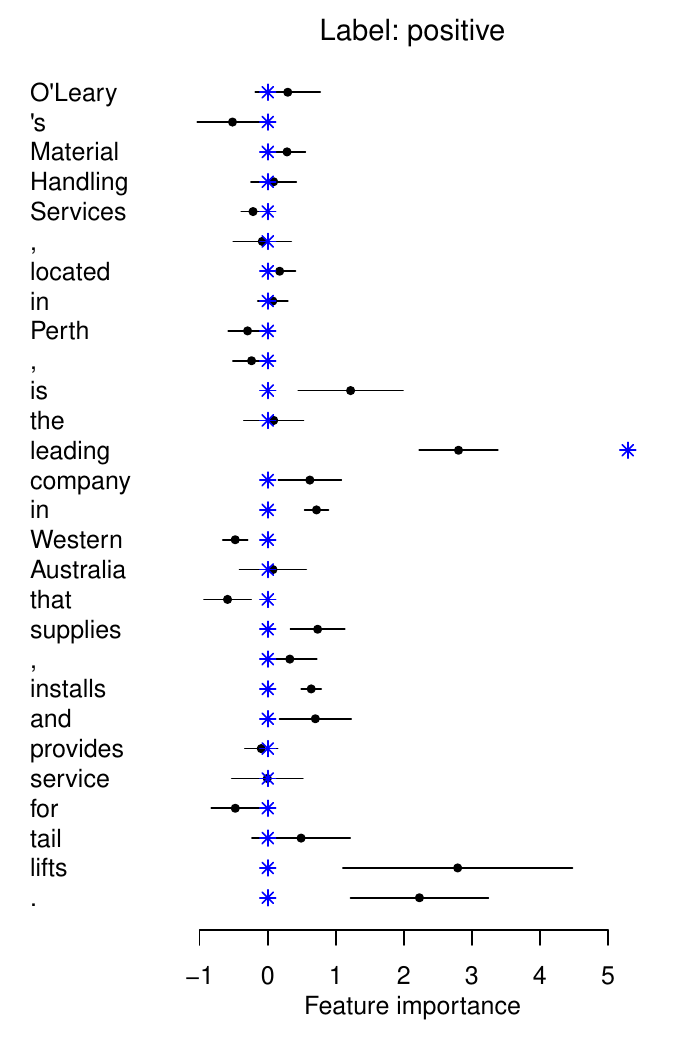}
\includegraphics[width=\ww\linewidth]{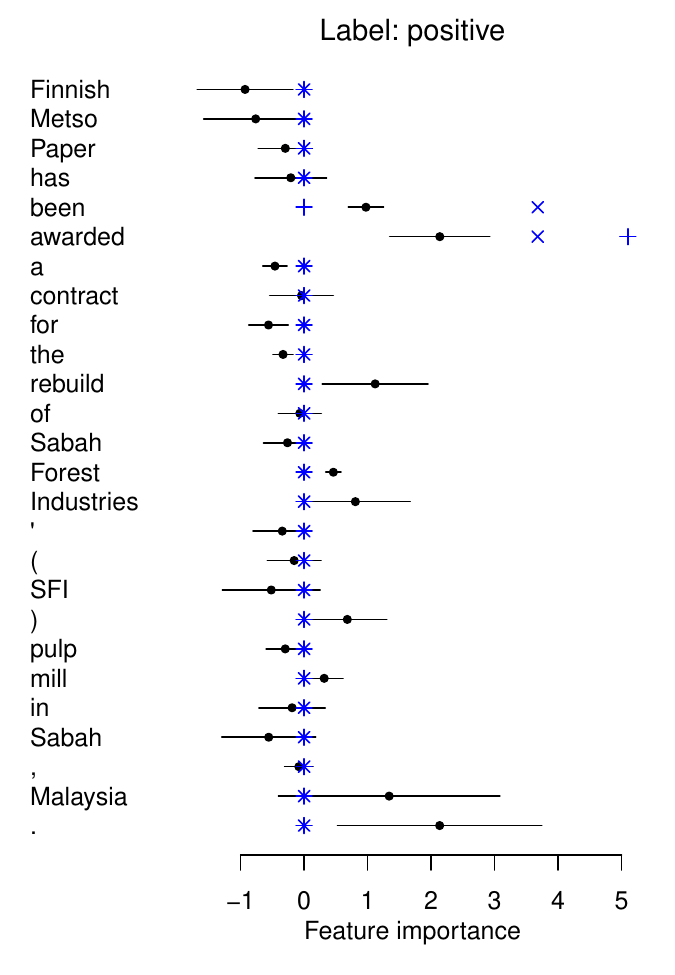}

\includegraphics[width=\ww\linewidth]{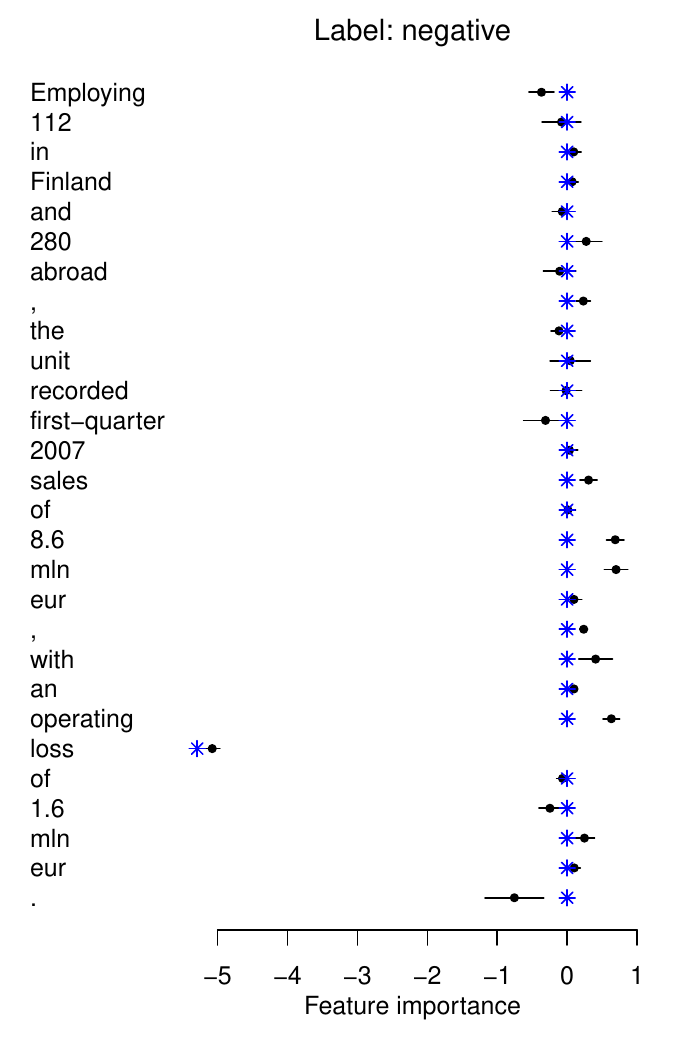}
\includegraphics[width=\ww\linewidth]{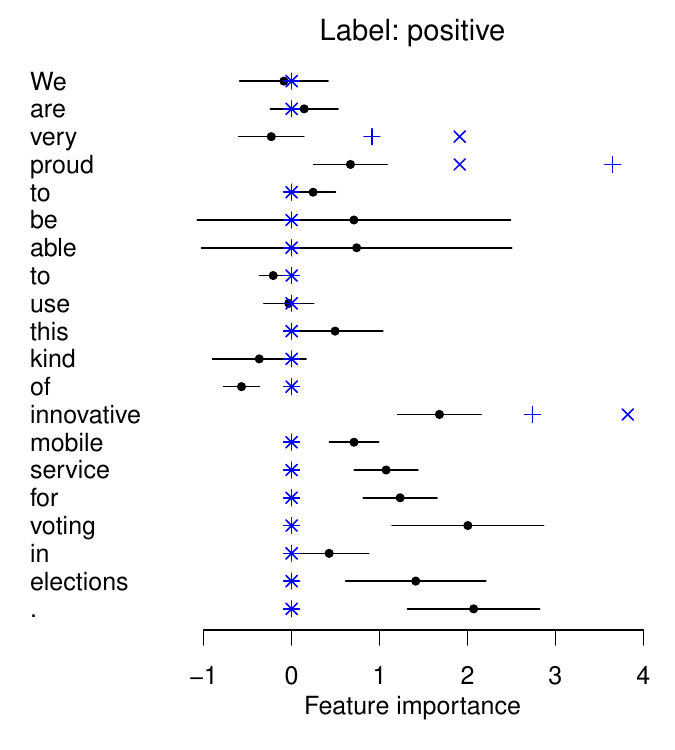}
\includegraphics[width=\ww\linewidth]{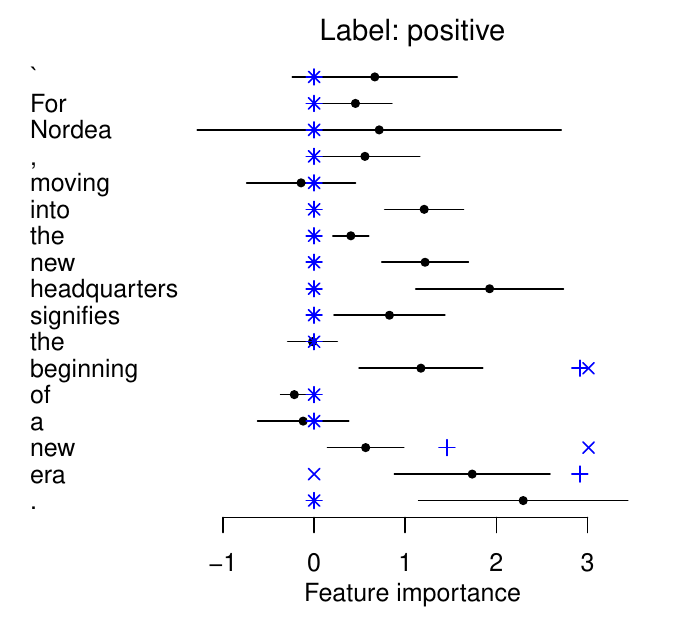}

\caption{\small \captionexplain}
\label{word_level_explain_appx_2}
\end{figure}

\begin{figure}[h!t]
\centering

\includegraphics[width=\ww\linewidth]{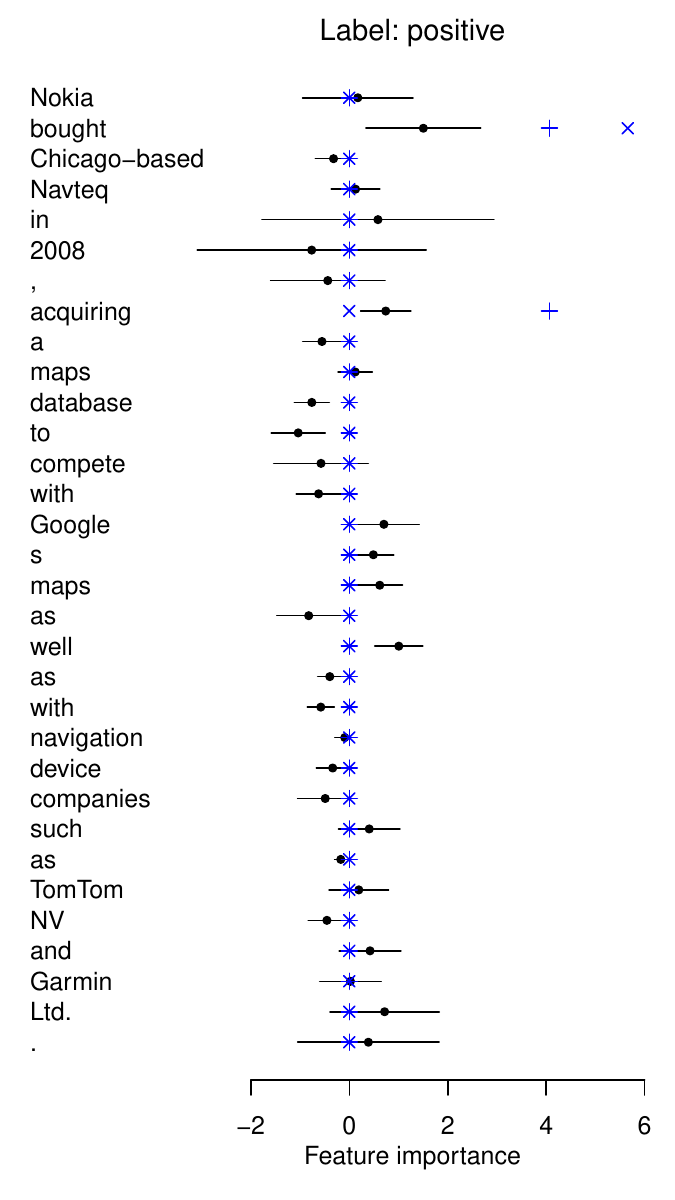}
\includegraphics[width=\ww\linewidth]{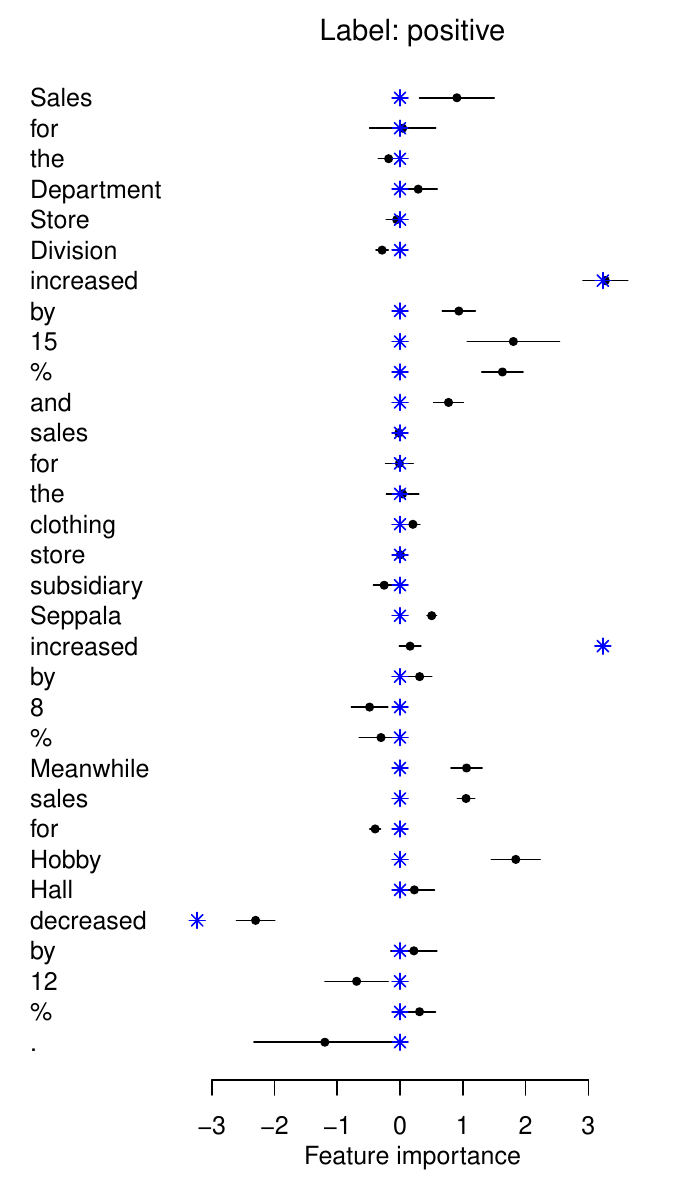}
\includegraphics[width=\ww\linewidth]{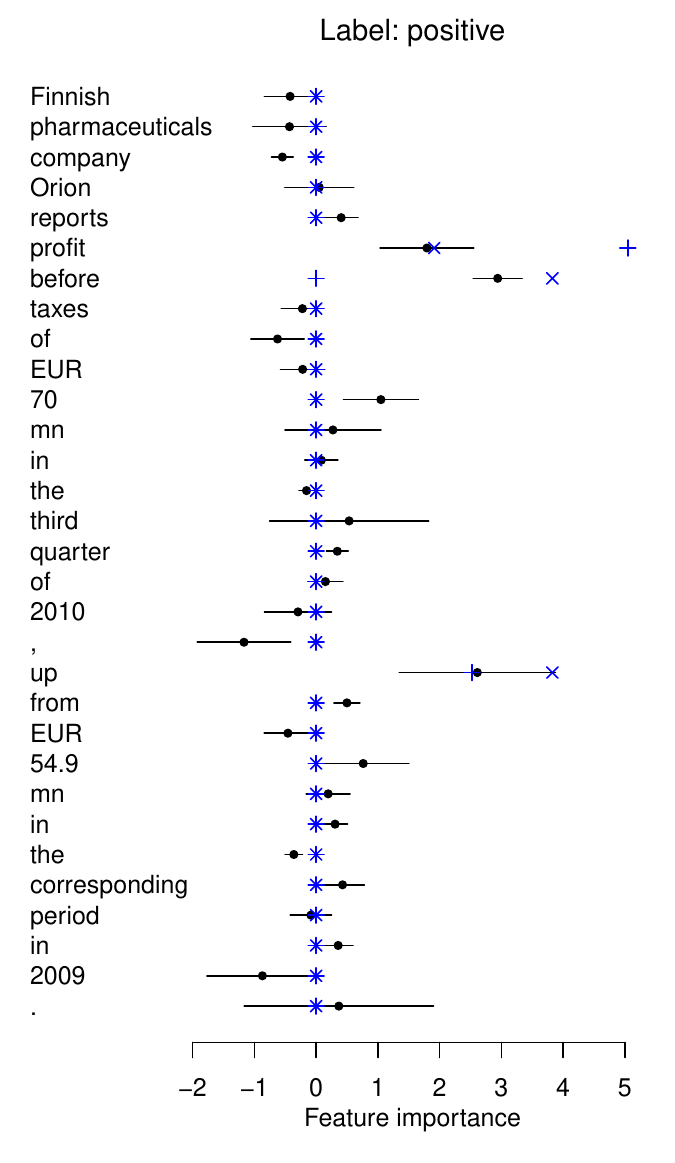}

\includegraphics[width=\ww\linewidth]{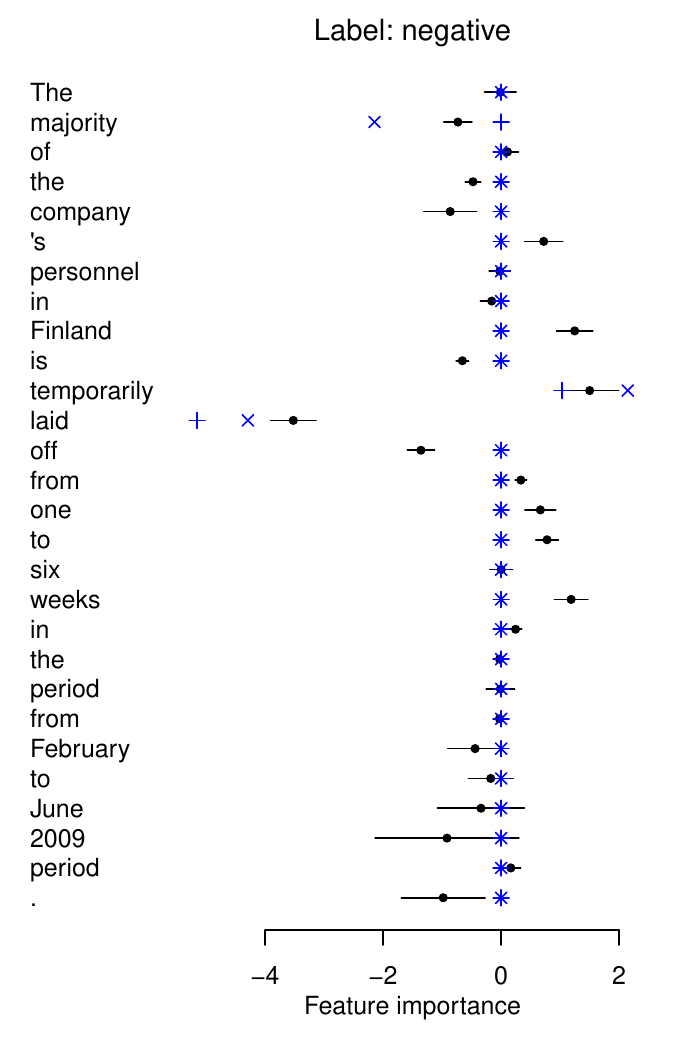}
\includegraphics[width=\ww\linewidth]{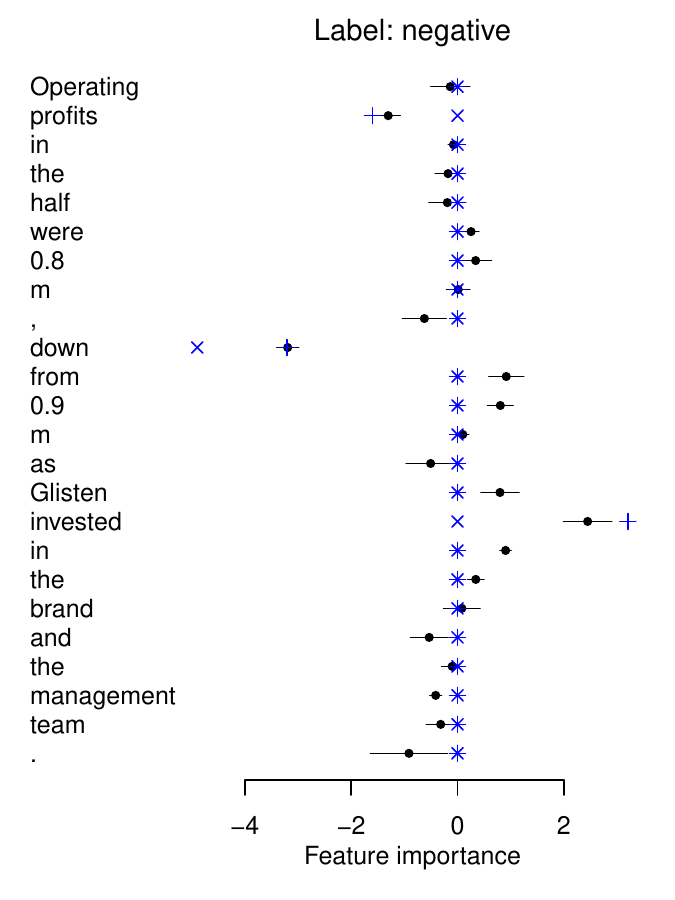}
\includegraphics[width=\ww\linewidth]{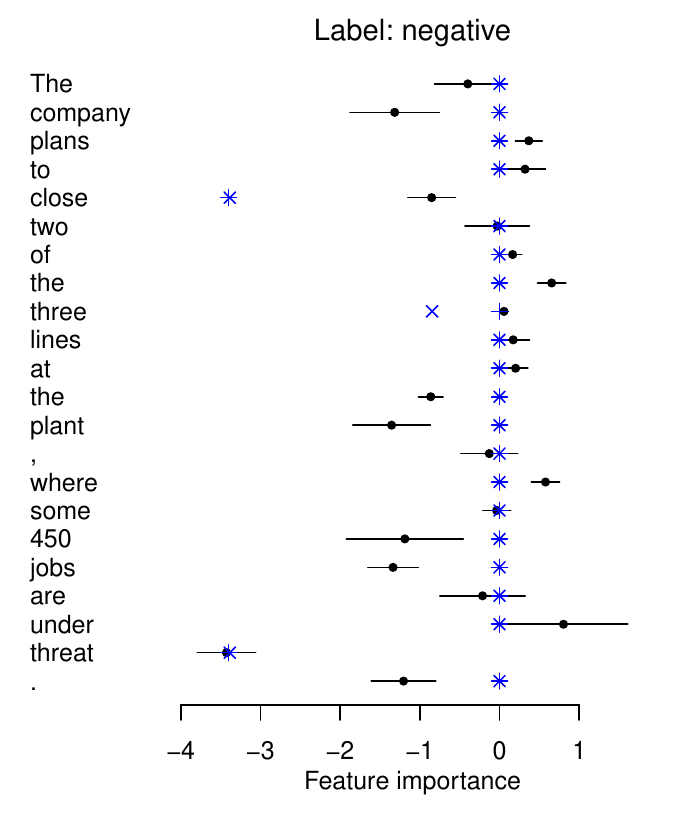}
\caption{\small \captionexplain}
\label{word_level_explain_appx_3}
\end{figure}

\end{document}